\documentclass[twoside]{article}
\usepackage[accepted]{aistats2024}

\usepackage{booktabs}
\usepackage{algorithm}
\usepackage[noend]{algorithmic}

\usepackage{amsthm}
\usepackage{amssymb}
\usepackage{color}
\usepackage{multirow}
\usepackage{xcolor,colortbl}
\usepackage{enumitem}
\usepackage{graphicx}
\usepackage[hidelinks]{hyperref}

\usepackage[round]{natbib}
\newcounter{mainsection}

\theoremstyle{plain}
\newtheorem{theorem}{Theorem}[section]
\newtheorem{proposition}[theorem]{Proposition}
\newtheorem{lemma}[theorem]{Lemma}

\newtheorem{maintheorem}{Theorem}[mainsection]
\newtheorem{mainproposition}[maintheorem]{Proposition}

\theoremstyle{definition}
\newtheorem{definition}[theorem]{Definition}
\newtheorem{assumption}[theorem]{Assumption}

\newtheorem{mainassumption}[maintheorem]{Assumption}

\theoremstyle{remark}
\newtheorem*{remark}{Remark}

\renewcommand{\paragraph}[1]{\textbf{#1}\ \ }

\def\ddefloop#1{\ifx\ddefloop#1\else\ddef{#1}\expandafter\ddefloop\fi}
\def\ddef#1{\expandafter\def\csname bb#1\endcsname{\ensuremath{\mathbb{#1}}}}
\ddefloop ABCDEFGHIJKLMNOPQRSTUVWXYZ\ddefloop

\def\ddef#1{\expandafter\def\csname c#1\endcsname{\ensuremath{\mathcal{#1}}}}
\ddefloop ABCDEFGHIJKLMNOPQRSTUVWXYZ\ddefloop

\def\ddef#1{\expandafter\def\csname v#1\endcsname{\ensuremath{\boldsymbol{#1}}}}
\ddefloop ABCDEFGHIJKLMNOPQRSTUVWXYZabcdefghijklmnopqrstuvwxyz\ddefloop

\def\ddef#1{\expandafter\def\csname v#1\endcsname{\ensuremath{\boldsymbol{\csname #1\endcsname}}}}
\ddefloop {alpha}{beta}{gamma}{delta}{epsilon}{varepsilon}{zeta}{eta}{theta}{vartheta}{iota}{kappa}{lambda}{mu}{nu}{xi}{pi}{varpi}{rho}{varrho}{sigma}{varsigma}{tau}{upsilon}{phi}{varphi}{chi}{psi}{omega}{Gamma}{Delta}{Theta}{Lambda}{Xi}{Pi}{Sigma}{varSigma}{Upsilon}{Phi}{Psi}{Omega}{ell}\ddefloop

\DeclareMathOperator*{\argmin}{argmin}

\definecolor{Gray}{gray}{0.90}
\newcolumntype{a}{>{\columncolor{Gray}}c}

\newcommand{\meansd}[2]{#1{ \tiny(#2)}}

\def\agg{\text{Agg}}

\def\method{BOBA}
\def\hsimp{honest simplex}
\def\hsubs{honest subspace}
\newcommand{\rev}[1]{#1}

\begin{document}

\runningauthor{Wenxuan Bao \unskip{,}\enspace Jun Wu \unskip{,}\enspace Jingrui He}

\twocolumn[
\aistatstitle{BOBA: Byzantine-Robust Federated Learning with Label Skewness}
\aistatsauthor{Wenxuan Bao$^{1}$ \And Jun Wu$^{1}$ \And Jingrui He$^{1}$}
\aistatsaddress{$^{1}$University of Illinois Urbana-Champaign \\ \texttt{\{wbao4,junwu3,jingrui\}@illinois.edu}}
]

\addtocontents{toc}{\protect\setcounter{tocdepth}{0}}  

\begin{abstract}
  In federated learning, most existing robust aggregation rules (AGRs) combat Byzantine attacks in the IID setting, where client data is assumed to be independent and identically distributed. In this paper, we address label skewness, a more realistic and challenging non-IID setting, where each client only has access to a few classes of data. In this setting, state-of-the-art AGRs suffer from selection bias, leading to significant performance drop for particular classes; they are also more vulnerable to Byzantine attacks due to the increased variation among gradients of honest clients. To address these limitations, we propose an efficient two-stage method named \textit{\method}. Theoretically, we prove the convergence of {\method} with an error of the optimal order. Our empirical evaluations demonstrate {\method}'s superior unbiasedness and robustness across diverse models and datasets when compared to various baselines. \rev{Our code is available at} \url{https://github.com/baowenxuan/BOBA}. 
\end{abstract}
\section{INTRODUCTION} \label{sec:introduction}

Federated learning (FL) \citep{FedAvg} is a machine learning system where multiple clients collaboratively train a global model under the orchestration of a central server, without sharing their own private and sensitive data. It has wide applications in sales, finance, healthcare \citep{con_and_app}, etc. However, FL systems are vulnerable to attacks and failures \citep{advance,privacy_and_robustness}. Notably, \textit{Byzantine attacks} can send arbitrary gradients to the server, causing sub-optimal convergence or even divergence \citep{krum}. To defend against Byzantine attacks, a common approach is to replace gradient averaging with robust aggregation rules (AGRs) \citep{geomed,trmean}. These methods have demonstrated their effectiveness in achieving Byzantine-robustness when client data adheres to the independent and identically distributed (IID) assumption. However, in practical applications, client data often deviate from the IID pattern \citep{FedAvg,survey_field,advance}. This non-IIDness introduces increased variation among honest clients' gradients, posing challenges in detecting and excluding Byzantine clients. 

Our work mainly focuses on label skewness, a typical non-IID setting where each client only has access to a few classes of data \citep{FedRS,class_imbalance}. In this setting, while clients share the same conditional data distribution given labels, their label distributions can vary a lot. For instance, in animal image classification, users from various regions may capture images of distinct species prevalent in their areas, even though these species share similar visual characteristics. Label skewness introduces two key challenges for model performance. First, it introduces a selection bias of clients, causing robust AGRs to favor certain clients over others, thus biasing the model. Secondly, it amplifies the variation among gradients of honest clients, making AGRs more vulnerable to Byzantine attacks. Thus more advanced techniques are required to tackle these challenges. 

Focusing on label skewness, we find that the gradients of honest clients distribute near a $(c-1)$-simplex, where $c$ is the number of classes. Leveraging this insight, we introduce \textit{\method} (Byzantine-rObust and unBiased Aggregator), a two-stage AGR to estimate this simplex effectively. In the first stage, we robustly estimate the low dimensional affine subspace containing the simplex and project all gradients onto the subspace. In the second stage, we use a few data samples on the server to estimate the $(c-1)$-simplex and further filter out potential Byzantine gradients. As a result, {\method} ensures that honest gradients remain largely unaffected, with only minor perturbations, while Byzantine gradients are either discarded or significantly weakened. Our contributions can be summarized as follows:
\begin{itemize}[topsep=0pt,itemsep=0pt]
    \item We make a systematic analysis of FL robustness challenges in the presence of label skewness, including the identification of two key challenges: selection bias and increased vulnerability (Sec. \ref{sec:challenge}).
    \item We introduce {\method}, which incorporates an objective addressing label skewness and robustness, along with an efficient optimization algorithm (Sec. \ref{sec:algorithm}).
    \item We provide theoretical analysis that derives gradient estimation error and convergence guarantee, demonstrating \method's unbiasedness and optimal order robustness (Sec. \ref{sec:theory}).
    \item We empirically evaluate the unbiasedness and robustness of {\method} across diverse models, datasets, and attack scenarios, outperforming various baseline AGRs and extending to more complex non-IID settings (Sec. \ref{sec:experiment}).
\end{itemize}
\section{RELATED WORKS}
\label{sec:related_works}

\paragraph{Robust AGRs with IID clients} 
Extensive research has been conducted on robust AGRs tailored for IID clients. These AGRs modify the server's gradient averaging step, and can be categorized into two main groups: majority-based and reference-based methods.
\textit{Majority-based AGRs} operate under the assumption that the gradients of honest clients tend to cluster together. They employ robust mean estimators, including coordinate-wise median \citep{trmean}, geometric median \citep{geomed, rfa}, and Krum \citep{krum}, to identify a vector close to the majority of gradients. While these methods have been theoretically proven to perform well in IID settings, our analysis reveals that they exhibit issues such as selection bias and increased vulnerability when confronted with label skewness scenarios. 
In many FL systems, the server possesses a limited amount of data \citep{nonIID,FedDF}. Although this data may be insufficient to independently train a satisfactory model, reference-based AGRs leverage server data to assess each client's update and adjust their contributions to enhance robustness. Loss-based rejections \citep{rej} evaluate client updates with their loss on server data, and drop clients whose updates are the most harmful. Zeno \citep{zeno} extends this approach by considering both loss and gradient scales. FLTrust \citep{fltrust} computes a server gradient using server data and reweighs client gradients based on their similarity to the server gradient. ByGARS \citep{bygars} optimizes the aggregation weights for client gradients using server data in a meta-learning framework. However, it is worth noting that these methods are not specifically designed to address the non-IID challenges inherent in FL scenarios. 

\paragraph{Robust AGRs with non-IID clients}
A few works have studied robustness with non-IID clients. \cite{bucket} combine IID AGRs with bucketing to enhance homogeneity in AGR inputs, albeit with a trade-off in robustness. Similar to {\method}, RAGE \citep{rage} also uses singular value decomposition (SVD) for robust aggregation. However, it uses SVD to remove Byzantine clients iteratively, whereas our work focuses on applying SVD to model the distribution of honest clients' gradients. \cite{cluster} group clients into IID clusters and train global models in each group. A topic related to selection bias is performance fairness, where each client should have similar accuracy. \cite{FedMGDA+} introduce a multi-task learning framework to learn a robust and fair global model. However, it is not robust to Byzantine attacks and can only guarantee Pareto optimal. Ditto \citep{ditto} learns personalized models to achieve fairness and robustness, but still requires training a robust global model. 

For additional related works on FL with label skewness and non-IIDness, please refer to Appendix \ref{appendix:related_works}.

\section{FL WITH LABEL SKEWNESS}
\label{sec:challenge}

\def\redsquare{\textcolor{red}{\rule{0.4em}{0.4em}}}
\def\greendot{\textcolor{green!60!black}{$\bullet$}}


\paragraph{Setup}
We study the FedSGD  \citep{FedAvg} system consisting of one central server and $n$ clients. Each client is either \textit{honest} (in honest set $\cH$) or \textit{Byzantine} (in Byzantine set $\cB$), with $|\cH|$ and $|\cB|$ representing the \textit{real} number of honest and Byzantine clients, respectively. In each communication round, the server broadcasts the parameter ${\vw}_G \in \bbR^d$ to all clients. Each honest client $i \in \cH$ computes the gradient with its own data $\{\vxi_{ij}\}_{j=1}^{m_i}$ sampled from $P_i$ and sends back the \textit{honest gradient} $\vg_i = \nabla_{\vw_G} \cL_i(\vw_G)$, where $\cL_i(\vw_G) = \frac{1}{m_i}\sum_{j=1}^{m_i} \ell(\vw_G; \vxi_{ij})$ and $\ell$ is the loss function. Each Byzantine client can send arbitrary \textit{Byzantine gradient} to the server. Finally, the server aggregates all $n$ gradients $\hat\vmu = \agg(\{\vg_i\}_{i=1}^n)$ and updates the parameter $\vw_G \leftarrow \vw_G - \eta \hat\vmu$, where $\agg(\cdot)$ is the aggregation rule (AGR), and $\eta$ is the learning rate. 

For each honest client $i \in \cH$, let $\bbE \vg_i$ be its \textit{expected gradient}, where the expectation is taken on data sampling from $P_i$. During training, the system minimizes the empirical risk, $\frac{1}{|\cH|} \sum_{i \in \cH} \cL_i(\vw_G)$. FL aims to train a model with low population risk, $\frac{1}{|\cH|} \sum_{i \in \cH} \bbE \cL_i(\vw_G)$. Let $\vmu = \frac{1}{|\cH|} \sum_{i \in \cH} \vg_i$ denote the gradient of empirical risk and $\bbE \vmu = \frac{1}{|\cH|} \sum_{i \in \cH} \bbE \vg_i$ denote its expectation, which is also the gradient of population risk. 

\paragraph{Byzantine attack}
In each round, Byzantine clients can send arbitrary vectors to the server, which may depend on current global model $\vw_G$ and honest gradients $\{\vg_i\}_{i \in \cH}$. They can also collude to perform stronger attacks, e.g., by sending the same vector. 

\textbf{Robust AGRs} aim to find a robust estimation of $\bbE \vmu$. Since the server has no prior knowledge about the exact number of Byzantines, we let $f$ be the \textit{Byzantine tolerance}, a hyperparameter such that AGRs guarantee to be robust when $|\cB| \leq f$. Similar to previous works \citep{zeno,fltrust,FedDF}, we assume the AGR has access to small amount of clean data to improve robustness. Notice that such data are collected and labeled by the server, rather than uploaded by clients \citep{fltrust}. Finally, since Byzantines in FL can change their index over rounds, the AGR can only use the information from the current round, including the global model, all clients' uploaded gradients, and server data. 



\subsection{Distribution of Honest Gradients}

\begin{figure}
    \centering
    \includegraphics[width=\linewidth]{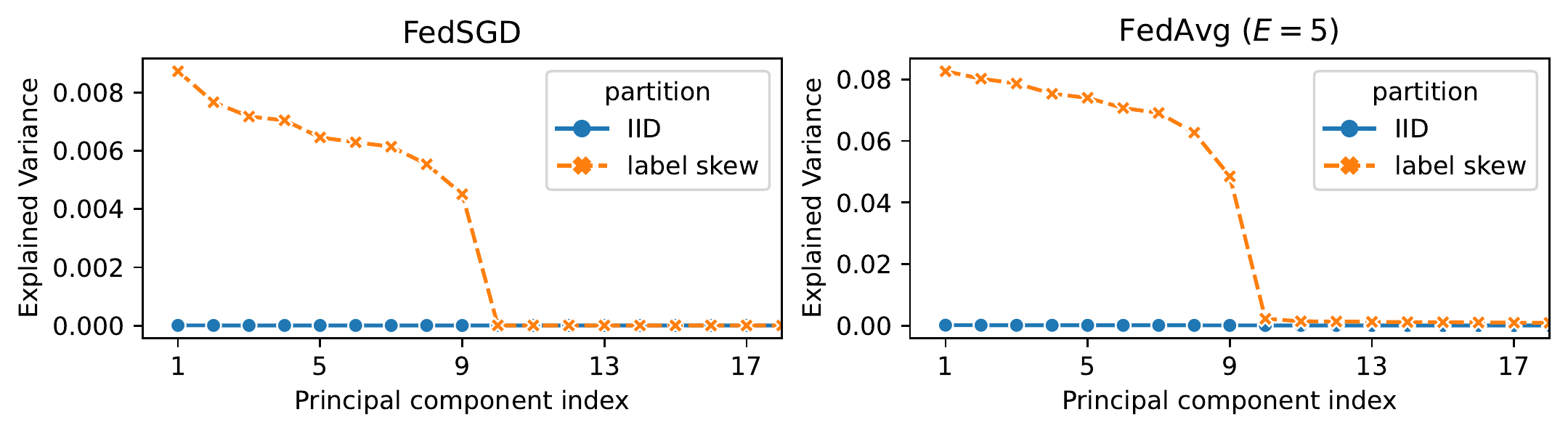}
    \vspace{-3ex}
    \caption{PCA of honest gradients on MNIST ($c = 10$). \rev{Over 99\% of the variance concentrate on the first $(c-1)$ principal components, verifying that honest gradients distribute near the honest subspace.} }
    \label{fig:pca}
    \vspace{-1ex}
\end{figure}

This subsection analyzes the distribution of honest gradients with label skewness. We start with some definitions. 
\begin{definition}[Inner, outer and total variations]
    For an honest client $i \in \cH$, its inner variation is $\bbE \| \vg_i - \bbE \vg_i \|_2^2$; its outer variation is $\| \bbE \vg_i - \bbE \vmu \|_2^2$, and its total variation is $\bbE \| \vg_i - \bbE \vmu \|_2^2$. 
\end{definition}
Inner variation measures the randomness of sampling data from client $i$'s local distribution $P_i$, while outer variation measures the difference between local distribution $P_i$ and global distribution $\frac{1}{|\cH|} \sum_{i \in \cH} P_i$, without randomness. 

In the IID setting, the outer variation is zero, implying that \textit{honest gradients $\{\vg_i\}_{i \in \cH}$ are distributed around the same center $\bbE \vmu$}. However, this implication does not hold under label skewness, since the outer variations are non-zero. We formally define label skewness and analyze the distribution of honest gradients. 

\begin{definition}[$c$-label skew distribution]
The data distributions $\{P_i\}_{i\in \cH}$ of honest clients are considered $c$-label skew distributions if they can be expressed as
\begin{align*}
    P_i(\vxi) = \sum_{z=1}^c p_{iz} Q_z(\vxi), \quad \forall i \in \cH
\end{align*}
where $P_i(\vxi)$ is the data distribution of client $i$, the label $z$ can take $c$ finite values, $p_{iz} \geq 0$ is the label distribution of client $i$ subject to $\sum_{z=1}^c p_{iz} = 1$, and $Q_z(\vxi) = P_i(\vxi | z)$ represents the conditional distribution given label $z$. Different clients share the same $\{Q_z(\vxi)\}_{z=1}^c$ while having distinct label distributions $\vp_i = [p_{i1}, \cdots, p_{iz}]^\top$. 
\end{definition}

The $c$-label skew distribution assumes the heterogeneity among honest clients can be characterized by their divergence in label distribution. With this condition, we can analyze the distribution of honest gradients.


\begin{proposition}[Expectation of honest gradients]\label{proposition:dist} 
With $c$-label skew distribution, $\forall i \in \cH$, we have \raggedright 
\resizebox{1.0\linewidth}{!}{
\begin{minipage}{\linewidth}
\begin{align*}
    \bbE \vg_i &= \sum_{\vxi} P_i(\vxi) \nabla_{\vw} \cL(\vw; \vxi) = \sum_{\vxi} \sum_{z = 1}^c p_{iz} Q_z(\vxi)\nabla_{\vw}\cL(\vw; \vxi) \\
    &= \sum_{z = 1}^c p_{iz} \nabla_{\vw}  \sum_{\vxi} Q_z(\vxi)\cL(\vw; \vxi) = \sum_{z=1}^c p_{iz} \bbE \vgamma_z 
\end{align*}
\vspace{1ex}
\end{minipage}
}
\hskip-0pt
where $\bbE \vgamma_z = \nabla_{\vw} \sum_{\vxi} Q_z(\vxi)\cL(\vw; \vxi)$ is the expected gradient computed with data from class $z$. 
\end{proposition}

Proposition \ref{proposition:dist} shows that each expected honest gradient is a convex combination of $\{\bbE \vgamma_z\}_{z=1}^c$, forming a $(c - 1)$-simplex in its range. We define the \textit{\hsimp} as $\{\sum_{z=1}^c p_z \bbE \vgamma_z : \sum_{z=1}^c p_z =1, p_z \geq 0 \}$, and the \textit{\hsubs} as $\{\sum_{z=1}^c p_z \bbE \vgamma_z : \sum_{z=1}^c p_z =1 \}$. 

As honest gradients are perturbations of their expectations, \textit{they distribute near the \hsimp}, approximately forming a $(c-1)$-dimensional affine subspace. Thus, if we conduct principal component analysis (PCA) on honest gradients, the variance should concentrate on the first $(c - 1)$ principal components. Figure \ref{fig:pca} verifies our finding on MNIST \citep{mnist}. Appendix \ref{appendix:exp:more_fl} gives details of this experiment. 



\begin{figure*}
    \centering
    \includegraphics[width=0.85\linewidth]{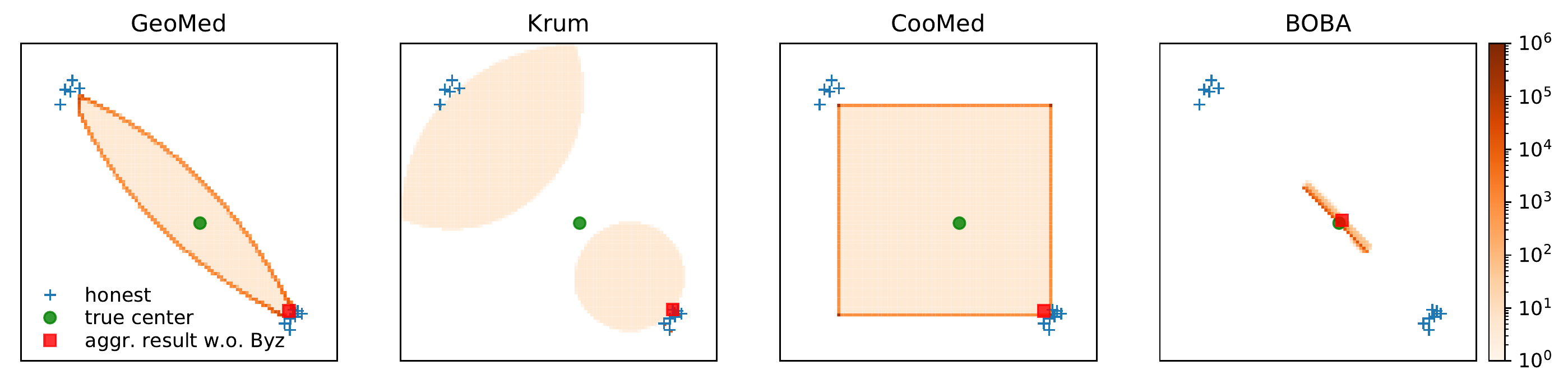}
    \vspace{-1ex}
    \caption{Comparison of aggregation results. (1) \textit{Selection bias}: Without attacks, the aggregation results ({\redsquare}) for GeoMed, Krum and CooMed are biased toward the majority class in the lower-right corner and deviate from the honest gradient center ({\greendot}), indicating their large biases. Meanwhile, {\method} is unbiased. (2) \textit{Increased vulnerability}: With different attacks, the aggregation results will be different. The orange region represents the heatmap (2D histogram) of possible aggregation results given various attacks, where larger radius indicates worse robustness. {\method} has smallest radius, showing its stronger robustness than IID AGRs. }
    \label{fig:vis_agg}
    \vspace{-1ex}
\end{figure*}

\subsection{Challenges of Label Skewness}

In the IID scenario, each honest gradient serves as an unbiased estimator of $\bbE \vmu$, simplifying the design of robust AGRs which merely require the identification of one honest gradient (or a close Byzantine gradient). However, in label skewness settings, each honest gradient can exhibit substantial deviations from $\bbE \vmu$, giving rise to two key challenges: \textit{selection bias} and \textit{increased vulnerability}. 

\paragraph{Selection bias}
Many robust AGRs, e.g., Krum \citep{krum}, select a subset of gradients for aggregation. With label skewness, these AGRs tend to select some clients more frequently, often discarding clients with higher outer variations or deviates from the majority. This selection bias introduces bias into the aggregation results, \textit{even in the absence of any attacks}. In Figure \ref{fig:vis_agg}, where honest gradients form two clusters, each representing a different class of samples, baseline AGRs consistently choose the majority class. This results in the FL model exclusively training on one class of samples, converging to a trivial solution. 

\paragraph{Increased vulnerability}
With label skewness, baseline AGRs are more vulnerable to attacks, resulting in larger variations in the aggregation results, primarily due to the increased total variation. In Figure \ref{fig:vis_agg}, the aggregation results of baseline AGRs exhibit a considerable range, much larger than the inner variation (variation of each cluster). Interestingly, this vulnerability occurs not only on the direction of outer variation, but also its orthogonal direction. 

In summary, IID AGRs are neither unbiased nor sufficiently robust in the more realistic label skewness setting. It is necessary to design a new robust AGR. 

\section{PROPOSED {\method} ALGORITHM}
\label{sec:algorithm}

\begin{algorithm}[tb]
    \caption{{\method} Framework} \label{alg:BOBA}
    \small
    \begin{algorithmic}[1]
        \REQUIRE $\vG = [\vg_1, \cdots, \vg_n]$, $\vGamma = [\vgamma_1, \cdots, \vgamma_c]$, $n, f, c, p_{\min}$
        \ENSURE Aggregation result $\hat\vmu$
        \STATE Initialize subspace $\hat\cP$: $\vm, \vU, \vSigma, \vV = \text{TrSVD}_{c-1}(\vGamma)$
        \WHILE{not converge}
            \STATE Update $\vr$: $\vG_{[n - f]} = \{ n - f$ gradients in $\vG$ with smallest $\| \vg_i - \Pi_{\hat\cP}(\vg_i) \|_2\}$ where $\Pi_{\hat\cP}(\vg_i) = \vU \vU^\top (\vg_i - \vm) + \vm$
            \STATE Update $\hat\cP$: $\vm, \vU, \vSigma, \vV = \text{TrSVD}_{c-1}(\vG_{[n-f]})$
        \ENDWHILE
        \STATE Encode: $\tilde{\vg_i} = \vU^\top (\vg_i - \vm), \forall i$; $\tilde{\vGamma} = \vU^\top (\vGamma - \vm \boldsymbol{1}^\top)$
        \STATE Estimate: $\hat \vp_i = \left[\begin{matrix}\tilde{\vGamma} \\ \boldsymbol{1}^\top \end{matrix} \right]^{-1} \left[ \begin{matrix} \tilde{\vg}_i \\ 1 \end{matrix} \right], \forall i$ 
        \STATE Filter: $\va = \cA(\{\hat\vp_i\}_{i=1}^n)$
        \STATE Aggregate: $\tilde \vmu = \sum_{i=1}^n a_i \tilde \vg_i / \sum_{i=1}^n a_i$
        \STATE Decode: $\hat\vmu = \vU \tilde\vg_G + \vm$
    \end{algorithmic}
\end{algorithm}

In this section, we propose {\method} and explain its two stages in detail. In stage 1, we robustly find the \hsubs, and project all gradients to this subspace. In stage 2, we estimate the vertices of the \hsimp, reconstruct the label distribution for each client, and drop clients with abnormal label distribution (i.e., with strongly negative entries). Intuitively, all honest gradients will be kept with small perturbation, while all Byzantine gradients will either be weakened (projected to the {\hsimp} in stage 1) or discarded (in stage 2). Therefore, the negative impact of Byzantine gradients can be largely mitigated. 


\subsection{Stage 1: Fitting the Honest Subspace}
\label{subsec:stage1}

The goal of stage 1 is to find a $(c-1)$-dimensional affine subspace close to all honest gradients under the influence of Byzantine gradients. When there are no Byzantines, a standard way to find the subspace is TrSVD, i.e., truncated singular value decomposition on centralized gradients, 
\begin{align*}
    \vm, \vU, \vSigma, \vV = \text{TrSVD}_{c-1}(\vG), \ \text{s.t.}
    \ \vU \vSigma \vV^\top \approx \vG - \vm \boldsymbol{1}^\top 
\end{align*}
where $\vG = [\vg_1, \cdots, \vg_n] \in \bbR^{d \times n}$ is the client gradient matrix, $\vm = \frac{1}{n} \vG \boldsymbol{1} \in \bbR^{d}$ is their average, $\vU \in \bbR^{d \times (c-1)}, \vV \in \bbR^{n \times (c-1)}$ are column-orthogonal and $\vSigma \in \bbR^{(c-1) \times (c-1)}$ is diagonal. TrSVD fits a $(c-1)$-dimensional affine subspace $\cP = \{\vU \vlambda + \vm : \vlambda \in \bbR^{c-1}\}$ minimizing the \textit{reconstruction loss}
\begin{align*}
    \ell(\cP) = \sum_{i=1}^n \left\| \vg_i - \Pi_{\cP}(\vg_i) \right\|_2^2 
\end{align*}
where $\Pi_{\cP}(\vg_i) = \vU \vU^\top (\vg_i - \vm) + \vm$ is a projection function that projects vectors to $\cP$. 

However, vanilla TrSVD is not robust to Byzantine attacks. When there are Byzantine gradients deviating from the \hsubs, the fitted subspace will be dragged to these Byzantine gradients at the cost of underfitting honest ones. For example in $\bbR^2$, when $n$ honest gradients are uniformly distributed on a segment of $\{(x, y): x \in [-1, 1], y = 0\}$. TrSVD will fit a subspace of $\{y = 0\}$. However, one Byzantine gradient of $(100n, 100n)$ can alter the fitted subspace to 
about $\{y = x\}$. Therefore, we design a new objective. 

\paragraph{Objective}
We design \textit{trimmed reconstruction loss} to robustify TrSVD: 
\begin{align*}
    \ell_t(\cP) = \min_{\tiny{\substack{\vr \in \{0,1\}^n \\ \sum_{i=1}^n r_i = n - f}}} \sum_{i=1}^n r_i \left \| \vg_i - \Pi_{\cP}(\vg_i) \right\|_2^2
\end{align*}
{\method} stage 1 fits an affine subspace $\hat\cP$ by minimizing the trimmed reconstruction loss above, which selects the $n - f$ nearest neighbors ($r_i = 1$) and ignores $f$ gradients furthest from $\hat\cP$ ($r_i = 0$). Intuitively, if Byzantines are far from the $\hat\cP$, they will be ignored so $\hat\cP$ is not affected; if Byzantines are close to $\hat\cP$, the $n - f$ nearest neighbors of $\hat\cP$ still includes at least $n - 2f$ honest gradients, which are enough to reconstruct the {\hsubs} (by Assumption \ref{assumption:singular} in Section 5). We show in Appendix \ref{subsubsec:stage1} that stage 1 is theoretically guaranteed to estimate the {\hsubs} robustly. 

The strongest colluding Byzantines may focus on another dimension different from the $c-1$ honest dimensions. But {\method} stage 1 will not identify the Byzantine dimension as honest. If it makes such a mistake, the $n - f$ nearest neighbors will form a $c$-dimensional affine subspace, including one Byzantine dimension and $c-1$ honest dimensions (since there are at least $n - 2f$ honest gradients in the $n - f$ nearest neighbors). Conducting TrSVD on these $n - f$ nearest neighbors results in large loss proportional to the outer variation, which is clearly sub-optimal. Meanwhile, correctly identifying all honest dimensions results in a loss unrelated to outer variations, which is much smaller. In our experiments, we also show that {\method} can resist such colluding Byzantines, e.g. IPM \citep{inner_prod} and LIE \citep{little}. 

\paragraph{Optimization}
To minimize trimmed reconstruction loss, we solve a joint minimization problem
\begin{align*}
    \hat\cP, \hat{\vr} = \argmin_{\tiny{\substack{\cP, \vr \in \{0,1\}^n \\ \sum_{i=1}^n r_i = n - f}}}  \ell_t (\cP, \vr) = \sum_{i=1}^n r_i \left \| \vg_i - \Pi_{\cP}(\vg_i) \right\|_2^2 \\[-20pt]
\end{align*}
Fixing $\cP$, the optimal $\vr$ selects the $n-f$ nearest neighbors of $\cP$; while fixing $\vr$, the optimal $\cP$ can be fitted by conducting TrSVD on the selected $n -f$ gradients. A naive way to minimize trimmed reconstruction loss is \textit{exhaustive searching} (BOBA-ES), which iterates every possible value of $\vr$, conducts TrSVD to fit $\cP$, and chooses the $\cP$ with the smallest trimmed reconstruction loss. It can guarantee the global minimum but have exponentially high computational complexity. 

Instead, we use \textit{alternating optimization}, with details in lines 2 - 4 in Algorithm \ref{alg:BOBA}. It alternatively updates $\cP$ and $\vr$ until convergence. Although the global minimum may not be guaranteed, alternating optimization can converge to a high-quality local minimum with just a few steps. Thus, it is more efficient and practical for large-scale FL. 

After minimization, we project every gradient to the fitted subspace $\hat\cP$. The projection can weaken Byzantine gradients by eliminating its components orthogonal to $\hat\cP$; meanwhile, it only introduces small bounded perturbation to honest gradients. However, only applying stage 1 does not fully guarantee robustness: a Byzantine may still have large components along $\hat\cP$ that bias the aggregation. We design stage 2 to further rule out such Byzantine gradients. 

\subsection{Stage 2: Finding the Honest Simplex}

In stage 2, {\method} uses a small amount of server data to estimate $c$ vertices of the \hsimp, and estimates the label distribution of each client. Gradients with negative entries in the label distribution lie outside the {\hsimp}, and will be discarded. 

Proposition \ref{proposition:dist} shows that each vertex of the \hsimp\ is the expected gradient computed with one class of data. Thus, we initialize $c$ virtual clients on the server, each with one class of data, and compute \textit{server gradients} $\{\vgamma_z\}_{z=1}^c$ with the same process of honest clients. To estimate the label distribution of a client $i$, we solve for $\{\hat{p}_{iz}\}_{z=1}^n$, s.t.
\begin{align*}
    \sum_{z=1}^c \hat p_{iz} \Pi_{\hat\cP}(\vgamma_z)= \Pi_{\hat\cP}(\vg_i) , \quad \sum_{z=1}^c \hat p_{iz} = 1 \\[-20pt]
\end{align*}
Solving this linear system in the gradient space $\bbR^d$ is inefficient. Instead, we split the projection into two steps: encoding ($\tilde{\vg}_i = \vU^\top (\vg_i - \vm)$) and decoding ($\Pi_{\hat\cP}(\vg_i) = \vU\tilde{\vg}_i + \vm$), and solve the linear system in the latent space $\bbR^{c-1}$, which has an explicit solution (see line 6 in Algorithm \ref{alg:BOBA}). 

If our estimation is perfect (e.g., when $\vg_i = \bbE \vg_i, \vgamma_z = \bbE \vgamma_z$), $\hat\vp_i$ will lie in the probability simplex, i.e. $\{\vp : \boldsymbol{1}^\top \vp = 1, \vp \geq \boldsymbol{0}\}$ if client $i$ is honest, while it can be arbitrary if client $i$ is Byzantine. So we can discard clients with negative entries in $\hat\vp_i$, since they must be Byzantines. However in practice, our estimation has a bounded error (Appendix \ref{subsubsec:stage2}). Thus, if an honest client does not have data from a class, which is very common, it can also have a slightly negative entry. Therefore, we design an acceptance criterion 
\begin{align*}
    \va = \cA(\{\hat\vp_i\}_{i = 1}^n),\ \ \text{where } a_i = \bbI\{\min_{z} p_{iz} \geq p_{\min} \} \\[-20pt]
\end{align*}
where $\bbI$ is the indicator function and $p_{\min} \leq 0$ is a hyper-parameter deciding the threshold of rejecting Byzantines. In our implementation, we will accept $n-f$ clients with largest $\min_{z = 1}^c p_{iz}$ if $\sum_{i=1}^n a_i \leq n - f$ (i.e., our acceptance criterion drops too many clients), since there should be at least $n-f$ honest clients. After dropping Byzantines ($a_i = 0$), we average the remaining projected gradients as the aggregation result of \method. 

\subsection{Computational complexity} 
\label{subsec:complexity}

The computational complexity of {\method} is $\cO(kcnd)$, if it conducts TrSVD for $k$ times. The complexity of TrSVD is $\cO(cnd)$ \citep{random_svd}, where $c$ is the number of classes, $n$ is the number of clients and $d$ is the dimension of gradients. When $k, c$ are small constants, {\method} has the same complexity as vanilla averaging, which is very efficient. Practically we also observe that $k$ is very small. In our experiments with MNIST, CIFAR-10 and AG-News, $k = 3.29, 3.20, 4.77$ on average, respectively. \rev{A detailed analysis of the complexity for each step is provided in Appendix \ref{appendix:proof:complexity}.}

\section{THEORETICAL ANALYSIS}
\label{sec:theory}

This section presents the convergence analysis of {\method}. We first establish a connection between convergence and gradient estimation error in Proposition \ref{proposition:convergence}. Subsequently, we demonstrate in Theorem \ref{proposition:boba} that {\method} has bounded gradient estimation error, ensuring guaranteed convergence. Through our analysis, we confirm that the order of {\method}'s gradient estimation error aligns with the lower bound of the gradient estimation error in non-IID setting, surpassing IID AGRs. This illustrates the unbiasedness and optimal order robustness of {\method}. \textit{Detailed proofs are deferred to Appendix \ref{appendix:proofs} due to space limit. } 

\begin{proposition}[Convergence]\label{proposition:convergence}
   With non-negative $L$-smooth population risk $\cL(\vw)$, we conduct SGD with noisy gradient $\hat\vmu = \hat{g}(\vw)$ and step size $\eta = \frac{1}{L}$. If the gradient estimation error $\bbE \| \hat\vmu - \bbE\vmu\|_2^2 = \bbE \| \hat{g}(\vw) - \nabla \cL(\vw) \|_2^2 \leq \Delta^2$ for all $\vw$, then for any weight initialization $\vw^{(0)}$, after $T$ steps, 
    \begin{align*}
        \frac{1}{T} \sum_{t=0}^{T-1} \bbE \left\| \nabla \cL(\vw^{(t)}) \right\|_2^2 
         \leq 2\frac{L}{T} \cL(\vw^{(0)}) + \Delta^2 \\[-20pt]
    \end{align*}
\end{proposition}

Proposition \ref{proposition:convergence} shows that with a robust AGR featuring bounded gradient estimation error, FedSGD converges to a flat region with small gradient in expectation, with convergence rate $\frac{1}{T}$ and error rate $\Delta^2$. Essentially, a smaller gradient estimation error contributes to improved model convergence. Subsequently, we proceed to derive the gradient estimation error of {\method}. To facilitate this analysis, we introduce the following assumptions: 

\begin{assumption}[Bounded variations] \label{assumption:variation}
For all $\vw$, 
    \begin{enumerate}[topsep=0pt,itemsep=0ex]
        \item Bounded honest client inner variations: $\exists \epsilon^2$ s.t., \\
        $
            \bbE \| \vg_i - \bbE\vg_i \|_2^2 \leq \epsilon^2, \forall i \in \cH
        $. 
        \item Bounded honest client outer variations: $\exists \delta^2$ s.t., \\
        $
            \| \bbE\vg_i - \bbE\vmu \|_2^2 \leq \delta^2, \forall i \in \cH
        $. 
        \item Bounded server inner variations: $\exists \epsilon_s^2$ s.t.,\\
        $
            \bbE \| \vgamma_z - \bbE \vgamma_z \|_2^2 \leq \epsilon_s^2, \forall z = 1, \cdots, c
        $. 
        \item Bounded server outer variations: $\exists \delta_s^2$ s.t., \\
        $
            \| \bbE\vgamma_z - \bbE\vmu \|_2^2 \leq \delta_s^2, \forall z = 1, \cdots, c
        $. 
    \end{enumerate}
\end{assumption}

Assumption \ref{assumption:variation} is standard in FL \citep{byrdsaga,survey_field}, and is applied to both honest clients and server since they both have clean data. 

\begin{assumption}[Bounded client singular value]\label{assumption:singular}
    There exists $\sigma > 0$ such that for all $\vw$, conducting centralized SVD on any $n - 2f$ expectations of honest gradients, the $(c-1)$-th singular value $\sigma_{c-1} \geq \sigma$. 
\end{assumption}

Assumption \ref{assumption:singular} is a natural extension of the standard ``$n - 2f > 0$'' assumption prevalent in IID AGRs \citep{krum,trmean,geomed}. This extension entails that, with $c$-label skewness, it is imperative for \textit{all honest components to simultaneously outweigh the Byzantine component}. To fulfill this requirement, removing any arbitrary subset of $f$ clients from the set of $n-f$ honest clients should still ensure that the remaining $n - 2f$ honest clients affinely span the \hsubs, indicated by $\sigma_{c-1} \geq \sigma, \exists \sigma > 0$. Failure to meet this condition could empower Byzantines to form a cluster to replace an honest component.

Assumption \ref{assumption:singular} also reveals that the robustness of an FL system with label skewness depends not only on $n, f$ and $c$, but also on the label distribution for each honest client. Considering $c = 2$, when $\{\bbE \vg_i\}_{i \in \cH}$ distributes uniformly on the {\hsimp} (a line segment), Assumption \ref{assumption:singular} holds as long as $n - 2f > 1$ (i.e., $f < \frac{n-1}{2}$), closely resembling the IID setting. However, when half of the honest clients have only positive samples, while the other half have only negative samples, $\{\bbE \vg_i\}_{i \in \cH}$ will only be distributed at the two vertices of the \hsimp. In this case, Assumption \ref{assumption:singular} only holds when $n - 2f > \frac{n - f}{2}$ (i.e., $f < \frac{n}{3}$).



With label skewness, a gradient distributed around the honest simplex can either be an honest gradient or a Byzantine gradient mimicking honest gradients to bias the aggregation without being detected. However, it is impossible for \textit{any} AGR to distinguish between the two scenarios. Consequently, this introduces an inevitable component to the gradient estimation error as demonstrated in Proposition \ref{proposition:lower}. 

\begin{proposition}[Lower bound of gradient estimation error for any AGR]\label{proposition:lower}
    Given any AGR, we can find $|\cH|$ honest gradients and $|\cB|$ Byzantine gradients, such that $\bbE \|\hat\vmu - \bbE \vmu \|_2^2 \geq \Omega(\beta^2 \delta^2)$, where $\beta = \frac{|\cB|}{n} = \frac{|\cB|}{|\cH| + |\cB|}$ is the fraction of Byzantine clients. 
\end{proposition}

We finally derive {\method}'s gradient estimation error and show that it matches with the error bound above. 

\begin{theorem}[Upper bound of gradient estimation error for BOBA]\label{proposition:boba}
    With Assumptions \ref{assumption:variation} and \ref{assumption:singular}, BOBA has
    \begin{align*}
        \bbE \| \hat\vmu - \bbE\vmu \|_2^2 \leq C_1 \epsilon^2 + C_2 \epsilon_s^2 + C_3 \beta^2 \delta_s^2
    \end{align*}
    where $\beta = \frac{|\cB|}{n}$ is the fraction of Byzantine clients, $C_1 = 4 + 8 ( \frac{1}{n - 2f} + \frac{\delta^2}{\sigma^2} ) (2 (n - f) + |\cH|)$, $C_2 = 16 ( \frac{1}{n - 2f} + \frac{\delta^2}{\sigma^2} ) (n - f) + 16 c (1 + c |p_{\min}|)^2 \beta^2$, $C_3 =  16 (1 + c |p_{\min}|)^2$. 
\end{theorem}

When the outer variation increases $t$ times, both $\delta$ and $\sigma$ increase $t$ times. When all clients are duplicated, $\delta^2$ does not change but $\sigma^2$ is doubled. Thus generally we have $\frac{\delta^2}{\sigma^2} \propto \frac{1}{n}$. When $\epsilon_s = \cO(\epsilon), \delta_s = \cO (\delta)$, $c = \cO(1), \frac{1}{n - 2f} = \cO(\frac{1}{n})$, $|\cH| = \cO(n)$, and $|p_{\min}| = \cO(1)$, we have $\| \hat\vmu - \bbE \vmu \|_2^2 = \cO(\epsilon^2 + \beta^2 \delta^2)$. We conclude that
\begin{itemize}[leftmargin=*,noitemsep,topsep=0pt]
    \item \textit{{\method} is unbiased}. Without attacks, {\method} preserves all honest gradients, resulting in a gradient estimation error unaffected by outer variation $\delta$. 
    \item \textit{{\method} has optimal order robustness}. With attacks, {\method}' gradient estimation error matches the optimal order in Proposition \ref{proposition:lower} in terms of the outer variation $\delta$, while IID AGRs only guarantee $\cO(\epsilon^2 + \delta^2)$ even when $\beta=0$ (see Appendix \ref{appendix:proof:lower_bound_existing_agr}). 
\end{itemize}

A detailed comparison and analysis of the gradient estimation error is presented in Appendix \ref{appendix:proof:grad_error_boba} and \ref{appendix:proof:grad_error_other}. 
\section{EXPERIMENTS}
\label{sec:experiment}

In this section, we conduct experiments to answer the following research questions. 
\begin{itemize}[itemsep=0pt,topsep=0pt]
    \item \textbf{RQ1}: Is {\method} unbiased and more robust to attacks than baseline AGRs? 
    \item \textbf{RQ2}: Is {\method} efficient? 
    \item \textbf{RQ3}: How is {\method} affected by the quality and quantity of server data, \rev{hyper-parameters, and different label skewness settings}? 
    \item \textbf{RQ4}: Can {\method} be extended to more complex non-IID settings and other FL frameworks? 
\end{itemize}

\paragraph{Setup}
We conduct the experiments on a wide range of models and datasets: a 3-layer MLP for MNIST \citep{mnist}, a 5-layer CNN for CIFAR-10 \citep{cifar}, and a GRU network for AG-News \citep{ag_news}. We partition training sets to $|\cH| = 100/100/160$ honest clients respectively with pathological partition \citep{FedAvg}, where each client has data from at most two classes. To evaluate unbiasedness, we use $|\cB| = 0$. To evaluate robustness, we add $|\cB| = 15/15/54$ Byzantine clients \textit{as supplements, not replacements}, resulting in totally $n = 115/115/214$ clients. This design simulates real-world FL systems where adversaries use additional devices to participate in FL training, instead of replacing existing users' devices. Meanwhile, since no data is removed from training, we can directly compare the accuracy with/without Byzantine clients. Appendix \ref{appendix:exp:setup} gives the detailed experimental settings. 

\paragraph{Attacks}
We consider six representative attacks: Gauss \citep{krum}, IPM \citep{inner_prod}, LIE \citep{little}, Mimic \citep{bucket}, MinMax, and MinSum \citep{dnc}. 

\paragraph{Baseline AGRs}
We consider 15 baseline AGRs: 
\begin{itemize}[itemsep=0pt,topsep=0pt] 
    \item \textit{Average} \citep{FedAvg} simply averages all gradients. It is unbiased but vulnerable to attacks. 
    \item \textit{Server} only uses server data to fit a model. We use it to verify that one cannot train a good model with server data only. 
    \item \textit{Majority-based IID AGRs}: coordinate median (CooMed), trimmed mean (TrMean) \citep{trmean}, Krum, Multi-Krum (MKrum) \citep{krum}, and geometric median (GeoMed) \citep{geomed}.
    \item \textit{Reference-based IID AGRs}. SelfRej, AvgRej \citep{rej}, Zeno \citep{zeno}, FLTrust \citep{fltrust} and ByGARS \citep{bygars}. 
    \item \textit{Non-IID AGRs}. Bucketing \citep{bucket} with Krum (B-Krum) or Multi-Krum (B-MKrum), and RAGE \citep{rage}. 
\end{itemize}
All AGRs are set to be robust to $f = 16$ Byzantines on MNIST/CIFAR-10 and $f = 60$ on AG-News. {\method} uses $p_{\min} = -0.5$. 
We assume limited server data: 20 per class for MNIST/CIFAR-10 and 30 per class for AG-News, much fewer than the samples on each client. 

\begin{table}
\vspace{-1ex}
\caption{Evaluation of unbiasedness (mean (s.d.) \% over five random seeds, $|\cH| = 100, 100, 160, |\cB| = 0$) \label{tab:selection_bias}}
\vspace{1ex}
\centering
\resizebox{0.95\linewidth}{!}{
\setlength{\tabcolsep}{4pt}{
\begin{tabular}{ccccccc}
\toprule
\multirow{2}*{Method} & \multicolumn{2}{c}{MNIST} & \multicolumn{2}{c}{CIFAR-10} & \multicolumn{2}{c}{AG-News} \\
\cmidrule(lr){2-3} \cmidrule(lr){4-5} \cmidrule(lr){6-7}
 & Acc $\uparrow$ & MRD $\downarrow$ & Acc $\uparrow$ & MRD $\downarrow$ & Acc $\uparrow$ & MRD $\downarrow$  \\
\midrule
Average     & \meansd{\textbf{92.5}}{0.1} & - 
            & \meansd{\textbf{71.7}}{0.8} & -  
            & \meansd{\underline{88.3}}{0.1} & - \\
Server      & \meansd{82.0}{0.5} & \meansd{18.8}{1.9}  
            & \meansd{24.4}{2.0} & \meansd{61.7}{1.9}  
            & \meansd{82.7}{1.4} & \meansd{8.8}{3.5} \\
CooMed      & \meansd{73.4}{5.8} & \meansd{62.9}{24.3} 
            & \meansd{18.0}{2.8} & \meansd{79.8}{3.3} 
            & \meansd{80.4}{4.5} & \meansd{18.6}{12.0} \\
TrMean      & \meansd{82.3}{2.7} & \meansd{59.4}{20.9} 
            & \meansd{22.3}{11.3} & \meansd{81.4}{2.2} 
            & \meansd{86.9}{0.5} & \meansd{5.8}{3.6} \\
Krum        & \meansd{39.6}{4.3} & \meansd{98.1}{0.2} 
            & \meansd{35.0}{3.0} & \meansd{81.5}{1.9} 
            & \meansd{66.8}{2.9} & \meansd{89.2}{7.0} \\
MKrum       & \meansd{91.7}{0.1} & \meansd{10.0}{2.3} 
            & \meansd{70.5}{0.7} & \meansd{11.1}{3.7} 
            & \meansd{88.0}{0.1} & \meansd{4.6}{2.1} \\
GeoMed      & \meansd{91.9}{0.1} & \meansd{3.1}{0.3}  
            & \meansd{\underline{71.6}}{0.8} & \meansd{\underline{5.1}}{1.1} 
            & \meansd{\textbf{88.4}}{0.1} & \meansd{\underline{0.4}}{0.2} \\
SelfRej     & \meansd{91.7}{0.1} & \meansd{9.6}{0.8} 
            & \meansd{70.1}{1.2} & \meansd{13.5}{6.1} 
            & \meansd{86.6}{1.8} & \meansd{13.5}{9.4} \\
AvgRej      & \meansd{91.1}{0.5} & \meansd{18.1}{8.0} 
            & \meansd{71.0}{0.5} & \meansd{11.2}{6.8} 
            & \meansd{85.8}{0.9} & \meansd{15.6}{6.2} \\
Zeno        & \meansd{91.7}{0.1} & \meansd{10.3}{2.0} 
            & \meansd{70.2}{0.8} & \meansd{11.5}{4.1} 
            & \meansd{86.4}{1.5} & \meansd{14.1}{8.6} \\
FLTrust     & \meansd{85.6}{0.6} & \meansd{18.9}{3.5} 
            & \meansd{53.1}{0.9} & \meansd{32.2}{2.7} 
            & \meansd{86.3}{0.4} & \meansd{5.8}{1.0} \\
ByGARS      & \meansd{76.7}{1.4} & \meansd{59.9}{10.2} 
            & \meansd{32.0}{1.7} & \meansd{60.7}{6.4} 
            & \meansd{44.9}{6.5} & \meansd{82.0}{4.3} \\
B-Krum      & \meansd{73.8}{4.8} & \meansd{93.8}{3.1} 
            & \meansd{59.0}{1.0} & \meansd{81.4}{2.2} 
            & \meansd{87.3}{0.6} & \meansd{5.0}{2.8} \\
B-MKrum     & \meansd{\underline{92.0}}{0.1} & \meansd{\underline{2.9}}{0.5} 
            & \meansd{70.9}{0.8} & \meansd{6.2}{0.9} 
            & \meansd{87.8}{0.3} & \meansd{3.3}{1.5} \\
RAGE        & \meansd{59.8}{0.5} & \meansd{90.1}{0.5} 
            & \meansd{58.3}{1.5} & \meansd{56.4}{10.0} 
            & \meansd{63.9}{6.1} & \meansd{80.2}{5.2} \\
BOBA        & \meansd{\textbf{92.5}}{0.1} & \meansd{\textbf{1.3}}{1.7} 
            & \meansd{70.9}{0.9} & \meansd{\textbf{4.0}}{1.7} 
            & \meansd{\underline{88.3}}{0.1} & \meansd{\textbf{0.2}}{0.1} \\
\bottomrule
\end{tabular}
}
}
\end{table}

\begin{table*}
\vspace{-1ex}
\caption{Evaluation of robustness (Accuracy, mean (s.d.) \% over five random seeds) \label{tab:robustness}}
\vspace{1ex}
\centering
 \resizebox{1.0\linewidth}{!}{
\setlength{\tabcolsep}{2pt}{
\begin{tabular}{cccccccaccccccacccccca}
\toprule
\multirow{2}*{Method} & \multicolumn{7}{c}{MNIST ($|\cH| = 100, |\cB| = 15$)} & \multicolumn{7}{c}{CIFAR-10 ($|\cH| = 100, |\cB| = 15$)} & \multicolumn{7}{c}{AG-News ($|\cH| = 160, |\cB| = 54$)} \\
\cmidrule(lr){2-8} \cmidrule(lr){9-15} \cmidrule(lr){16-22}
 & Gauss & IPM & LIE & Mimic & MinMax & MinSum & Wst
 & Gauss & IPM & LIE & Mimic & MinMax & MinSum & Wst
 & Gauss & IPM & LIE & Mimic & MinMax & MinSum & Wst \\
\midrule
Average     & \meansd{9.8}{0.0} & \meansd{9.8}{0.0} & \meansd{\underline{92.4}}{0.1} 
            & \meansd{\textbf{92.1}}{0.1} & \meansd{90.0}{0.2} & \meansd{90.8}{0.1}
            & 9.8
            
            & \meansd{10.0}{0.0} & \meansd{10.0}{0.0} & \meansd{\underline{68.2}}{0.8} 
            & \meansd{\textbf{70.3}}{0.8} & \meansd{33.2}{5.9} & \meansd{33.1}{5.3}
            & 10.0
            
            & \meansd{25.4}{2.6} & \meansd{25.0}{0.0} & \meansd{\underline{87.5}}{0.2} 
            & \meansd{87.2}{0.3} & \meansd{35.9}{3.6} & \meansd{30.5}{3.0} 
            & 25.0\\
            
CooMed      & \meansd{68.0}{6.9} & \meansd{42.0}{3.7} & \meansd{89.6}{0.3} 
            & \meansd{65.0}{6.2} & \meansd{77.2}{3.1} & \meansd{77.2}{3.1} 
            & 42.0
            
            & \meansd{18.2}{0.8} & \meansd{7.0}{1.3} & \meansd{22.0}{0.8} 
            & \meansd{14.9}{1.9} & \meansd{18.0}{2.3} & \meansd{18.0}{2.3}
            & 7.0
            
            & \meansd{86.0}{0.3} & \meansd{58.6}{9.9} & \meansd{81.7}{0.3} 
            & \meansd{82.2}{1.7} & \meansd{61.2}{17.6} & \meansd{60.9}{17.4} 
            & 58.6 \\

TrMean      & \meansd{91.7}{0.1} & \meansd{63.8}{10.0} & \meansd{88.9}{0.6} 
            & \meansd{83.2}{2.0} & \meansd{88.8}{0.2} & \meansd{88.8}{0.2}
            & 63.8
            
            & \meansd{57.3}{1.5} & \meansd{14.4}{2.6} & \meansd{30.6}{1.5} 
            & \meansd{30.1}{5.1} & \meansd{22.4}{2.4} & \meansd{23.2}{4.1}
            & 14.1
            
            & \meansd{88.1}{0.3} & \meansd{57.5}{7.7} & \meansd{85.2}{0.2} 
            & \meansd{82.4}{3.8} & \meansd{67.5}{16.3} & \meansd{74.4}{5.5} 
            & 57.5 \\
            
Krum        & \meansd{42.6}{3.8} & \meansd{42.6}{3.8} & \meansd{91.3}{0.1} 
            & \meansd{37.2}{6.4} & \meansd{44.0}{5.1} & \meansd{42.9}{4.4} 
            & 37.2
            
            & \meansd{38.4}{1.7} & \meansd{35.9}{3.7} & \meansd{40.1}{2.3} 
            & \meansd{31.8}{3.7} & \meansd{34.0}{2.5} & \meansd{39.1}{2.6} 
            & 31.8
            
            & \meansd{66.3}{1.9} & \meansd{66.8}{1.7} & \meansd{80.3}{1.0} 
            & \meansd{46.6}{0.4} & \meansd{66.2}{2.1} & \meansd{65.7}{3.3} 
            & 46.6 \\
            
MKrum       & \meansd{\underline{92.4}}{0.2} & \meansd{85.3}{5.3} & \meansd{92.0}{0.2} 
            & \meansd{91.4}{0.2} & \meansd{\textbf{92.4}}{0.1} & \meansd{\textbf{92.3}}{0.1}
            & 85.3
            
            & \meansd{71.7}{0.8} & \meansd{50.9}{11.2} & \meansd{66.0}{1.1} 
            & \meansd{69.6}{0.5} & \meansd{\underline{70.1}}{0.3} & \meansd{\underline{60.5}}{3.0} 
            & \underline{50.9}
            
            & \meansd{\underline{88.3}}{0.2} & \meansd{80.7}{6.0} & \meansd{86.6}{0.2} 
            & \meansd{83.4}{0.6} & \meansd{\textbf{88.3}}{0.1} & \meansd{\underline{85.9}}{0.3} 
            & 80.7 \\
            
GeoMed      & \meansd{91.9}{0.1} & \meansd{82.2}{0.5} & \meansd{91.6}{0.1} 
            & \meansd{89.5}{0.3} & \meansd{91.2}{0.1} & \meansd{91.3}{0.1}
            & 82.2
            
            & \meansd{71.5}{0.6} & \meansd{52.6}{2.5} & \meansd{43.9}{2.3} 
            & \meansd{62.1}{0.6} & \meansd{43.5}{3.0} & \meansd{43.4}{2.3}
            & 43.4
            
            & \meansd{\underline{88.3}}{0.1} & \meansd{77.5}{2.9} & \meansd{83.5}{0.2} 
            & \meansd{84.1}{0.2} & \meansd{83.5}{0.3} & \meansd{83.6}{0.3} 
            & 77.5 \\
            
SelfRej     & \meansd{\underline{92.4}}{0.2} & \meansd{71.1}{2.5} & \meansd{92.0}{0.1} 
            & \meansd{91.4}{0.1} & \meansd{87.6}{1.1} & \meansd{88.6}{0.7}
            & 71.5
            
            & \meansd{71.7}{0.9} & \meansd{14.2}{3.3} & \meansd{66.0}{1.2} 
            & \meansd{69.3}{0.9} & \meansd{32.1}{2.3} & \meansd{32.4}{1.9} 
            & 14.2
            
            & \meansd{\textbf{88.4}}{0.1} & \meansd{25.0}{0.0} & \meansd{86.4}{0.3} 
            & \meansd{84.4}{0.8} & \meansd{38.2}{10.8} & \meansd{32.6}{2.3} 
            & 25.0 \\
            
AvgRej      & \meansd{9.8}{0.0} & \meansd{\underline{91.0}}{0.4} & \meansd{91.8}{0.2} 
            & \meansd{90.7}{0.4} & \meansd{\underline{92.3}}{0.1} & \meansd{\underline{92.2}}{0.1} 
            & 9.8
            
            & \meansd{10.0}{0.0} & \meansd{\textbf{70.5}}{0.7} & \meansd{67.0}{1.2} 
            & \meansd{71.6}{0.5} & \meansd{61.7}{5.2} & \meansd{58.6}{4.6} 
            & 10.0
            
            & \meansd{41.1}{7.7} & \meansd{\textbf{88.0}}{0.3} & \meansd{84.6}{0.4}
            & \meansd{\textbf{88.3}}{0.1} & \meansd{40.7}{7.3} & \meansd{41.8}{12.1} 
            & 40.7 \\
            
Zeno        & \meansd{\underline{92.4}}{0.2} & \meansd{71.1}{2.4} & \meansd{92.0}{0.1} 
            & \meansd{91.4}{0.1} & \meansd{87.6}{1.1} & \meansd{88.6}{0.7} 
            & 71.1
            
            & \meansd{71.5}{0.5} & \meansd{14.1}{3.3} & \meansd{65.8}{1.0} 
            & \meansd{69.4}{0.5} & \meansd{32.3}{1.1} & \meansd{31.3}{3.8} 
            & 14.1
            
            & \meansd{\underline{88.3}}{0.1} & \meansd{25.0}{0.0} & \meansd{86.5}{0.2} 
            & \meansd{85.9}{2.1} & \meansd{53.9}{5.4} & \meansd{61.6}{13.3} 
            & 25.0 \\
            
FLTrust     & \meansd{85.6}{0.6} & \meansd{85.6}{0.6} & \meansd{88.4}{0.7} 
            & \meansd{85.5}{0.6} & \meansd{85.8}{0.6} & \meansd{85.6}{0.6} 
            & \underline{85.5} 
            
            & \meansd{53.0}{0.7} & \meansd{52.6}{1.1} & \meansd{48.9}{2.0} 
            & \meansd{53.3}{1.0} & \meansd{52.0}{1.7} & \meansd{51.9}{1.5} 
            & 48.9 
            
            & \meansd{86.2}{0.5} & \meansd{86.2}{0.4} & \meansd{86.2}{0.4} 
            & \meansd{85.7}{0.8} & \meansd{85.8}{0.9} & \meansd{85.8}{0.5} 
            & \underline{85.7} \\
            
ByGARS      & \meansd{76.7}{1.4} & \meansd{87.5}{0.7} & \meansd{85.0}{0.7} 
            & \meansd{77.1}{1.3} & \meansd{76.6}{1.3} & \meansd{76.6}{1.3} 
            & 76.6
            
            & \meansd{31.9}{1.7} & \meansd{53.6}{0.8} & \meansd{30.8}{2.6} 
            & \meansd{32.2}{1.3} & \meansd{26.9}{1.9} & \meansd{26.9}{1.6} 
            & 26.9
            
            & \meansd{45.4}{11.2} & \meansd{48.0}{8.1} & \meansd{44.5}{11.3} 
            & \meansd{77.2}{20.1} & \meansd{59.0}{22.6} & \meansd{40.7}{2.4} 
            & 40.7 \\
            
B-Krum      & \meansd{78.8}{2.8} & \meansd{80.0}{1.0} & \meansd{90.9}{0.4} 
            & \meansd{61.3}{2.2} & \meansd{79.3}{2.9} & \meansd{77.6}{2.5}
            & 61.3
            
            & \meansd{58.1}{2.3} & \meansd{58.1}{1.1} & \meansd{42.4}{2.4} 
            & \meansd{46.0}{2.6} & \meansd{58.8}{0.8} & \meansd{57.8}{1.1}
            & 42.4
            
            & \meansd{\underline{88.3}}{0.1} & \meansd{51.1}{30.0} & \meansd{87.0}{1.2} 
            & \meansd{81.6}{3.8} & \meansd{86.9}{0.4} & \meansd{86.2}{0.6} 
            & 51.1 \\
            
B-MKrum     & \meansd{\underline{92.4}}{0.1} & \meansd{85.4}{1.8} & \meansd{92.2}{0.1} 
            & \meansd{91.4}{0.0} & \meansd{91.8}{0.2} & \meansd{91.1}{0.1} 
            & 85.4
            
            & \meansd{\underline{71.8}}{0.6} & \meansd{32.0}{2.3} & \meansd{66.0}{0.7} 
            & \meansd{\underline{69.7}}{0.8} & \meansd{45.8}{4.9} & \meansd{42.9}{2.7}
            & 32.0
            
            & \meansd{\underline{88.3}}{0.2} & \meansd{24.9}{12.6} & \meansd{85.9}{0.2} 
            & \meansd{84.9}{0.2} & \meansd{63.7}{14.2} & \meansd{60.4}{28.3} 
            & 24.9 \\
            
RAGE        & \meansd{82.6}{1.0} & \meansd{60.5}{0.9} & \meansd{80.6}{14.0} 
            & \meansd{63.9}{2.3} & \meansd{60.4}{0.9} & \meansd{59.8}{0.5} 
            & 59.8
            
            & \meansd{71.7}{0.5} & \meansd{63.7}{1.3} & \meansd{48.3}{2.2} 
            & \meansd{60.2}{1.1} & \meansd{59.6}{3.0} & \meansd{56.8}{1.1}
            & 48.3
            
            & \meansd{28.5}{5.6} & \meansd{69.5}{2.6} & \meansd{61.2}{9.4} 
            & \meansd{48.8}{21.7} & \meansd{70.6}{1.0} & \meansd{65.5}{7.3} 
            & 28.5 \\
            
\method     & \meansd{\textbf{92.5}}{0.1} & \meansd{\textbf{91.6}}{0.2} & \meansd{\textbf{92.5}}{0.2} 
            & \meansd{\underline{91.7}}{0.4} & \meansd{92.0}{0.3} & \meansd{92.0}{0.6}
            & \textbf{91.6}
            
            & \meansd{\textbf{71.9}}{0.5} & \meansd{\underline{70.1}}{0.6} & \meansd{\textbf{69.2}}{0.7} 
            & \meansd{69.3}{1.1} & \meansd{\textbf{71.2}}{0.5} & \meansd{\textbf{71.4}}{0.5}
            & \textbf{69.2}
            
            & \meansd{\underline{88.3}}{0.1} & \meansd{\underline{87.7}}{0.7} & \meansd{\textbf{88.4}}{0.1} 
            & \meansd{\underline{87.3}}{0.3} & \meansd{\underline{88.1}}{0.1} & \meansd{\textbf{88.3}}{0.2} 
            & \textbf{87.3} \\
\bottomrule
\end{tabular}
}
}
\end{table*}

\paragraph{Evaluation of unbiasedness (RQ1)}
We evaluate the unbiasedness with $|\cB| = 0$. Besides accuracy, we introduce max-recall-drop (MRD) as a complement. It computes how the recall scores of each class differ from the model trained with Average (with $|\cB| = 0$) and picks the largest absolute drop. Smaller MRD indicates a less biased AGR. As selection bias may dramatically decrease some classes' recalls while increasing others, MRD can reflect selection bias better than accuracy. As shown in Table \ref{tab:selection_bias}, most baseline AGRs suffer from significant selection bias, resulting in large MRD. Among baselines, GeoMed and B-MKrum aim to retain as many gradients as possible, consequently achieving smaller MRDs. \textit{We observe that {\method} has accuracy very close to Average, and the smallest MRD among all robust AGRs. It verifies the superior unbiasedness of {\method}. }

\paragraph{Evaluation of robustness (RQ1)}
We evaluate the robustness with $|\cB| = 15, 15, 54$ on three datasets respectively with results shown in Table \ref{tab:robustness}. Considering that Byzantines would select the attack strategy that most effectively degrades model accuracy, we summarize the worst-case accuracy for each defense in the ``Wst'' column for a clear comparison. \textit{{\method} significantly improves the worst-case accuracy by \textbf{6.1\%, 18.3\%, 1.6\%} on three datasets, respectively, showing that {\method} has better robustness than baselines.} Interestingly, we observed that some AGRs (e.g., Mkrum and SelfRej) achieve higher accuracy under certain attacks (e.g., Gauss) compared to no attack conditions. This phenomenon arises from these AGRs relying on accurate estimates of the number of attackers. Without attacks, these AGRs overestimate the number of attackers ($f \gg |\cB|$), leading to dropping honest clients. However, with attacks ($f \approx |\cB|$), these AGRs drop fewer honest clients, resulting in higher accuracy. Considering that majority-based AGRs do not use server data, we also study whether server data can further improve their robustness in Appendix \ref{appendix:exp:boost_agr}. We show that server data cannot enhance the most competitive of these AGRs. 

\paragraph{Byzantines within the {\hsimp}}
Byzantine clients can upload vectors on the boundary of the honest simplex, thereby maximizing the bias in the aggregation results without being detected. The Mimic attack is an example of this type of attack. Although this attack cannot be detected by any AGRs, including BOBA, we found that this attack has a limited impact on model accuracy. 

\begin{figure}
    \centering
    \includegraphics[width=0.9\linewidth]{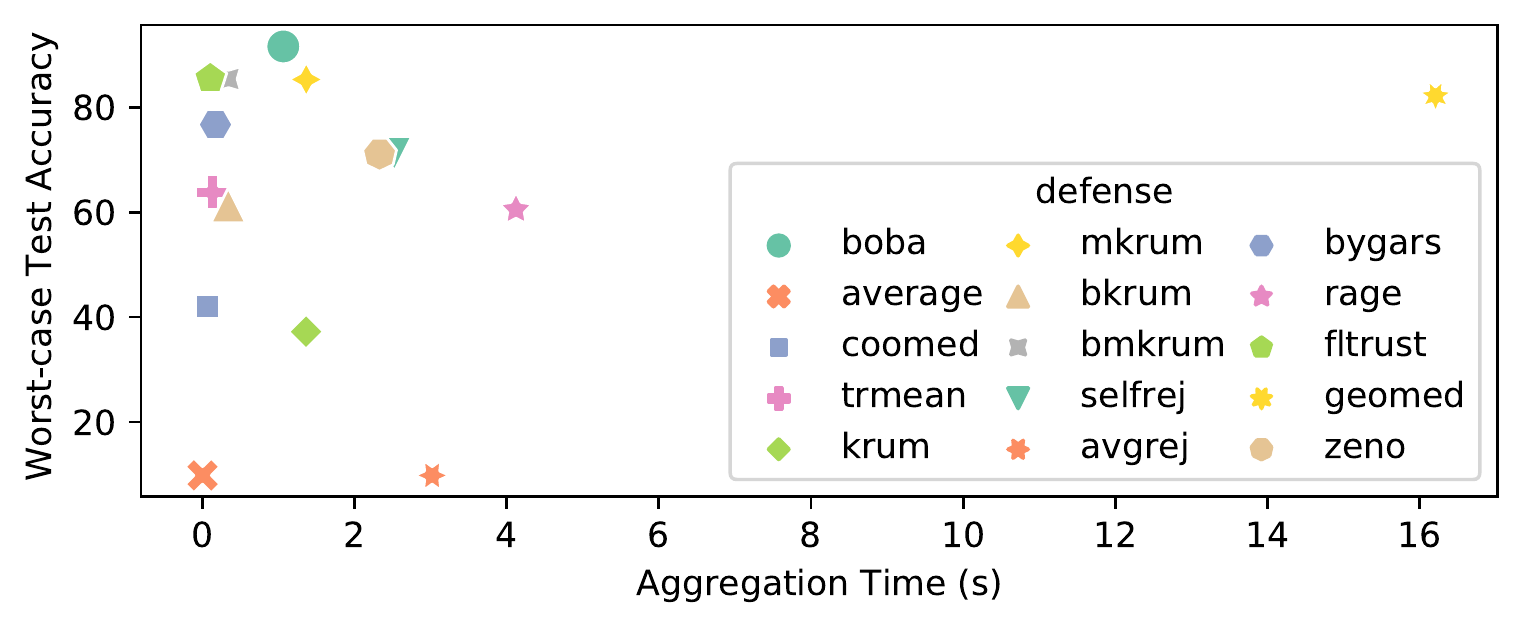}
    \vspace{-1ex}
    \caption{Running time of AGRs on MNIST \label{fig:time_small}}
    \vspace{-1ex}
\end{figure}

\paragraph{Efficiency (RQ2)}
We compare the aggregation time of {\method} with baselines on MNIST. \textit{Figure \ref{fig:time_small} shows that {\method} is faster than half of the baseline AGRs. }

\begin{figure}
    \centering
    \includegraphics[width=0.9\linewidth]{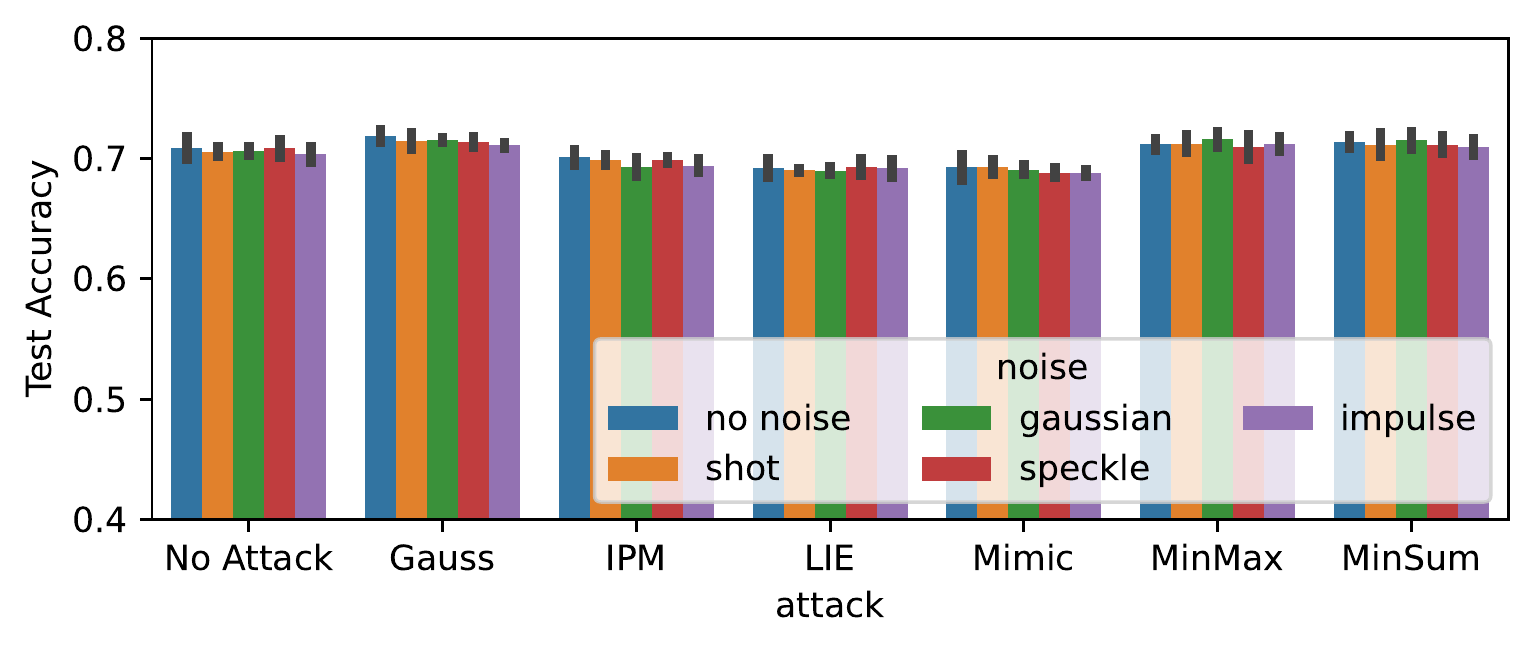}
    \vspace{-1ex}
    \caption{{\method} is robust to corrupted server data \label{fig:noise}}
    \vspace{-1ex}
\end{figure}

\paragraph{Effect of server data (RQ3)}
We investigate how the performance of {\method} is influenced by both the quality and quantity of server data. To simulate low-quality data, we introduce four types of random noises to the server data, following the approach proposed by \citep{corruption}. As illustrated in Figure \ref{fig:noise}, {\method} exhibits remarkable consistency across various noise types, highlighting its robustness to variations in server data quality. Additionally, as demonstrated in Appendix \ref{appendix:exp:server_data}, {\method} exhibits greater resilience to label skewness in server data compared to baseline reference-based AGRs. Moreover, {\method} proves to be effective even with a minimal amount of server data, surpassing all baseline AGRs with just 5 samples per class on CIFAR-10.

\rev{
\paragraph{Effect of hyper-parameters (RQ3)}
We show in Appendix \ref{appendix:exp:hyperparameters} that {\method} is robust to a wide range of $f$ and $p_{\min}$ under multiple fractions of Byzantines $\beta = |\cB| / n$. 
}

\rev{
\paragraph{More label skewness settings (RQ3)}
In Appendix \ref{appendix:exp:noniid}, we evaluate {\method} under two more label skewness settings: step partition \citep{FedBE} and Dirichlet partition \citep{dirichlet}. We also test {\method} under different levels of non-IIDness, and different participation rates. We observe that {\method} has consistent performance across all setting. 
}

\paragraph{Beyond label skewness (RQ4)}
In a real FL system, label skewness may not be the sole kind of distribution shifts. We consider a setting with both label skewness and feature skewness on CIFAR-10, where we additionally add different types of image corruption to each client \citep{corruption}. Results in Table \ref{tab:corruption} shows that {\method} still achieves significantly higher worst-case accuracy than baseline AGRs. 

\paragraph{More FL frameworks (RQ4)}
We extend {\method} to more FL frameworks, including FedAvg \citep{FedAvg} and FedProx \citep{FedProx} in Appendix \ref{appendix:exp:more_fl}. {\method} still remains effective for these frameworks. 

\begin{table}
\vspace{-1ex}
\caption{Performance (mean (s.d.) \% over five random seeds) on CIFAR-10 with label skewness and image corruptions (see full table in Appendix \ref{appendix:exp:feat})} \label{tab:corruption}
\vspace{1ex}
\centering
\setlength{\tabcolsep}{4pt}{
 \resizebox{1.0\linewidth}{!}{
\begin{tabular}{ccccccccca}
\toprule
\multirow{2}*{Method} & \multicolumn{2}{c}{$|\cB| = 0$} & \multicolumn{6}{c}{$|\cB| = 15$ (Acc $\uparrow$)}\\
\cmidrule(lr){2-3} \cmidrule(lr){4-10}
& Acc $\uparrow$ & MRD $\downarrow$ & Gauss & IPM & LIE & Mimic & MinMax & MinSum & Wst \\
\midrule
Average & \meansd{68.7}{0.4} & - & \meansd{10.0}{0.0} & \meansd{10.0}{0.0} & \meansd{64.6}{0.7} & \meansd{67.5}{0.5} & \meansd{27.9}{4.9} & \meansd{21.6}{7.5} & 10.0 \\
MKrum   & \meansd{66.8}{1.1} & \meansd{16.7}{11.7} & \meansd{68.2}{0.7} & \meansd{52.9}{10.2} & \meansd{63.1}{1.1} & \meansd{54.9}{25.1} & \meansd{67.2}{0.4} & \meansd{62.3}{2.1} & 52.9 \\
FLTrust & \meansd{50.1}{0.9} & \meansd{29.1}{2.1} & \meansd{50.0}{1.1} & \meansd{47.8}{1.7} & \meansd{47.3}{2.5} & \meansd{49.8}{0.9} & \meansd{49.0}{1.8} & \meansd{49.1}{1.9} & 47.3 \\
\method & \meansd{66.5}{1.0} & \meansd{6.7}{2.6} & \meansd{68.5}{0.3} & \meansd{66.0}{0.7} & \meansd{62.8}{1.6} & \meansd{66.2}{0.7} & \meansd{67.7}{0.5} & \meansd{67.5}{0.6} & 62.8 \\
\bottomrule
\end{tabular}
}
}
\end{table}
\begin{table}
\vspace{-1ex}
\caption{Ablation study (Accuracy, mean (s.d.) \% over five random seeds, AG-News with $|\cH| = 16, f=2$) \label{tab:ablation}}
\vspace{1ex}
\centering
\setlength{\tabcolsep}{4pt}{
 \resizebox{1.0\linewidth}{!}{
\begin{tabular}{cccccccc}
\toprule
\multirow{2}*{Method} & \multirow{2}*{$|\cB| = 0$} & \multicolumn{6}{c}{$|\cB| = 2$}\\
\cmidrule(lr){3-8}
& & Gauss & IPM & LIE & Mimic & MinMax & MinSum \\
\midrule
Average                 & \meansd{88.3}{0.1} & \meansd{25.8}{4.1} & \meansd{25.0}{0.0} & \meansd{88.3}{0.1} & \meansd{88.1}{0.2} & \meansd{82.7}{0.3} & \meansd{84.7}{0.1}\\
\method-ES              & \meansd{88.3}{0.1} & \meansd{88.3}{0.1} & \meansd{86.3}{0.5} & \meansd{88.3}{0.1} & \meansd{88.1}{0.1} & \meansd{88.3}{0.1} & \meansd{88.2}{0.1}\\
\method                 & \meansd{88.3}{0.1} & \meansd{88.3}{0.1} & \meansd{88.1}{0.3} & \meansd{88.3}{0.1} & \meansd{88.0}{0.1} & \meansd{88.4}{0.2} & \meansd{88.3}{0.2}\\
{\method} w.o. stage 1  & \meansd{83.0}{1.1} & \meansd{82.8}{0.8} & \meansd{82.2}{1.3} & \meansd{82.8}{1.0} & \meansd{82.6}{1.1} & \meansd{82.7}{1.4} & \meansd{82.7}{1.0}\\
{\method} w.o. stage 2  & \meansd{88.3}{0.1} & \meansd{24.8}{0.4} & \meansd{25.0}{0.0} & \meansd{88.3}{0.1} & \meansd{88.0}{0.1} & \meansd{88.4}{0.2} & \meansd{88.3}{0.2}\\
\bottomrule
\end{tabular}
}
}
\end{table}

\paragraph{Ablation Study}
We study how each component of {\method} contributes to the aggregation in Table \ref{tab:ablation}. 
\textit{{\method} w.o. stage 1} skips the subspace optimization and uses the subspace initialized with server gradients. Though being robust to attacks, it fails to fully utilize clients' data, and thus has a worse performance. \textit{{\method} w.o. stage 2} averages all projected gradients without discarding Byzantine gradients. It is unbiased, but not robust to attacks. \textit{\method-ES} uses exhaustive searching instead of alternating optimization to fit the \hsubs, globally minimizing the trimmed reconstruction loss. We observe that {\method} has performance comparable to \method-ES while calling TrSVD for much fewer times ($\approx 3$ v.s. $\binom{n}{f}$), \rev{which reduces the computation time from 5.69s to only 13.6ms}. We can conclude that (1) both stages in {\method} are necessary to guarantee performance and robustness, and (2) alternating optimization significantly improves the efficiency while maintaining the performance.


\section{CONCLUSION}
\label{sec:conclusion}

This paper focuses on Byzantine-robustness in FL with label skewness. We show that existing AGRs suffer from selection bias and increased vulnerability, and propose {\method} to alleviate these problems. We verify the unbiasedness and robustness of {\method} theoretically and empirically. 

\newpage

\section*{Acknowledgments}

This work is supported by National Science Foundation under Award No. IIS-1947203, IIS-2117902, and the U.S. Department of Homeland Security under Grant Award Number, 17STQAC00001-06-00. The views and conclusions are those of the authors and should not be interpreted as representing the official policies of the funding agencies or the government.

\bibliography{main}
\section*{Checklist}

\begin{enumerate}

\item For all models and algorithms presented, check if you include:
\begin{enumerate}
\item A clear description of the mathematical setting, assumptions, algorithm, and/or model. [Yes]
\item An analysis of the properties and complexity (time, space, sample size) of any algorithm. [Yes, in Subsection \ref{subsec:complexity}]
\item (Optional) Anonymized source code, with specification of all dependencies, including external libraries. [Yes]
\end{enumerate}

\item For any theoretical claim, check if you include:
\begin{enumerate}
\item Statements of the full set of assumptions of all theoretical results. [Yes]
\item Complete proofs of all theoretical results. [Yes, in Appendix \ref{appendix:proofs}]
\item Clear explanations of any assumptions. [Yes, in Section ? and Appendix \ref{subsubsec:assumption}]     
\end{enumerate}

\item For all figures and tables that present empirical results, check if you include:
\begin{enumerate}
\item The code, data, and instructions needed to reproduce the main experimental results (either in the supplemental material or as a URL). [Yes, the code is available at \url{https://github.com/baowenxuan/BOBA}]
\item All the training details (e.g., data splits, hyperparameters, how they were chosen). [Yes, in Appendix \ref{appendix:exp:setup}]
\item A clear definition of the specific measure or statistics and error bars (e.g., with respect to the random seed after running experiments multiple times). [Yes]
\item A description of the computing infrastructure used. (e.g., type of GPUs, internal cluster, or cloud provider). [Yes, in Appendix \ref{appendix:exp:setup}]
\end{enumerate}

\item If you are using existing assets (e.g., code, data, models) or curating/releasing new assets, check if you include:
\begin{enumerate}
\item Citations of the creator If your work uses existing assets. [Yes]
\item The license information of the assets, if applicable. [Not Applicable]
\item New assets either in the supplemental material or as a URL, if applicable. [Not Applicable]
\item Information about consent from data providers/curators. [Not Applicable]
\item Discussion of sensible content if applicable, e.g., personally identifiable information or offensive content. [Not Applicable]
\end{enumerate}

\item If you used crowdsourcing or conducted research with human subjects, check if you include:
\begin{enumerate}
\item The full text of instructions given to participants and screenshots. [Not Applicable]
\item Descriptions of potential participant risks, with links to Institutional Review Board (IRB) approvals if applicable. [Not Applicable]
\item The estimated hourly wage paid to participants and the total amount spent on participant compensation. [Not Applicable]
\end{enumerate}

\end{enumerate}


\onecolumn
\appendix
\addtocontents{toc}{\protect\setcounter{tocdepth}{3}}  

\tableofcontents
\newpage
\section{DETAILED RELATED WORKS}
\label{appendix:related_works}

\paragraph{Majority-based robust aggregator with IID clients}
Majority-based aggregators assume the gradients of honest clients should cluster together, and find a vector near to the majority of the gradients with robust mean estimators, including coordinate-wise median (CooMed) / trimmed mean (TrMean) \citep{trmean}, geometric median (GeoMed) \citep{geomed,rfa}, and Krum \citep{krum}. Specifically, CooMed / TrMean use median and trimmed mean on each dimension of the gradient separately. GeoMed minimizes the sum of L2 distances between each gradient and the aggregation result. Krum computes each gradient's sum of square L2 distances to its $k$ nearest neighbors, and picks the gradients with the lowest score. All these methods are robust mean estimators with bounded gradient estimation error, and are theoretically proven not to fail arbitrarily in IID settings. However, as shown in our analysis, they suffer from selection bias and increased vulnerability in label skewness settings. The bounds for gradient estimation error still hold, but become vacuous under severe non-IIDness.

\paragraph{Reference-based robust aggregator with IID clients}
Another line of works utilize server data to evaluate each client update, and reweigh these clients to achieve better robustness. Loss-based rejections \citep{rej} evaluate client updates with their loss on server data, and drop clients whose updates are the most harmful. Specifically, we consider SelfRej, which selects $n - f$ clients whose local models $\vw_i = \vw_G - \eta \vg_i$ have the smallest loss, and AvgRej, which selects $n - f$ clients that can lower the loss of averaged model the most. Zeno \citep{zeno} generalizes this idea by considering both loss and gradient scales, believing that honest gradients should lower the model loss with small gradient norm. FLTrust \citep{fltrust} uses server data to compute a server gradient, and reweighs the client gradients with their cosine similarity to the server gradient. ByGARS \citep{bygars} optimizes the aggregation weights of client gradients with server data in a meta-learning fashion. However in each update step, it still relies on the inner product of (normalized) client gradients and server gradient. Different from the original ByGARS which saves the aggregation weights as the initialization for the next round, we use zero initialization for every round to avoid using historical information. These methods use server data to improve aggregation, however, they are not specifically designed to tackle the non-IID challenge in FL.

\paragraph{Robust aggregator with non-IID clients}
Recently, a few works have studied robustness with non-IID clients. \cite{bucket} combine IID aggregation with bucketing, using averages of random subset of client gradients as inputs of an IID aggregator, e.g., Krum. It makes the inputs of the aggregator more homogeneous. However, bucketing also increases the ratio of Byzantine gradients, which sacrifices some robustness. For example, if there are $|\cB|$ Byzantine gradients among totally $n$ gradients, after bucketing with subset size $s$, there can be as much as $|\cB|$ corrupted gradients among totally $n/s$ gradients fed to the aggregator, which increases the ratio of Byzantines from $\frac{|\cB|}{n}$ to $s \cdot \frac{|\cB|}{n}$. Similar to {\method}, RAGE \citep{rage} also uses singular value decomposition (SVD) for robust aggregation. However, it uses SVD to remove Byzantine clients iteratively, whereas our work focuses on applying SVD to model the distribution of honest clients' gradients.

\paragraph{Robust aggregator using historical information}
Some works \citep{stateful,history} assume stateful clients or use historical information to improve robustness. They mainly focus on distributed learning, where the index for both honest and Byzantine clients remains the same across communication rounds. However, such assumptions do not hold in FL, especially cross-device FL, where the training clients are different across communication rounds. Therefore, we only focus on algorithms that do not use any historical information.

\paragraph{Robust aggregator for personalized FL}
While our paper focuses on global FL, where all clients share the same global model, robustness is also studied in personalized FL. \cite{cluster} divide clients into IID groups and train global models in each group. Ditto \citep{ditto} learns personalized models to achieve fairness and robustness, but still requires training a robust global model. \cite{robustmulti} propose a Byzantine-robust multi-task learning system. 

\paragraph{Non-IIDness in FL} 
Besides robustness, non-IIDness also raises optimization challenges in FL. When clients take multiple local steps, non-IIDness makes local updates diverge and thus degrades the model. A common method to handle non-IIDness is to share a limited amount of data as augmentation \citep{nonIID}, which can be collected in many real applications. To further protect privacy, some works replace the raw samples with aggregated samples \citep{FedMix}, or synthetic samples \citep{FedDPGAN}. Compared to them, our work assumes very limited server data. 

\paragraph{Label Skewness and Mixture Distribution}
Plenty of works focus on label skewness, a particular sub-class of non-IIDness. FedAwS \citep{FedAwS} studies an extreme case where each client has only access to one class, while FedRS \citep{FedRS} focuses on a general label skewness setting. A related non-IID setting is a mixture distribution \citep{mixture}, where each client's data distribution is a mixture of several shared distributions with its own mixture weights. {\method} mainly focuses on label skewness and can be easily extended to mixture distribution. 
\newpage
\section{MISSING PROOFS}
\label{appendix:proofs}

\subsection{Convergence Analysis}

In this subsection we provides classical convergence analysis which connects convergence to the gradient estimation error. We consider two cases: 
\begin{itemize}
    \item Smooth and non-negative loss (Proposition \ref{prop:converge_smooth})
    \item Smooth and strongly convex loss (Proposition \ref{prop:converge_smooth_convex})
\end{itemize}


We start with formal definitions. 

\begin{definition}[$L$-smoothness]
    A function $f: \bbR^d \to \bbR$ is $L$-smooth if for all $\vx, \vy \in \bbR^d$, 
    \begin{align*}
        \| \nabla f(\vx) - \nabla f(\vy) \|_2 \leq L \| \vx - \vy \|_2
    \end{align*}
    equivalently, for all $\vx, \vy \in \bbR^d$, 
    \begin{align*}
        f(\vy) \leq f(\vx) + \nabla f(\vx)^\top (\vy - \vx) + \frac{L}{2} \| \vx - \vy \|_2^2
    \end{align*}
\end{definition}


\begin{definition}[$\mu$-strong convexity]
    A function $f: \bbR^d \to \bbR$ is $\mu$-strongly convex if for all $\vx, \vy \in \bbR^d$, 
    \begin{align*}
        f(\vy) \geq f(\vx) + \nabla f(\vx)^\top (\vy - \vx) + \frac{\mu}{2} \| \vx - \vy \|_2^2
    \end{align*}
\end{definition}


\subsubsection{Convergence with Smooth and Non-Negative Loss}

In Proposition \ref{prop:converge_smooth}, we provides convergence analysis with $L$-smooth and non-negative loss. 

\setcounter{mainsection}{5}
\setcounter{maintheorem}{0}

\begin{mainproposition}[Convergence with smooth non-negative loss] \label{prop:converge_smooth}
    With non-negative $L$-smooth population risk $\cL(\vw)$, conducting SGD with noisy gradient $\hat\vmu = \hat{g}(\vw)$ and step size $\eta = \frac{1}{L}$. If the gradient estimation error $\bbE \| \hat\vmu - \bbE\vmu\|_2^2 = \bbE \| \hat{g}(\vw) - \nabla \cL(\vw) \|_2^2 \leq \Delta^2$ for all $\vw$, then for any weight initialization $\vw^{(0)}$, after $T$ steps, 
    \begin{align*}
        \frac{1}{T} \sum_{t=0}^{T-1} \bbE \left\| \nabla \cL(\vw^{(t)}) \right\|_2^2 
         \leq 2\frac{L}{T} \cL(\vw^{(0)}) + \Delta^2
    \end{align*}
\end{mainproposition}

\begin{proof}
    For any $\vw^{(t)}$, 
    \begin{align*}
        \cL(\vw^{(t+1)}) 
        &\leq \cL(\vw^{(t)}) + \nabla \cL(\vw^{(t)})^\top (\vw^{(t+1)} - \vw^{(t)}) + \frac{L}{2} \left\| \vw^{(t+1)} - \vw^{(t)} \right\|_2^2 \tag{$L$-smoothness}\\
        &= \cL(\vw^{(t)}) + \nabla \cL(\vw^{(t)})^\top \left[ -\eta \left( \hat{g}(\vw^{(t)}) - \nabla \cL(\vw^{(t)}) + \nabla \cL(\vw^{(t)}) \right) \right] \\
        &\quad\ + \frac{L}{2} \left\| - \eta \left( \hat{g}(\vw^{(t)}) - \nabla \cL(\vw^{(t)}) + \nabla \cL(\vw^{(t)}) \right) \right\|_2^2 \\
        &= \cL(\vw^{(t)}) + \left( \frac{L\eta^2}{2} - \eta \right) \left\| \nabla \cL(\vw^{(t)}) \right\|_2^2 + \left( L \eta^2 - \eta \right) \nabla\cL(\vw^{(t)})^\top \left( \hat{g}(\vw^{(t)}) - \nabla \cL(\vw^{(t)}) \right) \\
        &\quad\ + \frac{L\eta^2}{2} \left\| \hat{g}(\vw^{(t)}) - \nabla \cL(\vw^{(t)})  \right\|_2^2 \\
        &= \cL(\vw^{(t)}) - \frac{1}{2L} \left\| \nabla \cL(\vw^{(t)}) \right\|_2^2 + \frac{1}{2L} \left\| \hat{g}(\vw^{(t)}) - \nabla \cL(\vw^{(t)})  \right\|_2^2 \tag{$\eta = \frac{1}{L}$}
    \end{align*}
    Equivalently, 
    \begin{align*}
        \left\| \nabla \cL(\vw^{(t)}) \right\|_2^2 \leq 2L \left( \cL(\vw^{(t)}) - \cL(\vw^{(t+1)}) \right) + \left\| \hat{g}(\vw^{(t)}) - \nabla \cL(\vw^{(t)}) \right\|_2^2
    \end{align*}
    Average over $t = 0, \cdots, T-1$, we get
    \begin{align*}
         \frac{1}{T} \sum_{t=0}^{T-1} \left\| \nabla \cL(\vw^{(t)}) \right\|_2^2 
         &\leq 2\frac{L}{T} \left( \cL(\vw^{(0)}) - \cL(\vw^{(T)}) \right) + \frac{1}{T} \sum_{t=0}^{T-1} \left\| \hat{g}(\vw^{(t)}) - \nabla \cL(\vw^{(t)}) \right\|_2^2  \\
         &\leq 2\frac{L}{T} \cL(\vw^{(0)}) + \frac{1}{T} \sum_{t=0}^{T-1} \left\| \hat{g}(\vw^{(t)}) - \nabla \cL(\vw^{(t)}) \right\|_2^2 
    \end{align*}
    Finally, take expectations at both sides
    \begin{align*}
         \frac{1}{T} \sum_{t=0}^{T-1} \bbE \left\| \nabla \cL(\vw^{(t)}) \right\|_2^2 
         \leq 2\frac{L}{T} \cL(\vw^{(0)}) + \Delta^2
    \end{align*}
\end{proof}


\subsubsection{Convergence with Smooth and Strongly Convex Loss}

\begin{lemma} \label{lemma:one_step}
    Let $f(\vw)$ be $L$-smooth and $\mu$-strongly convex, conducting GD with exact gradient $\nabla f(\vw)$ and step size $\eta = \frac{2}{L + \mu}$. For all $t$, 
    \begin{align*}
        \| \vw^{(t+1)} - \vw^* \|_2 \leq \left(\frac{L - \mu}{L + \mu}\right) \| \vw^{(t)} - \vw^* \|_2
    \end{align*}
\end{lemma}
\begin{proof}
    See Theorem 3.12 in \cite{book_convergence}.
\end{proof}

\begin{proposition}[Convergence with smooth and strongly convex loss] \label{prop:converge_smooth_convex}
    With $L$-smooth and $\mu$-strongly convex population risk $\cL(\vw)$, conducting SGD with noisy gradient $\hat\vmu = \hat{g}(\vw)$ and step size $\eta = \frac{2}{L + \mu}$. If the gradient estimation error $\bbE \| \hat\vmu - \bbE\vmu\|_2^2 = \bbE \| \hat{g}(\vw) - \nabla \cL(\vw) \|_2^2 \leq \Delta^2$ for all $\vw$, then for any weight initialization $\vw^{(0)}$, after $T$ steps, 
    \begin{align*}
        \left\| \vw^{(T)} - \vw^* \right\|_2 \leq \left(\frac{L - \mu}{L + \mu}\right)^T \left\| \vw^{(0)} - \vw^* \right\|_2 + \frac{1}{\mu} \Delta
    \end{align*}
\end{proposition}

\begin{proof}
    For any $\vw^{(t)}$, 
    \begin{align*}
        \left\| \vw^{(t+1)} - \vw^* \right\|_2
        &= \left\| \vw^{(t)} - \eta\hat{g}\left(\vw^{(t)}\right) - \vw^* \right\|_2 \\
        &= \left\| \vw^{(t)} - \eta \nabla \cL\left(\vw^{(t)}\right) - \vw^* + \eta \left( \nabla \cL\left(\vw^{(t)}\right) - \hat{g}\left(\vw^{(t)}\right) \right)\right\|_2 \\
        &\leq \left\| \vw^{(t)} - \eta \nabla \cL\left(\vw^{(t)}\right) - \vw^* \right\|_2 + \eta \left\|  \nabla \cL\left(\vw^{(t)}\right) - \hat{g}\left(\vw^{(t)}\right) \right\|_2 \\
        &\leq \left(\frac{L - \mu}{L + \mu}\right) \left\| \vw^{(t)} - \vw^* \right\|_2 + \frac{2}{L + \mu} \left\|  \nabla \cL\left(\vw^{(t)}\right) - \hat{g}\left(\vw^{(t)}\right) \right\|_2 \tag{Lemma \ref{lemma:one_step} and $\eta = \frac{2}{L + \mu}$} 
    \end{align*}
    By induction, 
    \begin{align*}
        \left\| \vw^{(T)} - \vw^* \right\|_2 
        &\leq \left(\frac{L - \mu}{L + \mu}\right)^T \left\| \vw^{(0)} - \vw^* \right\|_2 + \sum_{t=0}^{T-1} \left(\frac{L - \mu}{L + \mu}\right)^t \frac{2}{L + \mu} \left\|  \nabla \cL\left(\vw^{(t)}\right) - \hat{g}\left(\vw^{(t)}\right) \right\|_2
    \end{align*}
    Notice that for any $\vw$, 
    \begin{align*}
        \bbE  \| \hat\vmu - \bbE\vmu\|_2 = \sqrt{ \bbE  \| \hat\vmu - \bbE\vmu\|_2^2 - \text{Var}\left( \| \hat\vmu - \bbE\vmu\|_2\right) } \leq  \sqrt{ \bbE  \| \hat\vmu - \bbE\vmu\|_2^2} \leq \Delta
    \end{align*}
    Finally, take expectations at both sides. 
    \begin{align*}
        \bbE \left\| \vw^{(T)} - \vw^* \right\|_2 
        &\leq \left(\frac{L - \mu}{L + \mu}\right)^T \left\| \vw^{(0)} - \vw^* \right\|_2 + \sum_{t=0}^{T-1} \left(\frac{L - \mu}{L + \mu}\right)^t \frac{2}{L + \mu} \bbE \left\|  \nabla \cL\left(\vw^{(t)}\right) - \hat{g}\left(\vw^{(t)}\right) \right\|_2 \\
        &\leq \left(\frac{L - \mu}{L + \mu}\right)^T \left\| \vw^{(0)} - \vw^* \right\|_2 + \sum_{t=0}^{T-1} \left(\frac{L - \mu}{L + \mu}\right)^t \frac{2}{L + \mu} \Delta \\
        &\leq \left(\frac{L - \mu}{L + \mu}\right)^T \left\| \vw^{(0)} - \vw^* \right\|_2 + \sum_{t=0}^{\infty} \left(\frac{L - \mu}{L + \mu}\right)^t \frac{2}{L + \mu} \Delta  \\
        &= \left(\frac{L - \mu}{L + \mu}\right)^T \left\| \vw^{(0)} - \vw^* \right\|_2 + \frac{1}{1 - \frac{L - \mu}{L + \mu}} \frac{2}{L + \mu} \Delta  \\
        &= \left(\frac{L - \mu}{L + \mu}\right)^T \left\| \vw^{(0)} - \vw^* \right\|_2 + \frac{1}{\mu} \Delta
    \end{align*}
\end{proof}

\begin{remark}
    Some previous literature, including \cite{trmean}, use $\frac{1}{L}$ as the step size, which results in the same parameter estimation error but sub-optimal convergence rate
    \begin{align*}
        \bbE \left\| \vw^{(T)} - \vw^* \right\|_2 \leq \left(\frac{L - \mu}{L}\right)^T \left\| \vw^{(0)} - \vw^* \right\|_2 + \frac{1}{\mu} \Delta
    \end{align*}
    where $1 > \frac{L - \mu}{L}\ > \frac{L - \mu}{L + \mu}$. The proof can be found in Theorem 3.10 in \cite{book_convergence}. Instead, we choose step size $\eta = \frac{2}{L + \mu}$ which improves the convergence rate. 
\end{remark}


\newpage

\subsection{Upper Bound of Gradient Estimation Error of {\method}}
\label{appendix:proof:grad_error_boba}

In this subsection we prove bounded gradient estimation error for {\method}. Since the full proof is long, we split it to parts for clarity: 
\begin{itemize}
    \item Subsubsection \ref{subsubsec:notation} summarizes the notation used in the proof. 
    \item Subsubsection \ref{subsubsec:assumption} gives formal assumptions. 
    \item Subsubsection \ref{subsubsec:lemma} provides useful lemmas used in the proof. 
    \item Subsubsection \ref{subsubsec:trl} proves that {\method} stage 1 can converge to an affine subspace with upper bounded trimmed reconstruction loss in expectation. 
    \item Subsubsection \ref{subsubsec:stage1} proves the robustness of {\method} stage 1, i.e., the fitted subspace is closed enough to the \hsubs. 
    \item Subsubsection \ref{subsubsec:stage2} proves the robustness of {\method} stage 2, i.e., all honest gradients will not be discarded. 
    \item Subsubsection \ref{subsubsec:robust} wraps up the previous subsubsections, and proves the robustness of {\method}. 
\end{itemize}

\newpage
\subsubsection{Notation} \label{subsubsec:notation}

We summarize all notations we use in our proof in Table \ref{tab:notation}. 

\begin{table}[h!]
  \centering
  \vspace{-1ex}
  \caption{Notation}
  \vspace{1ex}
  \small
  \begin{tabular}{cl}
    \toprule
    Notation            & Description \\
    \midrule
    $d$                 & dimensionality of model parameters and gradient  \\
    $c$                 & number of classes \\
    $n$                 & number of clients \\
    $\cH$               & set of honest clients \\
    $|\cH|$             & number of honest clients \\
    $\cB$               & set of Byzantine clients \\
    $|\cB|$             & real number of Byzantine clients \\
    $f$                 & declared number of Byzantine clients. The aggregator is robust when $f \geq |\cB|$ \\
    $\vg_i$             & gradient uploaded by client $i$, $i = 1, \cdots, n$ \\
    $\vmu$              & average of all honest gradients, $\vmu = \frac{1}{|\cH|} \sum_{i \in \cH} \vg_i$. This is the gradient of empirical loss. \\
    $\bbE \vg_i$        & expectation of honest gradient $\vg_i, i \in \cH$. Note that Byzantine gradient does not have expectation. \\
    $\bbE \vmu$         & expectation of $\vmu$. This is the gradient of population loss. \\
    $\hat\vmu$          & aggregation result, $\hat\vmu = \agg(\{ \vg_i \}_{i=1}^n)$ \\
    \midrule
    $\epsilon$          & upper bound of client inner variation, formally defined in Assumption \ref{assumption:deviation} \\
    $\delta$            & upper bound of client outer variation, formally defined in Assumption \ref{assumption:deviation} \\
    $\epsilon_s$        & upper bound of server inner variation, formally defined in Assumption \ref{assumption:deviation} \\
    $\delta_s$          & upper bound of server outer variation, formally defined in Assumption \ref{assumption:deviation} \\
    $\sigma$            & lower bound of client singular value, formally defined in Assumption \ref{assumption:singular} \\
    $\sigma_s$          & lower bound of server singular value, formally defined in Assumption \ref{assumption:singular} \\
    \midrule
    $\cP$               & an affine subspace \\
    $\Pi_{\cP}$         & an affine projection function \\
    $\cF(\cP)$        & $n - f$ gradients used to fit $\cP$ (among $\{\vg_i\}_{i=1}^n$) \\
    $\cN(\cP)$        & $n - f$ nearest neighbors of $\cP$ (among $\{\vg_i\}_{i=1}^n$) \\
    $\ell_t(\cP)$         & trimmed reconstruction loss of $\cP$, $\ell_t(\cP) = \sum_{i \in \cN(\cP)} \| \vg_i - \Pi_{\cP} (\vg_i) \|_2^2$ \\
    $\cP^*$             & the {\hsubs}. It goes through the expectation of honest gradients $\{ \bbE \vg_i \}_{i \in \cH}$ \\
    $\hat\cP$           & the projection function fitted by \method \\
    $\cS$ & $n - 2f$ clients that is both honest and in $n - f$ nearest neighbors of $\hat\cP$, $\cS = \{s_1, \cdots, s_{n - 2f}\} \subset (\cH \cap \cN(\hat\cP))$ \\
    $\partial \vS$      & matrix of differences between projections to fitted and ideal affine subspaces of expected gradients in $\cS$, \\
    & $\partial \vS = [\Pi_{\hat\cP}(\bbE \vg_{s_1}) - \Pi_{\cP^*}(\bbE \vg_{s_1}), \cdots, \Pi_{\hat\cP}(\bbE \vg_{s_{n - 2f}}) - \Pi_{\cP^*}(\bbE \vg_{s_{n - 2f}})] \in \bbR^{d \times (n - 2f)}$ \\
    $\Delta \vg_i$      & difference between (fitted) projection and expectation of honest gradient $\vg_i, i \in \cH$, $\Delta \vg_i =  \Pi_{\hat\cP}(\vg_i) - \bbE \vg_i$ \\
    \midrule
    $\vgamma_z$         & server gradient of class $z$, $z = 1, \cdots, c$ \\
    $\bbE \vgamma_z$    & expectation of server gradient $\vgamma_z, z = 1, \cdots, c$ \\
    $\vGamma$           & matrix of server gradients, $\vGamma = [\vgamma_1, \cdots, \vgamma_c]$ \\
    $\bbE \vGamma$      & matrix of expectations of server gradients, $\bbE \vGamma = [\bbE \vgamma_1, \cdots, \bbE \vgamma_c]$ \\
    $\Pi_{\hat\cP}(\vGamma)$  & matrix of projections of server gradients, $\Pi_{\hat\cP}(\vGamma) = [\Pi_{\hat\cP}(\vgamma_1), \cdots, \Pi_{\hat\cP}(\vgamma_c)]$ \\
    $\Delta \vGamma$    & matrix of differences between (fitted) projection and expectation of server gradients, \\ 
                        & $\Delta\vGamma = \Pi_{\hat\cP}(\vGamma) - \bbE \vGamma = [\Pi_{\hat\cP}(\vgamma_1) - \bbE\vgamma_1, \cdots, \Pi_{\hat\cP}(\vgamma_c) - \bbE\vgamma_c]$ \\
    \midrule
    $\vp_i$             & true label distribution of honest client $i \in \cH$ \\
    $\hat{\vp}_i$       & estimated label distribution of client $i$ \\ 
    $p_{\min}$               & hyperparameter of \method, $p_{\min} \leq 0$ in our case \\
    $\hat{\vp}_\cH$     & average of all estimated label distributions of honest clients, $\hat{\vp}_\cH = \frac{1}{|\cH|} \sum_{i \in \cH} \hat{\vp}_i$ \\
    $\hat{\vp}_\cB$     & average of all estimated label distributions of Byzantine clients that evading stage 2, \\ & $\hat\vp_{\cB} = \frac{1}{|\cB|} \sum_{b \in \cB} \hat\vp_b$ when all Byzantine gradients evade stage 2 \\
    \bottomrule
  \end{tabular}
  \label{tab:notation}
\end{table}

\newpage
\subsubsection{Assumptions} \label{subsubsec:assumption}

In this part, we re-introduce the assumptions mentioned in the main text and provide more explanations in remarks. 

\setcounter{mainsection}{5}
\setcounter{maintheorem}{1}

\begin{mainassumption}[Bounded variations]\label{assumption:deviation}
    \ 
    \begin{enumerate}
        \item Honest client inner variation: 
        $
            \bbE \| \vg_i - \bbE\vg_i \|_2^2 \leq \epsilon^2, \forall i \in \cH
        $.  
        \item Honest client outer variation: 
        $
            \| \bbE\vg_i - \bbE\vmu \|_2^2 \leq \delta^2, \forall i \in \cH
        $. 
        \item Server inner variation:
        $
            \bbE \| \vgamma_z - \bbE \vgamma_z \|_2^2 \leq \epsilon_s^2, \forall z = 1, \cdots, c
        $. 
        \item Server outer variation: 
        $
            \| \bbE\vgamma_z - \bbE\vmu \|_2^2 \leq \delta_s^2, \forall z = 1, \cdots, c
        $. 
    \end{enumerate}
\end{mainassumption}

\begin{remark} \ 
    \begin{itemize}
        \item Assumption \ref{assumption:deviation}(1) and \ref{assumption:deviation}(2) are standard assumptions in both FL and Byzantine-robust FL, e.g., Assumption 6.1.1 (vi) and (vii) in \cite{survey_field}. 
        \item Since server gradients are also `honest', Assumption \ref{assumption:deviation}(3) and \ref{assumption:deviation}(4) simply rewrites Assumption \ref{assumption:deviation}(1) and \ref{assumption:deviation}(2) with updated notation. 
    \end{itemize}
\end{remark}

\begin{mainassumption}[Bounded singular values]\label{assumption:singular_apdx}
    \ 
    \begin{enumerate}
        \item Honest client singular value: conducting centralized SVD on any $n - 2f$ expectations of honest gradients, the $(c-1)$-th singular value $\sigma_{c-1} \geq \sigma > 0$. 
        \item Server singular value: conducting centralized SVD on all $c$ expectations of server gradients, the $(c-1)$-th singular value $\sigma_{c-1} \geq \sigma_s > 0$. 
    \end{enumerate}
\end{mainassumption}

\begin{remark}
    \ 
    \begin{itemize}
        \item Assumption \ref{assumption:singular_apdx}(1) is a natural extension of the standard ``$n - 2f > 0$'' assumption prevalent in IID AGRs \citep{krum,trmean,geomed}. This extension entails that, with $c$-label skewness, it is imperative for \textit{all honest components to simultaneously outweigh the Byzantine component}. To fulfill this requirement, removing any arbitrary subset of $f$ clients from the set of $n-f$ honest clients should still ensure that the remaining $n - 2f$ honest clients affinely span the \hsubs, indicated by $\sigma_{c-1} \geq \sigma, \exists \sigma > 0$. Failure to meet this condition could empower Byzantines to form a cluster to replace an honest component.

        Assumption \ref{assumption:singular_apdx}(1) also reveals that the robustness of an FL system with label skewness depends not only on $n, f$ and $c$, but also on the label distribution for each honest client. Considering $c = 2$, when $\{\bbE \vg_i\}_{i \in \cH}$ distributes uniformly on the {\hsimp} (a line segment), Assumption \ref{assumption:singular_apdx}(1) holds as long as $n - 2f > 1$ (i.e., $f < \frac{n-1}{2}$), closely resembling the IID setting. However, when half of the honest clients have only positive samples, while the other half have only negative samples, $\{\bbE \vg_i\}_{i \in \cH}$ will only be distributed at the two vertices of the \hsimp. In this case, Assumption \ref{assumption:singular_apdx}(1) only holds when $n - 2f > \frac{n - f}{2}$ (i.e., $f < \frac{n}{3}$).
        
        \item Assumption \ref{assumption:singular_apdx}(2) assumes that $c$ server gradients form the vertices of a $(c - 1)$-honest simplex, while they do not degrade, i.e., they are not on any $(c - 2)$-simplex.  We omit Assumption \ref{assumption:singular_apdx}(2) in the main text for clarity, since Assumption \ref{assumption:singular_apdx}(1) is a sufficient condition for Assumption \ref{assumption:singular_apdx}(2). 
    \end{itemize}
\end{remark}

\newpage
\subsubsection{Useful Lemmas of Affine Projection} \label{subsubsec:lemma}

In our proof, we frequently use lemmas related to affine subspace and affine projection. For clarity, we formally define these notions and summarize these lemmas. 

\begin{definition}[Affine Subspace and Affine Projection]
    $\cP$ is a $c$-dimensional affine subspace in $\bbR^d$ if there exists a column-orthogonal $\vU \in \bbR^{d \times c}$ and a bias vector $\vm \in \bbR^d$, s.t.
    \begin{align*}
        \cP = \{\vU \vlambda + \vm: \vlambda \in \bbR^c\}
    \end{align*}
    The corresponding affine projection function $\Pi_\cP$ is an affine projection function orthogonally projecting vectors to $\cP$. 
    \begin{align*}
        \Pi_{\cP}(\vw) = \vP (\vw - \vm) + \vm, \quad \forall \vw \in \bbR^d
    \end{align*}
    where $\vP = \vU \vU^\top \in \bbR^{d \times d}$ is a projection matrix whose eigenvalues have $c$ ones and $d - c$ zeros. 
\end{definition}

Then, we present useful lemmas of affine projection. 


\begin{lemma}[Nearest neighbor projection]\label{property:1}
    For any affine projection function $\Pi_{\cP}: \bbR^d \to \bbR^d$ and two vectors $\vu, \vv \in \bbR^d$, 
    \begin{align*}
        \| \Pi_{\cP}(\vu) - \vu \|_2 \leq \| \Pi_{\cP}(\vv) - \vu \|_2
    \end{align*}
\end{lemma}
\begin{proof}
    We first prove that $\Pi_{\cP}(\vv) - \Pi_{\cP}(\vu)$ and $\Pi_{\cP}(\vu) - \vu$ are orthogonal. 
    \begin{align*}
        (\Pi_{\cP}(\vv) - \Pi_{\cP}(\vu))^\top (\Pi_{\cP}(\vu) - \vu)
        &=[(\vP (\vv - \vm) + \vm) - (\vP (\vu - \vm) + \vm) ]^\top [\vP (\vu - \vm) + \vm - \vu] \\
        &= [\vP (\vv - \vu)]^\top [(\vP - \vI) (\vu - \vm)] \\
        &= (\vv - \vu)^\top [\vP^\top (\vP - \vI)](\vu - \vm) \\
        &= (\vv - \vu)^\top \boldsymbol{0} (\vu - \vm) \\
        &= 0
    \end{align*}
    With this result, 
    \begin{align*}
        \| \Pi_{\cP}(\vv) - \vu \|_2 
        &= \sqrt{ \| \Pi_{\cP}(\vv) - \Pi_{\cP}(\vu) \|_2^2 + \| \Pi_{\cP}(\vu) - \vu \|_2^2  } \\
        &\geq \| \Pi_{\cP}(\vu) - \vu \|_2
    \end{align*}
\end{proof}


\begin{lemma}[1-contraction]\label{property:2}
    For any projection function $\Pi_{\cP}:\bbR^d \to \bbR^d$ and two vectors $\vu, \vv \in \bbR^d$, 
    \begin{align*}
        \| \Pi_{\cP} (\vu) - \Pi_{\cP} (\vv) \|_2 \leq \| \vu - \vv \|_2
    \end{align*}
\end{lemma}
\begin{proof}
    \begin{align*}
        \| \Pi_{\cP} (\vu) - \Pi_{\cP} (\vv) \|_2 &= \| [\vP (\vu - \vm) + \vm ] - [\vP (\vv - \vm) + \vm] \|_2 \\
        &= \| \vP (\vu - \vv) \|_2 \\
        &\leq \| \vP \|_2 \| \vu - \vv \|_2 \\
        &\leq \| \vu - \vv \|_2
    \end{align*}
\end{proof}


\begin{lemma}[Commutativity of affine projection and affine combination]\label{property:3}
    For any projection function $\Pi_{\cP}:\bbR^d \to \bbR^d$, a set of $n$ vectors $\{\vu_i\}_{i=1}^n \in \bbR^d$ and coefficients $\{\lambda_i\}_{i=1}^n$ subject to $\sum_{i=1}^n \lambda_i = 1$, 
    \begin{align*}
        \Pi_{\cP}\left( \sum_{i = 1}^n \lambda_i \vu_i \right) = \sum_{i = 1}^n \lambda_i \Pi_{\cP} (\vu_i)
    \end{align*}
\end{lemma}
\begin{proof}
    \begin{align*}
        \Pi_{\cP}\left( \sum_{i = 1}^n \lambda_i \vu_i \right) 
        &= \vP \left(\sum_{i = 1}^n \lambda_i \vu_i - \vm \right) + \vm \\
        &= \sum_{i = 1}^n \lambda_i [\vP (\vu_i - \vm) + \vm ] \\
        &= \sum_{i = 1}^n \lambda_i \Pi_{\cP} (\vu_i)
    \end{align*}
\end{proof}
\begin{remark}
    The sum-to-one constraint on coefficients is crucial as the projection is affine, not linear. 
\end{remark}


\begin{lemma}\label{property:4}
    For any two projection functions $\Pi_{\cP_1}, \Pi_{\cP_2}:\bbR^d \to \bbR^d$, a set of $n$ vectors $\{\vu_i\}_{i=1}^n \in \bbR^d$, coefficients $\{\lambda_i\}_{i=1}^n$ subject to $\sum_{i=1}^n \lambda_i = 1$ and an affine combination $\vv = \sum_{i=1}^n \lambda_i \vu_i$, 
    \begin{align*}
        \| \partial \vv \|_2 \leq \| \partial \vU \|_2 \cdot \| \vlambda \|_2
    \end{align*}
    where $\vlambda = [\lambda_1, \cdots, \lambda_n]^\top \in \bbR^n$, $\partial \vv = \Pi_{\cP_1}(\vv) - \Pi_{\cP_2}(\vv) \in \bbR^{d}$ and $\partial\vU = [\Pi_{\cP_1}(\vu_1) - \Pi_{\cP_2}(\vu_1), \cdots, \Pi_{\cP_1}(\vu_n) - \Pi_{\cP_2}(\vu_n)] \in \bbR^{d \times n}$.
\end{lemma}
\begin{proof}
    \begin{align*}
        \| \partial \vv \|_2 &= \| \Pi_{\cP_1}(\vv) - \Pi_{\cP_2}(\vv) \|_2 \\
        &= \left\| \Pi_{\cP_1}\left( \sum_{i=1}^n \lambda_i \vu_i \right) - \Pi_{\cP_2}\left( \sum_{i=1}^n \lambda_i \vu_i \right) \right\|_2 \\
        &= \left\| \sum_{i=1}^n \lambda_i \Pi_{\cP_1}\left( \vu_i \right) -  \sum_{i=1}^n \lambda_i \Pi_{\cP_2}\left(\vu_i \right) \right\|_2 \tag{Lemma \ref{property:3}}\\
        &= \left\| \sum_{i=1}^n \lambda_i [\Pi_{\cP_1}\left( \vu_i \right) - \Pi_{\cP_2}\left(\vu_i \right)] \right\|_2 \\
        &= \| \partial \vU \vlambda \|_2 \\
        &\leq \| \partial\vU \|_2 \cdot \| \vlambda \|_2
    \end{align*}
\end{proof}
\begin{remark}
     This lemma shows how to bound the projection error of another vector based on the ``basis'' vectors. (Strictly speaking, $\{\vu_i\}_{i=1}^n$ are not basis, as they can be dependent. In this case, we can find a $\vlambda$ with the smallest norm to get the tightest bound of $\| \Pi_{\cP_1}(\vv) - \Pi_{\cP_2}(\vv) \|_2$. )
\end{remark}

\newpage
\subsubsection{Bounded Trimmed Reconstruction Loss} \label{subsubsec:trl}

{\method} stage 1 minimizes the \textit{trimmed reconstruction loss} among clients' gradients. In this subsubsection, we prove that for both {\method}-ES (which use exhaustive searching) and {\method} (which use more efficient alternating optimization) can fit an affine subspace with upper bounded trimmed reconstruction loss in expectation, while this loss is not affected by outer deviation. We first formally define trimmed reconstruction loss in Definition \ref{dfn:trl}, and then derive upper bounds of the trimmed reconstruction losses for {\method}-ES and {\method} in Lemma \ref{lemma:loss_es} and \ref{lemma:loss_ao}, respectfully. Finally, we empirically compare their trimmed reconstruction loss. 

\begin{definition}[Trimmed reconstruction loss] \label{dfn:trl}
    Given gradients $\vg_1, \cdots, \vg_n$ and Byzantine tolerance $f$, the trimmed reconstruction loss of an affine subspace $\cP$ is
    \begin{align*}
        \ell_t(\cP) = \min_{\tiny{\substack{\vr \in \{0,1\}^n \\ \sum_{i=1}^n r_i = n - f}}} \sum_{i=1}^n r_i \left \| \vg_i - \Pi_{\cP}(\vg_i) \right\|_2^2
    \end{align*}
    which is the sum of squared distance from $\cP$ to its $n-f$ nearest neighbors. 
\end{definition}


\begin{lemma}[Trimmed reconstruction loss of {\method}-ES] \label{lemma:loss_es}
    Let $\hat\cP$ denote the subspace fitted by {\method}-ES (exhaustive searching) stage 1 and $\ell_t(\hat\cP)$ be its corresponding trimmed reconstruction loss. We have
    \begin{align*}
        \ell_t(\hat\cP) \leq \frac{n - f}{|\cH|} \sum_{i \in \cH} \| \vg_i - \bbE \vg_i \|_2^2
    \end{align*}
    Meanwhile, if we take expectation at both sides
    \begin{align*}
        \bbE \ell_t(\hat\cP) \leq (n - f) \epsilon^2
    \end{align*}
\end{lemma}

\begin{proof}
    {\method}-ES iterates through all subsets of gradients with cardinality $n - f$, and pick the subset with we it fits an affine subspace with smallest trimmed reconstruction loss. For any affine subspace $\cP$ fitted by $n-f$ gradients, denote $\cF(\cP)$ as the $n-f$ gradients with which $\cP$ is fitted, and $\cN(\cP)$ as the $n-f$ nearest neighbors of $\cP$. Also, denote $\cP^*$ as the honest subspace. For any $\cP'$ fitted by $n-f$ \textit{honest} gradients denoted as $\cF(\cP')$ (notice that $n - f \leq |\cH|$), 
    \begin{align*}
        \ell_t(\hat\cP) &\leq \ell_t(\cP') \tag{Optimality of {\method}-ES} \\
        &= \sum_{i \in \cN(\cP')} \left\| \Pi_{\cP'}(\vg_i) - \vg_i \right\|_2^2 \\
        &\leq \sum_{i \in \cF(\cP')} \left\| \Pi_{\cP'}(\vg_i) - \vg_i \right\|_2^2 \\
        &\leq \sum_{i \in \cF(\cP')} \left\| \Pi_{\cP^*}(\vg_i) - \vg_i \right\|_2^2 \tag{Optimality of SVD} \\
        &\leq \sum_{i \in \cF(\cP')} \left\| \Pi_{\cP^*}(\bbE \vg_i) - \vg_i \right\|_2^2 \tag{Lemma \ref{property:1}} \\
        &= \sum_{i \in \cF(\cP')} \left\| \bbE \vg_i - \vg_i \right\|_2^2 
    \end{align*}
    Finally, we iterate all subset of honest gradients with cardinality $n - f$. There will be $\binom{|\cH|}{n-f}$ subsets, while each honest gradient is chosen for $\binom{|\cH| - 1}{n-f - 1}$ times. Therefore, 
    \begin{align*}
        \binom{|\cH|}{n-f} \ell_t(\hat\cP) \leq \binom{|\cH| - 1}{n-f - 1} \sum_{i \in \cH} \left\| \bbE \vg_i - \vg_i \right\|_2^2 
        \quad \Rightarrow \quad \ell_t(\hat\cP) \leq \frac{n - f}{|\cH|} \sum_{i \in \cH} \left\| \bbE \vg_i - \vg_i \right\|_2^2 
    \end{align*}
\end{proof}


\newpage
\begin{lemma}[Trimmed reconstruction loss of {\method}] \label{lemma:loss_ao}
    Let $\hat\cP$ denote the subspace fitted by {\method} (alternating optimization) stage 1 and $\ell_t(\hat\cP)$ be its corresponding trimmed reconstruction loss. We have
    \begin{align*}
        \ell_t(\hat\cP) \leq 2 \frac{n-f}{|\cH|} \left( \sum_{i \in \cH} \| \vg_i - \bbE \vg_i \|_2^2 + \sum_{i \in \cH} \sum_{z=1}^c p_{iz} \| \bbE \vgamma_z - \vgamma_z  \|_2^2 \right)
    \end{align*}
    Meanwhile, if we take expectation at both sides
    \begin{align*}
        \bbE \ell_t(\hat\cP) \leq 2 (n - f) (\epsilon^2 + \epsilon_s^2)
    \end{align*}
\end{lemma}

\begin{proof}
    We denote $\hat\cP_0$ as the affine subspace initialized by server gradients $\vgamma_1, \cdots, \vgamma_c$. Since the trimmed reconstruction loss is monotone non-increasing during the alternating optimization, we have
    \begin{align*}
        \ell_t(\hat\cP) \leq \ell_t(\hat\cP_0)
    \end{align*}
    For each honest client $i \in \cH$, its expected gradient can be expressed as a \textit{convex} combination of expected server gradients, i.e., 
    \begin{align*}
        \bbE \vg_i = \sum_{z=1}^c p_{iz} \bbE \vgamma_z \tag{Proposition 3.3}
    \end{align*}
    where $[p_{i1}, \cdots, p_{iz}]^\top$ is the label distribution of client $i$. We have
    \begin{align*}
        \left\| \bbE \vg_i - \Pi_{\hat\cP_0}(\bbE \vg_i) \right\|_2^2 
        &= \left\| \Pi_{\cP^*}(\bbE\vg_i) - \Pi_{\hat\cP_0} (\bbE\vg_i) \right\|_2^2 \\
        &= \left\| \Pi_{\cP^*}\left( \sum_{z=1}^c p_{iz} \bbE \vgamma_z \right) - \Pi_{\hat\cP_0} \left(\sum_{z=1}^c p_{iz} \bbE \vgamma_z \right) \right\|_2^2 \\
        &= \left\| \sum_{z=1}^c p_{iz} \left( \Pi_{\cP^*}\left( \bbE \vgamma_z \right) - \Pi_{\hat\cP_0} \left(\bbE \vgamma_z \right) \right) \right\|_2^2 \tag{Lemma \ref{property:3}} \\
        &\leq \sum_{z=1}^c p_{iz} \| \Pi_{\cP^*}\left( \bbE \vgamma_z \right) - \Pi_{\hat\cP_0} \left(\bbE \vgamma_z \right) \|_2^2 \tag{Convexity of $\| \vx \|_2^2$}\\
        &= \sum_{z=1}^c p_{iz} \| \bbE \vgamma_z - \Pi_{\hat\cP_0} \left(\bbE \vgamma_z \right) \|_2^2 \\
        &\leq \sum_{z=1}^c p_{iz} \| \bbE \vgamma_z - \Pi_{\hat\cP_0} \left(\vgamma_z \right) \|_2^2 \tag{Lemma \ref{property:1}}\\
        &= \sum_{z=1}^c p_{iz} \| \bbE \vgamma_z - \vgamma_z  \|_2^2
    \end{align*}
    Therefore, 
    \begin{align*}
        \| \vg_i - \Pi_{\hat\cP_0} (\vg_i) \|_2^2 
        &\leq \| \vg_i - \Pi_{\hat\cP_0} (\bbE \vg_i) \|_2^2 \tag{Lemma \ref{property:2}} \\
        &= \| (\vg_i - \bbE \vg_i) + (\bbE \vg_i - \Pi_{\hat\cP_0} (\bbE \vg_i)) \|_2^2 \\
        &\leq 2 \| \vg_i - \bbE \vg_i \|_2^2 + 2 \| \bbE \vg_i - \Pi_{\hat\cP_0} (\bbE \vg_i) \|_2^2 
    \end{align*}
    Finally, denote $\cN(\hat\cP_0)$ as the $n-f$ nearest neighbors of $\hat\cP_0$
    \begin{align*}
        \ell_t(\hat\cP) &\leq \ell_t(\hat\cP_0) \\
        &= \sum_{i \in \cN(\hat\cP_0)} \| \vg_i - \Pi_{\hat\cP_0} (\vg_i) \|_2^2 \\
        &\leq \frac{n-f}{|\cH|}\sum_{i \in \cH} \| \vg_i - \Pi_{\hat\cP_0} (\vg_i) \|_2^2 \\
        &\leq 2 \frac{n-f}{|\cH|} \left( \sum_{i \in \cH} \| \vg_i - \bbE \vg_i \|_2^2 + \sum_{i \in \cH} \sum_{z=1}^c p_{iz} \| \bbE \vgamma_z - \vgamma_z  \|_2^2 \right)
    \end{align*}
\end{proof}

\begin{remark}
    When $\epsilon_s = \cO(\epsilon)$, $\bbE \ell_t(\hat\cP) = \cO(n \epsilon^2)$ for both {\method}-ES and {\method}. 
\end{remark}


\paragraph{Empirical comparison}

In Lemma \ref{lemma:loss_es} and \ref{lemma:loss_ao}, we derive upper bounds of the trimmed reconstruction loss, and show that they have the same order. Additionally, we empirically compare the trimmed reconstruction loss of two algorithms. Specifically, for each type of attack, we employed both {\method} and {\method}-ES in every round and recorded their respective trimmed reconstruction losses. Since {\method} always yield trimmed reconstruction losses greater than or equal to those of {\method}-ES, we plotted the ratio of their losses, i.e. $\frac{\ell_t(\text{BOBA})}{\ell_t(\text{BOBA-ES})}$. 

\begin{figure*}[h!]
    \centering
    \includegraphics[width=0.9\linewidth]{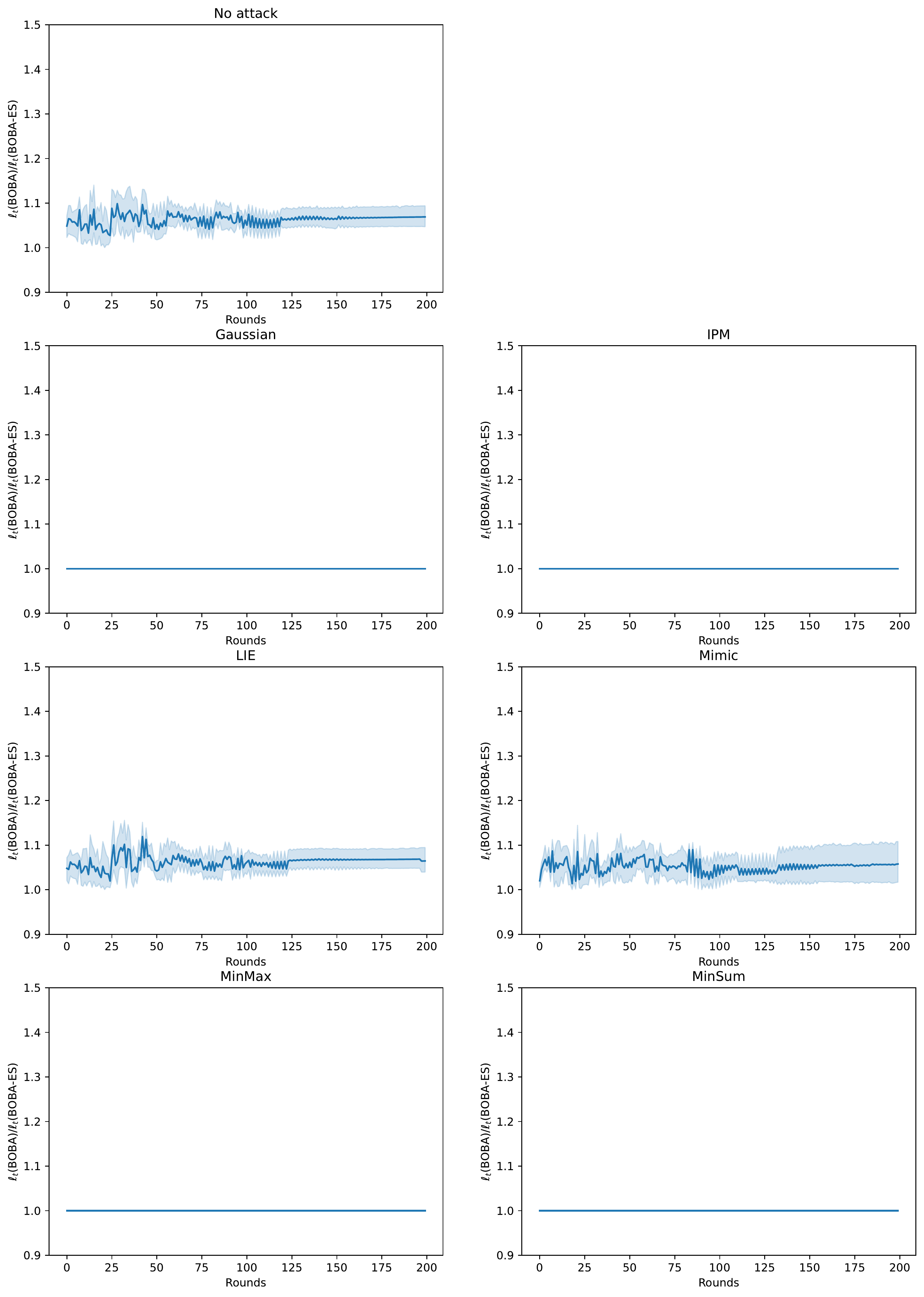}
    \caption{Comparison of trimmed reconstruction loss of {\method} and {\method}-ES}
    \label{fig:loss_compare}
\end{figure*}

We have organized the results in Figure \ref{fig:loss_compare}. It is noteworthy that when facing Gauss/IPM/MinMax/MinSum attacks, {\method} can achieve the same trimmed reconstruction loss as {\method}-ES. However, when not subjected to attacks or when facing LIE/Mimic attacks, due to the existence of multiple subspaces that can yield similar trimmed reconstruction loss, {\method} converges to a slightly higher trimmed reconstruction loss compared to {\method}-ES. Nevertheless, their convergence results are highly similar, with the loss ratio seldom exceeding 1.2. This demonstrates that {\method} can yield very similar results to {\method}-ES. It is important to note that in the main text, we also provide a comprehensive comparison of the performance of both algorithms.

\newpage
\subsubsection{Robustness of {\method} Stage 1} \label{subsubsec:stage1}

In {\method} stage 1, we fit an affine subspace close to all honest gradients, and project all gradient into this fitted subspace. In this subsubsection, we prove the robustness of stage 1, i.e., honest gradients will only be slightly perturbed in stage 1. Specifically
\begin{itemize}
    \item Lemma \ref{lemma:stage_1} proves that each honest gradient's projection is close enough to its expectation. 
    \item Lemma \ref{lemma:stage_1_avg} proves that the average of honest gradients' projections is close enough to the expectation of the average of honest gradients. 
    \item Lemma \ref{lemma:stage_1_server} proves that each server gradient's projection is close enough to its expectation. 
\end{itemize}

\begin{lemma}[Robustness of stage 1] \label{lemma:stage_1}
    Let $\hat\cP$ denote the subspace fitted by BOBA stage 1 and $\ell_t(\hat\cP)$ be its corresponding trimmed reconstruction loss. 
    For any honest gradient $\vg_h$, $\forall h \in \cH$, we have
    \begin{align*}
        \left\| \Pi_{\hat{\cP}}(\vg_h) - \bbE \vg_h \right\|_2^2 \leq 2\| \vg_h - \bbE \vg_h \|_2^2 + 4 \left( \frac{1}{n - 2f} + \frac{4 \delta^2}{\sigma^2} \right) \left( \ell_t(\hat\cP) + \sum_{i \in \cH} \left\| \vg_i - \bbE \vg_i \right\|_2^2 \right)
    \end{align*}
    Meanwhile, if we take expectation at both sides, 
    \begin{align*}
        \bbE \left\| \Pi_{\hat{\cP}}(\vg_h) - \vg_h \right\|_2^2 
        &\leq 2 \epsilon^2 + 4 \left( \frac{1}{n-2f} + \frac{4\delta^2}{\sigma^2}\right) \left( \bbE \ell_t(\hat\cP) + |\cH| \epsilon^2 \right) \\
        &\leq \begin{cases}
            \left( 2 + 4 \left( \frac{1}{n-2f} + \frac{4\delta^2}{\sigma^2}\right) (n - f + |\cH|) \right) \epsilon^2 & \textnormal{(BOBA-ES)} \\
            \left( 2 + 4 \left( \frac{1}{n-2f} + \frac{4\delta^2}{\sigma^2}\right) (2(n - f) + |\cH|) \right) \epsilon^2 + 8 \left( \frac{1}{n-2f} + \frac{4\delta^2}{\sigma^2}\right) (n - f) \epsilon_s^2 & \textnormal{(BOBA)} 
        \end{cases}
    \end{align*}
\end{lemma}

\begin{proof}
    The core of the proof is Lemma \ref{property:4}. We split the full proof into four steps. 
    \begin{itemize}
        \item Step 1: Find $n-2f$ expected gradients $\bbE\vg_{s_1}, \bbE\vg_{s_2}, \cdots, \bbE\vg_{s_{n - 2f}}$ that affinely span the {\hsubs}. Their projections to the fitted subspace and the honest subspace are close. 
        \item Step 2: Express $\bbE\vg_h$ as a affine combination of $\bbE\vg_{s_1}, \bbE\vg_{s_2}, \cdots, \bbE\vg_{s_{n - 2f}}$ with coefficient $\vlambda$. Derive an upper bound for $\| \vlambda \|_2$. 
        \item Step 3: Use Lemma \ref{property:4} to show that $\bbE\vg_h$'s projections to the fitted subspace and the honest subspace are close. 
        \item Step 4: Use triangle inequality to postprocess the inequality. 
    \end{itemize}

    \textbf{Step 1. }
    Let $\cN(\hat\cP)$ denote the $n-f$ nearest neighbors of $\hat\cP$. Among these $n-f$ gradients, at least $n - 2f$ gradients are honest. We use $\vg_{s_1}, \vg_{s_2}, \cdots, \vg_{s_{n - 2f}}$ to denote these $n - 2f$ honest gradients and $\cS = \{s_1, s_2, \cdots, s_{n - 2f}\}$ denote their indices. Since $\cS \subset \cN(\hat\cP)$, we have
    \begin{align*}
        \sum_{i \in \cS} \left\| \Pi_{\hat\cP}(\vg_i) - \vg_i \right\|_2^2 \leq \sum_{i \in \cN(\hat\cP)} \left\| \Pi_{\hat\cP}(\vg_i) - \vg_i \right\|_2^2 = \ell_t(\hat\cP)
    \end{align*}
    Let $\cP^*$ denote the {\hsubs}, i.e., the subspace on which expected gradients $\{\bbE \vg_i\}_{i \in \cH}$ lie. Then, 
    \begin{align*}
        \sum_{i \in \cS} \left\| \Pi_{\hat\cP}(\bbE \vg_i) - \Pi_{\cP^*}(\bbE \vg_i) \right\|_2^2 
        &= \sum_{i \in \cS} \left\| \Pi_{\hat\cP}(\bbE \vg_i) - \bbE \vg_i \right\|_2^2 \\
        &\leq \sum_{i \in \cS} \left\| \Pi_{\hat\cP}(\vg_i) - \bbE \vg_i \right\|_2^2 \tag{Lemma \ref{property:2}} \\
        &\leq \sum_{i \in \cS} \left( \left\| \Pi_{\hat\cP}(\vg_i) - \vg_i \right\|_2 + \left\| \vg_i - \bbE \vg_i \right\|_2 \right)^2 \\
        &\leq \sum_{i \in \cS} \left( 2 \left\| \Pi_{\hat\cP}(\vg_i) - \vg_i \right\|_2^2 + 2 \left\| \vg_i - \bbE \vg_i \right\|_2^2 \right) \\
        &\leq 2 \sum_{i \in \cS} \left\| \Pi_{\hat\cP}(\vg_i) - \vg_i \right\|_2^2 + 2 \sum_{i \in \cH} \left\| \vg_i - \bbE \vg_i \right\|_2^2 \\
        &= 2 \ell_t(\hat\cP) + 2 \sum_{i \in \cH} \left\| \vg_i - \bbE \vg_i \right\|_2^2 
    \end{align*}
    Define $\partial \vS = [\Pi_{\hat\cP}(\bbE \vg_{s_1}) - \Pi_{\cP^*}(\bbE \vg_{s_1}), \cdots, \Pi_{\hat\cP}(\bbE \vg_{s_{n - 2f}}) - \Pi_{\cP^*}(\bbE \vg_{s_{n - 2f}})] \in \bbR^{d \times (n - 2f)}$, we have
    \begin{align*}
        \| \partial \vS \|_2^2 
        \leq \| \partial \vS \|_F^2 
        = \sum_{i \in \cS} \left\| \Pi_{\hat\cP}(\bbE \vg_i) - \Pi_{\cP^*}(\bbE \vg_i) \right\|_2^2 
        \leq 2 \ell_t(\hat\cP) + 2 \sum_{i \in \cH} \left\| \vg_i - \bbE \vg_i \right\|_2^2 
    \end{align*}

    \textbf{Step 2. }
    By Assumption \ref{assumption:deviation}(1), $\bbE\vg_{s_1}, \bbE\vg_{s_2}, \cdots, \bbE\vg_{s_{n - 2f}}$ can affinely span the {\hsubs}. Therefore we can express $\bbE \vg_h$ as an affine combination of them, i.e., there exists $\vlambda = [\lambda_1, \cdots, \lambda_{n - 2f}]^\top \in \bbR^{n - 2f}$, s.t., 
    \begin{align*}
        \bbE \vg_h = \sum_{i=1}^{n - 2f} \lambda_i \bbE \vg_{s_i}, \quad \sum_{i=1}^{n - 2f} \lambda_i = 1
    \end{align*}
    With bounded singular values (Assumption \ref{assumption:singular}(1)), we can find a $\vlambda$ with small bounded norm. Although the solution of $\vlambda$ is usually not unique (when $n - 2f > c$), we only need one solution with small norm. We first ``centralize'' gradients. Let $\bbE\vm_s = \frac{1}{n - 2f} \sum_{i = 1}^{n - 2f} \bbE\vg_{s_i}$, we need to solve the following centralized linear system
    \begin{align*}
        \bbE \vg_h - \bbE \vm_s = \sum_{i=1}^{n - 2f} \theta_i ( \bbE \vg_{s_i} - \bbE \vm_s) = \vA_s \vtheta
    \end{align*}
    where $\vA_s = [\bbE \vg_{s_1} - \bbE \vm_s, \cdots, \bbE \vg_{s_{n-2f}} - \bbE \vm_s] \in \bbR^{d \times (n-2f)}$. 
    One solution of $\vtheta$ is $\theta = \vA_s^+ (\bbE \vg_h - \bbE \vm_s)$, where $\vA_s^+$ is the Moore-Penrose inverse of $\vA_s$. The norm of this solution is bounded by
    \begin{align*}
        \| \vtheta \|_2 &= \| \vA_s^+ (\bbE \vg_h - \bbE \vm_s) \|_2 \\
        &\leq \| \vA_s^+ \|_2 \cdot \| \bbE \vg_h - \bbE \vm_s \|_2 \\
        &\leq \frac{1}{\sigma} \cdot \| \bbE \vg_h - \bbE \vm_s \|_2 \tag{Assumption \ref{assumption:singular}(1)} \\
        &\leq \frac{1}{\sigma}  \cdot \left\| (\bbE \vg_h - \bbE \vmu) - \frac{1}{n - 2f} \sum_{i=1}^{n - 2f} (\bbE \vg_{s_i} - \bbE \vmu) \right\|_2 \\
        &\leq \frac{1}{\sigma}  \cdot \left( \| \bbE \vg_h - \bbE \vmu \|_2 + \frac{1}{n - 2f} \sum_{i=1}^{n - 2f} \| \bbE \vg_{s_i} - \bbE \vmu \|_2  \right) \\
        &\leq \frac{2\delta}{\sigma} \tag{Assumption \ref{assumption:deviation}(2)}
    \end{align*}
    Each solution of the centralized linear system $\vtheta$ corresponds to a solution of the original linear system
    \begin{align*}
        \vlambda = \frac{1}{n-2f} \boldsymbol{1} + \left( \vI - \frac{1}{n-2f} \boldsymbol{1}\boldsymbol{1}^\top \right) \vtheta
    \end{align*}
    Therefore, 
    \begin{align*}
        \| \vlambda \|_2 &= \left \| \frac{1}{n - 2f} \boldsymbol{1} + \left( \vI - \frac{1}{n - 2f} \boldsymbol{1} \boldsymbol{1}^\top \right) \vtheta \right \|_2 \\
        &= \sqrt{\left\| \frac{1}{n - 2f} \boldsymbol{1} \right\|_2^2 + \left\|\left( \vI - \frac{1}{n - 2f} \boldsymbol{1} \boldsymbol{1}^\top \right) \vtheta \right\|_2^2 } \tag{Orthogonality} \\
        &\leq \sqrt{\left\| \frac{1}{n - 2f} \boldsymbol{1} \right\|_2^2 + \left\|\vI - \frac{1}{n - 2f} \boldsymbol{1} \boldsymbol{1}^\top \right\|_2^2 \cdot \left \| \vtheta \right\|_2^2 } \\
        &\leq \sqrt{\left\| \frac{1}{n - 2f} \boldsymbol{1} \right\|_2^2 + \left \| \vtheta \right\|_2^2 } \\
        &\leq \sqrt{ \frac{1}{n - 2f} + \frac{4 \delta^2}{\sigma^2} }
    \end{align*}
    It is also easy to verify that $\boldsymbol{1}^\top \vlambda = 1$. 

    \textbf{Step 3. }
    Since then, we construct a $\vlambda$ satisfying the condition of Lemma \ref{property:4}. Therefore
    \begin{align*}
        \left\| \Pi_{\hat\cP}(\bbE \vg_h) - \Pi_{\cP^*}(\bbE \vg_h) \right\|_2^2 &\leq \| \partial \vS \|_2^2 \cdot \|\vlambda \|_2^2 \tag{Lemma \ref{property:4}} \\
        &\leq \left( \frac{1}{n - 2f} + \frac{4 \delta^2}{\sigma^2} \right) \left( 2 \ell_t(\hat\cP) + 2 \sum_{i \in \cH} \left\| \vg_i - \bbE \vg_i \right\|_2^2  \right)
    \end{align*}

    \textbf{Step 4. }
    Finally, 
    \begin{align*}
        \| \Pi_{\hat\cP}(\vg_h) - \bbE \vg_h \|_2^2 
        &= \| \Pi_{\hat\cP}(\vg_h) - \Pi_{\cP^*}(\bbE \vg_h) \|_2^2 \\
        &\leq 2 \| \Pi_{\hat\cP}(\vg_h) - \Pi_{\hat\cP}(\bbE \vg_h)\|_2^2 + 2 \| \Pi_{\hat\cP}(\bbE \vg_h) - \Pi_{\cP^*}(\bbE \vg_h) \|_2^2 \\
        &\leq 2\| \vg_h - \bbE \vg_h \|_2^2 + 2\| \Pi_{\hat\cP}(\bbE \vg_h) - \Pi_{\cP^*}(\bbE \vg_h) \|_2^2 \tag{Lemma \ref{property:2}}\\
        &\leq 2\| \vg_h - \bbE \vg_h \|_2^2 + 2 \left( \frac{1}{n - 2f} + \frac{4 \delta^2}{\sigma^2} \right) \left( 2 \ell_t(\hat\cP) + 2 \sum_{i \in \cH} \left\| \vg_i - \bbE \vg_i \right\|_2^2  \right) \\
        &= 2\| \vg_h - \bbE \vg_h \|_2^2 + 4 \left( \frac{1}{n - 2f} + \frac{4 \delta^2}{\sigma^2} \right) \left( \ell_t(\hat\cP) + \sum_{i \in \cH} \left\| \vg_i - \bbE \vg_i \right\|_2^2  \right) \\
    \end{align*}
\end{proof}


Lemma \ref{lemma:stage_1} shows that for each honest gradient, its projection is close to its expectation. Next, we demonstrate in Lemma \ref{lemma:stage_1_avg} that a similar property holds for the average of honest gradients. 

\begin{lemma} \label{lemma:stage_1_avg}
    Let $\hat\cP$ denote the subspace fitted by BOBA stage 1 and $\ell_t(\hat\cP)$ be its corresponding trimmed reconstruction loss. 
    Denote $\vmu = \frac{1}{|\cH|} \sum_{i \in \cH} \vg_i$ and $\hat\vmu_{\cH} = \frac{1}{|\cH|} \sum_{i \in \cH} \Pi_{\hat\cP} (\vg_i)$, we have
    \begin{align*}
        \left\| \hat\vmu_{\cH} - \bbE \vmu \right\|_2^2 \leq 2 \frac{1}{|\cH|} \sum_{i \in \cH} \left\| \vg_i - \bbE \vg_i \right\|_2^2 + 4 \left( \frac{1}{n - 2f} + \frac{\delta^2}{\sigma^2} \right) \left(\ell_t(\hat\cP) + \sum_{i \in \cH} \left\| \vg_i - \bbE \vg_i \right\|_2^2  \right)
    \end{align*}
    Meanwhile, if we take expectation at both sides, 
    \begin{align*}
        \bbE \left\| \hat\vmu_{\cH} - \bbE \vmu \right\|_2^2 
        &\leq 2 \epsilon^2 + 4 \left( \frac{1}{n-2f} + \frac{\delta^2}{\sigma^2}\right) \left( \bbE \ell_t(\hat\cP) + |\cH| \epsilon^2 \right) \\
        &\leq \begin{cases}
            \left( 2 + 4 \left( \frac{1}{n-2f} + \frac{\delta^2}{\sigma^2}\right) (n - f + |\cH|) \right) \epsilon^2 & \textnormal{(BOBA-ES)} \\
            \left( 2 + 4 \left( \frac{1}{n-2f} + \frac{\delta^2}{\sigma^2}\right) (2(n - f) + |\cH|) \right) \epsilon^2 + 8 \left( \frac{1}{n-2f} + \frac{\delta^2}{\sigma^2}\right) (n - f) \epsilon_s^2 & \textnormal{(BOBA)} 
        \end{cases}
    \end{align*}
\end{lemma}
\begin{proof}
    The average of honest gradients can also be seen as a honest gradient. Therefore, we can use the same proof framework as Lemma \ref{lemma:stage_1}, while providing tighter bound. 
    
    \textbf{Step 1. } Omitted, identical to the step 1 in the proof of \ref{lemma:stage_1}. 

    \textbf{Step 2. } Similar to step 2 in the proof of \ref{lemma:stage_1}, while we can derive a tighter bound of $\| \vlambda \|_2$
    \begin{align*}
        \| \vtheta \|_2 &= \| \vA_s^+ (\bbE \vmu - \bbE \vm_s) \|_2 \\
        &\leq \| \vA_s^+ \|_2 \cdot \| \bbE \vmu - \bbE \vm_s \|_2 \\
        &\leq \frac{1}{\sigma} \cdot \| \bbE \vmu - \bbE \vm_s \|_2 \tag{Assumption \ref{assumption:singular}(1)} \\
        &\leq \frac{1}{\sigma}  \cdot \left\| - \frac{1}{n - 2f} \sum_{i=1}^{n - 2f} (\bbE \vg_{s_i} - \bbE \vmu) \right\|_2 \\
        &\leq \frac{1}{\sigma}  \cdot \frac{1}{n - 2f} \sum_{i=1}^{n - 2f} \| \bbE \vg_{s_i} - \bbE \vmu \|_2  \\
        &\leq \frac{\delta}{\sigma} \tag{Assumption \ref{assumption:deviation}(2)}
    \end{align*}
    And therefore $\| \vlambda \|_2 \leq \sqrt{\frac{1}{n - 2f} + \frac{\delta^2}{\sigma^2}}$. 
    
    \textbf{Step 3. } Also identical to the step 3 in the proof of \ref{lemma:stage_1}. 
    \begin{align*}
        \| \Pi_{\hat\cP}(\bbE \vmu) - \Pi_{\cP^*}(\bbE \vmu) \|_2^2 \leq  \left( \frac{1}{n - 2f} + \frac{\delta^2}{\sigma^2} \right) \left( 2 \ell_t(\hat\cP) + 2 \sum_{i \in \cH} \left\| \vg_i - \bbE \vg_i \right\|_2^2  \right)
    \end{align*}

    \textbf{Step 4. } Finally, 
    \begin{align*}
        \| \hat\vmu_{\cH} - \bbE \vmu \|_2^2 
        &= \left\| \frac{1}{|\cH|} \sum_{i \in \cH} \Pi_{\hat\cP} (\vg_i) - \bbE \vmu \right\|_2^2  \\
        &= \left\| \Pi_{\hat\cP} \left( \frac{1}{|\cH|} \sum_{i \in \cH} \vg_i \right) - \bbE \vmu \right\|_2^2  \tag{Lemma \ref{property:3}}\\
        &= \left\| \Pi_{\hat\cP} ( \vmu ) - \bbE \vmu \right\|_2^2 \\
        &= \| \Pi_{\hat\cP}(\vmu) - \Pi_{\cP^*}(\bbE \vmu) \|_2^2 \\
        &\leq 2 \| \Pi_{\hat\cP}(\vmu) - \Pi_{\hat\cP}(\bbE \vmu)\|_2^2 + 2 \| \Pi_{\hat\cP}(\bbE \vmu) - \Pi_{\cP^*}(\bbE \vmu) \|_2^2 \\
        &\leq 2\| \vmu - \bbE \vmu \|_2^2 + 2\| \Pi_{\hat\cP}(\bbE \vmu) - \Pi_{\cP^*}(\bbE \vmu) \|_2^2 \tag{Lemma \ref{property:2}}\\
        &= 2 \left \| \frac{1}{|\cH|}\sum_{i \in \cH} (\vg_i - \bbE\vg_i)  \right\|_2^2 + 2\| \Pi_{\hat\cP}(\bbE \vmu) - \Pi_{\cP^*}(\bbE \vmu) \|_2^2  \\
        &\leq 2 \frac{1}{|\cH|}\sum_{i \in \cH} \left\|\vg_i - \bbE\vg_i \right\|_2^2 + 2\| \Pi_{\hat\cP}(\bbE \vmu) - \Pi_{\cP^*}(\bbE \vmu) \|_2^2  \tag{Convexity of $\| \vx \|_2^2$}\\
        &= 2 \frac{1}{|\cH|} \sum_{i \in \cH} \left\| \vg_i - \bbE \vg_i \right\|_2^2 + 4 \left( \frac{1}{n - 2f} + \frac{\delta^2}{\sigma^2} \right) \left(\ell_t(\hat\cP) + \sum_{i \in \cH} \left\| \vg_i - \bbE \vg_i \right\|_2^2  \right)
    \end{align*}

\end{proof}

\begin{lemma}[Robustness of Stage 1 for server gradients] \label{lemma:stage_1_server}
    Let $\hat\cP$ denote the subspace fitted by BOBA stage 1 and $\ell_t(\hat\cP)$ be its corresponding trimmed reconstruction loss. For server gradients $\vgamma_1, \cdots, \vgamma_c$, 
    \begin{align*}
        \| \Delta \vGamma \|_2^2 \leq 2 \sum_{z=1}^c\| \vgamma_z - \bbE \vgamma_z \|_2^2 + 4 c\left( \frac{1}{n - 2f} + \frac{(\delta + \delta_s)^2}{\sigma^2} \right) \left( \ell_t(\hat\cP) + \sum_{i \in \cH} \left\| \vg_i - \bbE \vg_i \right\|_2^2  \right) 
    \end{align*}
    where $\Delta \vGamma = [\Pi_{\hat\cP}(\vgamma_1) - \bbE \vgamma_1, \cdots, \Pi_{\hat\cP}(\vgamma_c) - \bbE \vgamma_c] \in \bbR^{d \times c}$. Meanwhile, if we take expectation at both sides, 
    \begin{align*}
        \bbE \| \Delta \vGamma \|_2^2 
        &\leq 2 c \epsilon_s^2 + 4 c\left( \frac{1}{n - 2f} + \frac{(\delta + \delta_s)^2}{\sigma^2} \right) \left( \bbE \ell_t(\hat\cP) + |\cH| \epsilon^2  \right)  \\
        &\leq \begin{cases}
            4 c \left( \frac{1}{n-2f} + \frac{(\delta + \delta_s)^2}{\sigma^2}\right) (n - f + |\cH|) \epsilon^2 + \left(2c + 4 c \left( \frac{1}{n-2f} + \frac{(\delta + \delta_s)^2}{\sigma^2}\right) (n - f) \right) \epsilon_s^2 & \textnormal{(BOBA-ES)} \\
            4 c \left( \frac{1}{n-2f} + \frac{(\delta + \delta_s)^2}{\sigma^2}\right) (2(n - f) + |\cH|) \epsilon^2 + \left( 2c + 8c \left( \frac{1}{n-2f} + \frac{\delta + \delta_s)^2}{\sigma^2}\right) (n - f) \right) \epsilon_s^2 & \textnormal{(BOBA)} 
        \end{cases}
    \end{align*}
\end{lemma}
\begin{proof}
    Each server gradient can also be seen as a honest gradients. Therefore, we can use the same proof framework as Lemma \ref{lemma:stage_1}. We first derive upper bound of $\| \Pi_{\hat\cP}(\vgamma_z) - \bbE \vgamma_z \|_2^2$ for each $z \in \{1, \cdots, c\}$. 

    \textbf{Step 1. } Omitted, identical to the step 1 in the proof of \ref{lemma:stage_1}. 

    \textbf{Step 2. } Similar to step 2 in the proof of \ref{lemma:stage_1}, while the bound of $\| \vlambda \|_2$ need to be updated: 
    \begin{align*}
        \| \vtheta \|_2 &= \| \vA_s^+ (\bbE \vgamma_z - \bbE \vm_s) \|_2 \\
        &\leq \| \vA_s^+ \|_2 \cdot \| \bbE \vgamma_z - \bbE \vm_s \|_2 \\
        &\leq \frac{1}{\sigma} \cdot \| \bbE \vgamma_z - \bbE \vm_s \|_2 \tag{Assumption \ref{assumption:singular}(1)} \\
        &\leq \frac{1}{\sigma}  \cdot \left\| (\bbE \vgamma_z - \bbE \vmu) - \frac{1}{n - 2f} \sum_{i=1}^{n - 2f} (\bbE \vg_{s_i} - \bbE \vmu) \right\|_2 \\
        &\leq \frac{1}{\sigma}  \cdot \left( \| \bbE \vgamma_z - \bbE \vmu \|_2 + \frac{1}{n - 2f} \sum_{i=1}^{n - 2f} \| \bbE \vg_{s_i} - \bbE \vmu \|_2  \right) \\
        &\leq \frac{\delta + \delta_s}{\sigma} \tag{Assumption \ref{assumption:deviation}(2) and (4)}
    \end{align*}
    And therefore $\| \vlambda \|_2 \leq \sqrt{\frac{1}{n - 2f} + \frac{(\delta + \delta_s)^2}{\sigma^2}}$. 

    \textbf{Step 3. } Also identical to the step 3 in the proof of \ref{lemma:stage_1}. 
    \begin{align*}
        \| \Pi_{\hat\cP}(\bbE \vgamma_z) - \Pi_{\cP^*}(\bbE \vgamma_z) \|_2^2 \leq  \left( \frac{1}{n - 2f} + \frac{(\delta + \delta_s)^2}{\sigma^2} \right) \left( 2 \ell_t(\hat\cP) + 2 \sum_{i \in \cH} \left\| \vg_i - \bbE \vg_i \right\|_2^2  \right)
    \end{align*}

    \textbf{Step 4. }
    Finally, 
    \begin{align*}
        \| \Pi_{\hat\cP}(\vgamma_z) - \bbE \vgamma_z \|_2^2 
        &= \| \Pi_{\hat\cP}(\vgamma_z) - \Pi_{\cP^*}(\bbE \vgamma_z) \|_2^2 \\
        &\leq 2 \| \Pi_{\hat\cP}(\vgamma_z) - \Pi_{\hat\cP}(\bbE \vgamma_z)\|_2^2 + 2 \| \Pi_{\hat\cP}(\bbE \vgamma_z) - \Pi_{\cP^*}(\bbE \vgamma_z) \|_2^2 \\
        &\leq 2\| \vgamma_z - \bbE \vgamma_z \|_2^2 + 2\| \Pi_{\hat\cP}(\bbE \vgamma_z) - \Pi_{\cP^*}(\bbE \vgamma_z) \|_2^2 \tag{Lemma \ref{property:2}}\\
        &= 2\| \vgamma_z - \bbE \vgamma_z \|_2^2 + 4 \left( \frac{1}{n - 2f} + \frac{(\delta + \delta_s)^2}{\sigma^2} \right) \left( \ell_t(\hat\cP) + \sum_{i \in \cH} \left\| \vg_i - \bbE \vg_i \right\|_2^2  \right) \\
        \| \Delta \vGamma \|_2^2 &\leq \| \Delta \vGamma \|_F^2 \\
        &= \sum_{z=1}^c \| \Pi_{\hat\cP}(\vgamma_z) - \bbE \vgamma_z \|_2^2  \\
        &\leq 2 \sum_{z=1}^c\| \vgamma_z - \bbE \vgamma_z \|_2^2 + 4 c\left( \frac{1}{n - 2f} + \frac{(\delta + \delta_s)^2}{\sigma^2} \right) \left( \ell_t(\hat\cP) + \sum_{i \in \cH} \left\| \vg_i - \bbE \vg_i \right\|_2^2  \right) 
    \end{align*}
    
\end{proof}

\newpage
\subsubsection{Robustness of {\method} Stage 2} \label{subsubsec:stage2}

In {\method} stage 2, we estimate the label distribution for each client and discard abnormal clients with strongly negative elements in their label distribution. We use a filtering strategy with a hyper-parameter $p_{\min}$. In this subsubsection, we show that we can find a hyper-parameter $p_{\min}$ such that $|p_{\min}| \geq \| \hat \vp_h - \vp_h \|_2$, where $\vp_h$ is the true label distribution and $\hat \vp_h$ is the estimated label distribution, for a honest client $h \in \cH$. 

\begin{lemma}[Weyl's perturbation bound for singular values]\label{lemma:weyl}
    Let $\vA$ be a matrix with singular value $\sigma_1 \geq \cdots \geq \sigma_n$ and $\hat\vA = \vA + \Delta \vA$ be a perturbation of $\vA$, with corresponding singular value $\hat\sigma_1, \cdots, \hat\sigma_n$, we have
    \begin{align*}
        | \hat\sigma_i - \sigma_i | \leq \| \Delta \vA \|_2
    \end{align*}
\end{lemma}

\begin{proof}
    See proof by \cite{svd}. 
\end{proof}

We re-introduce some useful notation. Let
\begin{align*}
    \bbE \vGamma &= [\bbE \vgamma_1, \cdots, \bbE \vgamma_c] \in \bbR^{d \times c}\\
    \Pi_{\hat\cP} (\vGamma) &= [\Pi_{\hat\cP}(\vgamma_1), \cdots, \Pi_{\hat\cP}(\vgamma_c)] \in \bbR^{d \times c}\\
    \Delta \vGamma &= \Pi_{\hat\cP} (\vGamma) - \bbE \vGamma = [\Pi_{\hat\cP}(\vgamma_1) - \bbE \vgamma_1, \cdots, \Pi_{\hat\cP}(\vgamma_c) - \bbE \vgamma_c] \in \bbR^{d \times c}\\
    \Delta \vg_h &= \Pi_{\hat\cP} (\vg_h) - \bbE \vg_h \in \bbR^{d}
\end{align*}
The true and estimated label distributions of honest client $h \in \cH$ are denoted as $\vp_h, \hat\vp_h$, which follow
\begin{align*}
    \bbE \vg_h &= (\bbE \vGamma) \vp_h, \quad \Pi_{\hat\cP} (\vg_h) = \Pi_{\hat\cP}(\vGamma) \hat \vp_h
\end{align*}

\begin{lemma}[Robustness of stage 2] \label{lemma:stage_2}
    For any honest gradient $\vg_h$, we have
    \begin{align*}
        \left \| \Delta \vp_h \right \|_2  \leq \frac{1}{\sigma_s - \| \Delta \vGamma \|_2} \cdot \left[ \| \Delta \vg_h \|_2 + \sqrt{2} \| \Delta \vGamma \|_2 \right] 
    \end{align*}
    where $\Delta \vp_h = \hat\vp_h - \vp_h$. 
\end{lemma}

\begin{proof}
    We compare two linear systems: 
    \begin{align*}
        (\bbE \vGamma) \vp_h = \bbE \vg_h &, \quad \boldsymbol{1}^\top \vp_h = 1 \tag{System 1} \\
        (\Pi_{\hat\cP} (\vGamma)) \hat\vp_h = \Pi_{\hat\cP} (\vg_h) &, \quad  \boldsymbol{1}^\top \hat\vp_h = 1 \tag{System 2} 
    \end{align*}
    Different from solving the affine combination at step 3 of Lemma \ref{lemma:stage_1}, the solutions here to both linear systems are \textit{unique}. Therefore, we can use any method to express $\Delta \vp_h = \hat\vp_h - \vp_h$ and then get a corresponding bound of its 2-norm.

    It is also worth noting that the linear system in the algorithm/code is solved in latent space $\bbR^{c-1}$ instead of original space $\bbR^d$, which is much more efficient. However in this proof, we consider the problem in $\bbR^d$ to compare the fitted projection with the ideal projection. We still get the same solution of $\vp_h$ and $\hat\vp_h$, thus the bound is valid. 

    We first centralized both systems to remove the affine constraint. Let
    \begin{align*}
        \vA &= \bbE \vGamma \left( \vI - \frac{1}{c} \boldsymbol{1} \boldsymbol{1}^\top \right) &
        \hat \vA &= \Pi_{\hat\cP} (\vGamma) \left( \vI - \frac{1}{c} \boldsymbol{1} \boldsymbol{1}^\top \right)  &
        \Delta \vA &= \hat \vA - \vA \\
        \vb &= \bbE \vg_h - \bbE \vGamma \cdot \frac{1}{c} \boldsymbol{1} &
        \hat\vb &= \Pi_{\hat\cP}(\vg_h) - \Pi_{\hat\cP}(\vGamma) \cdot \frac{1}{c} \boldsymbol{1} & 
        \Delta \vb &= \hat \vb - \vb
    \end{align*}
    Previously, we have bounded $\| \Delta \vg_h \|_2$ and $\| \Delta \vGamma \|_2$ in Lemma \ref{lemma:stage_1} and \ref{lemma:stage_1_server}, respectively. We use them to give bounds of $\| \Delta \vA \|_2$ and $\| \Delta \vb \|_2$. 
    \begin{align*}
        \| \Delta \vA \|_2 &= \left \| \hat\vA - \vA \right\|_2 \\
        &= \left \| \Pi_{\hat\cP} (\vGamma) \left( \vI - \frac{1}{c} \boldsymbol{1} \boldsymbol{1}^\top \right) - \bbE \vGamma \left( \vI - \frac{1}{c} \boldsymbol{1} \boldsymbol{1}^\top \right) \right \|_2 \\
        &= \left \| \left( \Pi_{\hat\cP} (\vGamma) - \bbE \vGamma\right) \left( \vI - \frac{1}{c} \boldsymbol{1} \boldsymbol{1}^\top \right) \right \|_2 \\
        &\leq \left \| \Pi_{\hat\cP} (\vGamma) - \bbE \vGamma \right \|_2 \cdot \left \| \vI - \frac{1}{c} \boldsymbol{1} \boldsymbol{1}^\top \right\|_2 \\
        &\leq \left \| \Pi_{\hat\cP} (\vGamma) - \bbE \vGamma \right \|_2 \\
        &= \| \Delta \vGamma \|_2
    \end{align*}
    and similarly, 
    \begin{align*}
        \| \Delta \vb \|_2 &= \| \hat\vb - \vb \|_2 \\
        &= \left\| \left(\Pi_{\Pi_{\hat\cP}}(\vg_h) - \Pi_{\hat\cP}(\vGamma) \cdot \frac{1}{c} \boldsymbol{1}\right) - \left(\bbE \vg_h - \bbE \vGamma  \cdot \frac{1}{c} \boldsymbol{1} \right)\right\|_2 \\
        &= \left\| \left(\Pi_{\hat\cP}(\vg_h) - \bbE \vg_h \right) - \left( \Pi_{\hat\cP}(\vGamma) - \bbE \vGamma  \right) \cdot \frac{1}{c} \boldsymbol{1}  \right\|_2 \\
        &\leq \left\| \Pi_{\hat\cP}(\vg_h) - \bbE \vg_h \right\|_2 + \left\| \Pi_{\hat\cP}(\vGamma) - \bbE \vGamma \right\|_2 \cdot \left\| \frac{1}{c} \boldsymbol{1} \right\|_2 \\
        &= \| \Delta \vg_h \|_2 + \frac{1}{\sqrt{c}} \| \Delta \vGamma \|_2
    \end{align*}
    
    Then, instead of the original systems, we analyze the centralized systems
    \begin{align*}
        \vA \vx = \vb, \quad \hat\vA \hat\vx = \hat\vb
    \end{align*}
    with
    \begin{align*}
        \vx &= \vp_h - \frac{1}{c} \boldsymbol{1} &
        \hat\vx &= \hat\vp_h - \frac{1}{c} \boldsymbol{1} &
        \Delta \vx = \hat\vx - \vx
    \end{align*}
    By standard perturbation analysis of linear system, 
    \begin{align*}
        \hat\vA \hat\vx - \vA \vx &= \hat\vb - \vb \\
        \hat\vA (\Delta \vx) + (\Delta \vA) \vx &= \Delta\vb \\
        \hat\vA (\Delta \vx) &= \Delta\vb - (\Delta \vA) \vx  
    \end{align*}
    On the left hand side, $\Delta \vx \in \bbR^c$ but the rank of $\hat\vA$ is only $c - 1$. Usually, this results in an unbounded norm of $\Delta \vx$, as it can grow arbitrarily in the direction of the $c$-th right singular vector of $\hat\vA$. However, the $c$-th right singular vector of $\hat\vA$ is $\frac{1}{\sqrt{c}}\boldsymbol{1}$. 
    \begin{align*}
        \hat\vA \boldsymbol{1} = \Pi_{\hat\cP}(\vGamma) \left( \vI - \frac{1}{c} \boldsymbol{1} \boldsymbol{1}^\top \right) \boldsymbol{1} = \Pi_{\hat\cP}(\vGamma) \left( \boldsymbol{1} - \boldsymbol{1} \right) = \boldsymbol{0}
    \end{align*}
    But $\Delta \vx$ cannot grow in the direction of $\boldsymbol{1}$
    \begin{align*}
        \boldsymbol{1}^\top \Delta \vx &= \boldsymbol{1}^\top \left[\left(\hat\vp_h - \frac{1}{c} \boldsymbol{1} \right) - \left(\vp_h - \frac{1}{c} \boldsymbol{1} \right) \right] = \boldsymbol{1}^\top\hat\vp_h - \boldsymbol{1}^\top \vp_h = 1 - 1 = 0
    \end{align*}
    Thus, we can still bound $\Delta \vx$ with the $(c-1)$-th singular value of $\hat\vA$ (instead of the smallest singular value, $0$). We have
    \begin{align*}
        \hat \sigma_{c-1} \| \Delta \vx \|_2 &\leq \left\| \hat\vA (\Delta \vx) \right\|_2 \\
        &= \left \| \Delta\vb - (\Delta \vA) \vx \right\|_2 \\
        &\leq \| \Delta\vb \|_2 + \| \Delta \vA \|_2 \cdot \|\vx \|_2 \\
        \| \Delta \vx \|_2 &\leq \frac{1}{\hat \sigma_{c-1}} \left( \| \Delta\vb \|_2 + \| \Delta \vA \|_2 \cdot \|\vx \|_2 \right)
    \end{align*}
    $\| \Delta \vA \|_2$ and $\| \Delta \vb \|_2$ are already bounded, we still need to bound $\frac{1}{\hat \sigma_{c-1}}$ and $\| \vx \|_2$. 
    
    $\hat \sigma_{c-1}$ is the $(c-1)$-th singular value of $\hat\vA$, and is perturbed from $\sigma_{c-1}$, the $(c-1)$-th singular value of $\vA$. By Assumption \ref{assumption:singular} and Weyl's perturbation bound for singular value (Lemma \ref{lemma:weyl})
    \begin{align*}
        \hat \sigma_{c-1} &\geq \sigma_{c-1} - |\hat \sigma_{c-1} - \sigma_{c-1}| \\
        &\geq \sigma_{c-1} - \| \Delta \vA \|_2 \tag{Lemma \ref{lemma:weyl}} \\
        &\geq \sigma_s - \| \Delta \vA \|_2 \tag{Assumption \ref{assumption:singular}} \\
        &\geq \sigma_s - \| \Delta \vGamma \|_2 
    \end{align*}
    
    The 2-norm of $\vx$ can also be bounded, 
    \begin{align*}
        \| \vx \|_2 &= \left \| \vp_h - \frac{1}{c} \boldsymbol{1} \right \|_2 \\
        &= \sqrt{\left( \vp_h - \frac{1}{c} \boldsymbol{1} \right)^\top \left( \vp_h - \frac{1}{c} \boldsymbol{1} \right)} \\
        &= \sqrt{\vp_h^\top \vp_h - \frac{1}{c}} \\
        &\leq \sqrt{1 - \frac{1}{c}}
    \end{align*}
    
    Putting everything together, we have
    \begin{align*}
        \| \Delta \vp_h \|_2 &= \| \Delta \vx \|_2 \\
        &\leq \frac{1}{\hat \sigma_{c-1}} \cdot \left(\| \Delta\vb \|_2 + \| \Delta \vA \|_2 \cdot \|\vx \|_2\right) \\
        &\leq \frac{1}{\sigma_s - \| \Delta \vGamma \|_2} \cdot \left( \| \Delta \vg_h \|_2 + \frac{1}{\sqrt{c}} \| \Delta \vGamma \|_2 + \sqrt{1 - \frac{1}{c}}\| \Delta \vGamma \|_2 \right) \\
        &= \frac{1}{\sigma_s - \| \Delta \vGamma \|_2} \cdot \left[ \| \Delta \vg_h \|_2 + \left(\sqrt{\frac{1}{c}} +  \sqrt{1 - \frac{1}{c}} \right) \| \Delta \vGamma \|_2 \right] \\
        &\leq \frac{1}{\sigma_s - \| \Delta \vGamma \|_2} \cdot \left[ \| \Delta \vg_h \|_2 + \sqrt{2} \| \Delta \vGamma \|_2 \right] 
    \end{align*}
    when $\sigma_s - \| \Delta \vGamma \|_2 > 0$. 

\end{proof}


\begin{remark}
    We consider the case where $\| \Delta \vg_h \|_2 = \cO(\epsilon)$ and $\| \Delta \vGamma \|_2 = \cO(\sqrt{c} \epsilon)$ (see remarks of Lemma \ref{lemma:stage_1} and \ref{lemma:stage_1_server}). When the outer deviation dominates the inner deviation, $\sigma_s \gg \| \Delta \vGamma \|_2$, thus $\| \Delta \vp_h\|_2 = \cO(\frac{\sqrt{c}\epsilon}{\sigma_s})$. This means that we can set a small $|p_{\min}|$ and still preserve all honest gradients. 
\end{remark}

\newpage
\subsubsection{Robustness of {\method}} \label{subsubsec:robust}
So far, we have already proved two things. 
\begin{itemize}
    \item In stage 1, all honest gradients are only slightly perturbed. 
    \item In stage 2, all honest gradients are preserved, given 
\end{itemize}
In this subsubsection, we wrap up the theoretical results and provide the unbiasedness and robustness of {\method}. Specifically, 
\begin{itemize}
    \item When there are no Byzantine attack, {\method} is \textit{unbiased}. 
    \item When there are Byzantine attack, {\method} is \textit{robust} and has gradient estimation error of optimal order matching with the theoretical lower bound. 
\end{itemize}

\setcounter{mainsection}{5}
\setcounter{maintheorem}{4}

\begin{maintheorem}
    Let $\hat\vmu$ denote the aggregation result of {\method}. We have, 
    \begin{align*}
        \| \hat\vmu - \bbE \vmu \|_2^2
        &\leq 4 \frac{1}{|\cH|} \sum_{i \in \cH} \left\| \vg_i - \bbE \vg_i \right\|_2^2 + 8 \left( \frac{1}{n - 2f} + \frac{\delta^2}{\sigma^2} \right) \left(\ell_t(\hat\cP) + \sum_{i \in \cH} \left\| \vg_i - \bbE \vg_i \right\|_2^2  \right) \\
        &\quad\ + \beta^2 8 (1 + c |p_{\min}|)^2 \left( 2 \sum_{z=1}^c \| \vgamma_z - \bbE \vgamma_z \|_2^2 + 2 \delta_s^2 \right)
    \end{align*}
    Then we take expectation on both sides, 
    \begin{align*}
        \bbE \| \hat\vmu - \bbE \vmu \|_2^2 
        &\leq 4 \epsilon^2 + 8 \left( \frac{1}{n - 2f} + \frac{\delta^2}{\sigma^2} \right) \left(\bbE \ell_t(\hat\cP) + |\cH| \epsilon^2 \right) + 8 \beta^2 (1 + c |p_{\min}|)^2 (2 c \epsilon_s^2 + 2 \delta_s^2) \\
        &\leq \begin{cases}
            \left( 4 + 8 \left( \frac{1}{n - 2f} + \frac{\delta^2}{\sigma^2} \right)(n - f + |\cH|)\right)\epsilon^2 + 16 c (1 + c |p_{\min}|)^2 \beta^2 \epsilon_s^2 \\ \quad\quad + 16 (1 + c |p_{\min}|)^2 \beta^2 \delta_s^2 &  \textnormal{(BOBA-ES)} \\
            \left( 4 + 8 \left( \frac{1}{n - 2f} + \frac{\delta^2}{\sigma^2} \right)(2 (n - f) + |\cH|)\right)\epsilon^2 \\ \quad\quad+ \left( 16 \left( \frac{1}{n - 2f} + \frac{\delta^2}{\sigma^2} \right)(n - f) + 16 c (1 + c |p_{\min}|)^2 \beta^2 \right) \epsilon_s^2 \\ \quad\quad + 16 (1 + c |p_{\min}|)^2 \beta^2 \delta_s^2 &  \textnormal{(BOBA)}
        \end{cases}
    \end{align*}
\end{maintheorem}

\begin{proof}
    When some Byzantine clients are accepted, they can affect the aggregation result via biasing the average of estimated label distribution. Without loss of generality, we consider the worst case: all Byzantine gradients are accepted by {\method} stage 2. 

    We first decompose the gradient estimation error into two parts. Define $\hat\vmu_{\cH} = \frac{1}{|\cH|} \sum_{i \in \cH} \Pi_{\hat\cP}(\vg_i)$, we have
    \begin{align*}
        \| \hat\vmu - \bbE \vmu \|_2^2 \leq 2 \| \hat\vmu_{\cH} - \bbE \vmu \|_2^2  + 2 \| \hat \vmu - \hat \vmu_{\cH} \|_2^2
    \end{align*}
    The first term is already bounded in Lemma \ref{lemma:stage_1_avg}. We further bound the second term. Notice that, 
    \begin{align*}
        \hat\vmu_{\cH} &= \frac{1}{|\cH|} \sum_{i \in \cH} \Pi_{\hat\cP}(\vg_i) = \frac{1}{|\cH|} \sum_{i \in \cH} \Pi_{\hat\cP}(\vGamma) \hat\vp_i 
        = \Pi_{\hat\cP}(\vGamma) \left(\frac{1}{|\cH|} \sum_{i \in \cH} \hat\vp_i\right)
    \end{align*}
    where $\hat\vp_i$ is the estimated label distribution of client $i$. Similarly, 
    \begin{align*}
        \hat \vmu = \Pi_{\hat\cP}(\vGamma) \left(\frac{1}{n} \sum_{i=1}^n \hat\vp_i\right)
    \end{align*}
    We define 
    \begin{align*}
        \hat\vp_{\cH} = \frac{1}{|\cH|} \sum_{i \in \cH} \hat\vp_i, \quad \hat\vp_{\cB} = \frac{1}{|\cB|} \sum_{i \in \cB} \hat\vp_i, \quad \hat\vp = \frac{1}{n} \sum_{i=1}^n \hat\vp_i
    \end{align*}
    Then, 
    \begin{align*}
        \left\| \hat \vmu - \hat \vmu_{\cH} \right\|_2^2 
        &= \left\| \Pi_{\hat\cP}(\vGamma) \left(\hat\vp -  \hat\vp_{\cH} \right)\right\|_2^2 \\
        &= \left\| \left( \Pi_{\hat\cP}(\vGamma) - \Pi_{\hat\cP}(\bbE \vmu) \boldsymbol{1}^\top \right) \left(\hat\vp - \hat\vp_{\cH} \right)\right\|_2^2 \\
        &= \left \|\sum_{z = 1}^c (\hat\vp - \hat\vp_{\cH})_z \cdot \left(\Pi_{\hat\cP}(\vgamma_z) - \Pi_{\hat\cP}(\bbE\vmu)\right) \right\|_2^2 \\
        &\leq \left( \sum_{z = 1}^c |(\hat\vp - \hat\vp_{\cH})_z| \cdot \left \|\Pi_{\hat\cP}(\vgamma_z) - \Pi_{\hat\cP}(\bbE\vmu)\right\|_2 \right)^2 \\
        &\leq \left( \| \hat\vp -  \hat\vp_{\cH} \|_1 \cdot \left( \max_{z} \|\Pi_{\hat\cP}(\vgamma_z) - \Pi_{\hat\cP}(\bbE\vmu) \|_2 \right)  \right)^2 \\
        &= \| \hat\vp -  \hat\vp_{\cH} \|_1^2 \cdot \left( \max_{z} \|\Pi_{\hat\cP}(\vgamma_z) - \Pi_{\hat\cP}(\bbE\vmu) \|_2^2 \right)
    \end{align*}
    We first derive a bound for $\| \hat\vp -  \hat\vp_{\cH} \|_1$. In {\method} stage 2, a gradient will be accepted if and only if its estimated label distribution lies in the $(c-1)$-simplex of
    \begin{align*}
        \hat\vp_i \in \{\vq: \vq \geq p_{\min} \boldsymbol{1} , \boldsymbol{1}^\top \vq = 1 \}
    \end{align*}
    Since both $\hat\vp_{\cH}$ and $\hat\vp_{\cB}$ are averages of some $\hat\vp_i$ that lie inside the simplex above, we have
    \begin{align*}
        \hat\vp_{\cH}, \hat\vp_{\cB} \in \{\vq: \vq \geq p_{\min} \boldsymbol{1} , \boldsymbol{1}^\top \vq = 1 \}
    \end{align*}
    Therefore, denote $\beta = \frac{|\cB|}{n}$, we have
    \begin{align*}
        \| \hat\vp - \hat\vp_{\cH} \|_1 &= \left \| \frac{|\cH|}{n} \hat\vp_{\cH} + \frac{|\cB|}{n} \hat\vp_{\cB} -  \hat\vp_{\cH} \right \|_1 \\
        &= \frac{|\cB|}{n} \| \hat\vp_{\cB} - \hat\vp_{\cH} \|_1 \\
        &\leq \frac{|\cB|}{n} \left( \| \hat\vp_{\cB} - p_{\min} \boldsymbol{1} \|_1 + \| \hat\vp_{\cH}- p_{\min} \boldsymbol{1} \|_1 \right) \\
        &= \beta 2 (1 + c |p_{\min}|) 
    \end{align*}
    Then we derive a bound for $\max_{z} \|\Pi_{\hat\cP}(\vgamma_z) - \Pi_{\hat\cP}(\bbE\vmu) \|_2^2$. For each server gradient $\vgamma_z$, 
    \begin{align*}
        \max_z \| \Pi_{\hat\cP}(\vgamma_z) - \Pi_{\hat\cP}(\bbE \vmu) \|_2^2 
        &\leq \max_z \| \vgamma_z - \bbE \vmu \|_2^2 \tag{Lemma \ref{property:2}} \\
        &\leq \max_z \left( 2 \| \vgamma_z - \bbE \vgamma_z \|_2^2 + 2 \| \bbE \vgamma_z - \bbE \vmu \|_2^2 \right) \\
        &\leq 2 \sum_{z=1}^c \| \vgamma_z - \bbE \vgamma_z \|_2^2 + \max_z 2 \| \bbE \vgamma_z - \bbE \vmu \|_2^2 \\
        &\leq 2 \sum_{z=1}^c \| \vgamma_z - \bbE \vgamma_z \|_2^2 + 2 \delta_s^2 \tag{Assumption \ref{assumption:deviation}}
    \end{align*}
    Therefore, 
    \begin{align*}
        \left\| \hat \vmu - \hat \vmu_{\cH} \right\|_2^2 \leq \beta^2 4 (1 + c |p_{\min}|)^2 \left( 2 \sum_{z=1}^c \| \vgamma_z - \bbE \vgamma_z \|_2^2 + 2 \delta_s^2 \right)
    \end{align*}
    Put all together
    \begin{align*}
        \| \hat\vmu - \bbE \vmu \|_2^2
        &\leq 4 \frac{1}{|\cH|} \sum_{i \in \cH} \left\| \vg_i - \bbE \vg_i \right\|_2^2 + 8 \left( \frac{1}{n - 2f} + \frac{\delta^2}{\sigma^2} \right) \left(\ell_t(\hat\cP) + \sum_{i \in \cH} \left\| \vg_i - \bbE \vg_i \right\|_2^2  \right) \\
        &\quad\ + \beta^2 8 (1 + c |p_{\min}|)^2 \left( 2 \sum_{z=1}^c \| \vgamma_z - \bbE \vgamma_z \|_2^2 + 2 \delta_s^2 \right)
    \end{align*}
    Then we take expectation on both sides, 
    \begin{align*}
        \bbE \| \hat\vmu - \bbE \vmu \|_2^2 
        &\leq 4 \epsilon^2 + 8 \left( \frac{1}{n - 2f} + \frac{\delta^2}{\sigma^2} \right) \left(\bbE \ell_t(\hat\cP) + |\cH| \epsilon^2 \right) + 8 \beta^2 (1 + c |p_{\min}|)^2 (2 c \epsilon_s^2 + 2 \delta_s^2) \\
        &\leq \begin{cases}
            \left( 4 + 8 \left( \frac{1}{n - 2f} + \frac{\delta^2}{\sigma^2} \right)(n - f + |\cH|)\right)\epsilon^2 + 16 c (1 + c |p_{\min}|)^2 \beta^2 \epsilon_s^2 \\ \quad\quad + 16 (1 + c |p_{\min}|)^2 \beta^2 \delta_s^2 &  \textnormal{(BOBA-ES)} \\
            \left( 4 + 8 \left( \frac{1}{n - 2f} + \frac{\delta^2}{\sigma^2} \right)(2 (n - f) + |\cH|)\right)\epsilon^2 \\ \quad\quad+ \left( 16 \left( \frac{1}{n - 2f} + \frac{\delta^2}{\sigma^2} \right)(n - f) + 16 c (1 + c |p_{\min}|)^2 \beta^2 \right) \epsilon_s^2 \\ \quad\quad + 16 (1 + c |p_{\min}|)^2 \beta^2 \delta_s^2 &  \textnormal{(BOBA)}
        \end{cases}
    \end{align*}
\end{proof}

\begin{remark}
    We analyze the order of the gradient estimation error. When the outer variation increases $t$ times, i.e., $\bbE \vg_i \leftarrow \bbE t \vg_i$, both $\delta$ and $\sigma$ increase $t$ times. When all clients are duplication, i.e., $\vG \leftarrow [\vG, \vG]$, and $f \leftarrow 2f$, we have that $\delta^2$ does not change but $\sigma^2$ is doubled. Thus generally we have $\frac{\delta^2}{\sigma^2} \propto \frac{1}{n}$. When $\epsilon_s = \cO(\epsilon), \delta_s = \cO (\delta)$, $c = \cO(1), \frac{1}{n - 2f} = \cO(\frac{1}{n})$, $|\cH| = \cO(n)$, and $|p_{\min}| = \cO(1)$, we have $\| \hat\vmu - \bbE \vmu \|_2^2 = \cO(\epsilon^2 + \beta^2 \delta^2)$. We can conclude that
    \begin{align*}
        \bbE \| \hat\vmu - \bbE \vmu \|_2^2 = \cO(\epsilon^2 + \beta^2 \delta^2)
    \end{align*}
    Especially, when $\beta = 0$, we have
    \begin{align*}
        \bbE \| \hat\vmu - \bbE \vmu \|_2^2 = \cO(\epsilon^2)
    \end{align*}
\end{remark}

\paragraph{Comparison to Bucketing}
\cite{bucket} also have a similar claim in their Theorem II. However, their theorem heavily relies on the assumption that AGR is aware of $\beta$, the real fraction of Byzantine clients, as defined in their Definition A. (Their paper use $\delta$.) However, in practical FL systems, the fraction of Byzantine clients can be dynamic, and the AGR usually do not have precise knowledge of it. On the contrary, our {\method} algorithm does not require exact estimation of $\beta$; it only needs to satisfy Assumption \ref{assumption:singular} and the condition $f \geq |\cB|$. We also show in Appendix C.4 that {\method} has consistent performance under a wide range of $f$ and $|\cB|$.

\newpage

\subsection{Lower Bounds of Gradient Estimation Error}
\label{appendix:proof:grad_error_other}

\subsubsection{Lower Bounds of Gradient Estimation Error for Any AGR}

In the IID setting, as the inner variation approaches zero (i.e., $\epsilon \to 0$), the gradient estimation error of robust AGR typically also tends to zero (i.e., $\| \hat\vmu - \bbE \vmu \|_2^2 \to 0$). Consider the extreme case where each honest client uploads the same vector. In this scenario, robust AGR only needs to select the mode, i.e., the vector with the highest frequency from the collected vectors. This implies that the aggregation result is entirely immune to the influence of Byzantine clients. 

However, this intuition does not hold in non-IID settings, including cases with label skewness. On the contrary, for any AGR, as long as it remains unaware of the identity of Byzantine clients (i.e., which clients are honest and which are Byzantine), the best-case gradient estimation error can only be guaranteed to be $\cO(\beta^2 \delta^2)$, rather than approaching zero, even when $\epsilon$ is zero. We rigorously state the above proposition as Proposition \ref{prop:lower}.

\setcounter{mainsection}{5}
\setcounter{maintheorem}{3}

\begin{mainproposition}[Lower bound of gradient estimation error for any AGR]\label{prop:lower}
    Given any AGR, we can find $|\cH|$ honest gradients and $|\cB|$ Byzantine gradients, such that $\bbE \|\hat\vmu - \bbE \vmu \|_2^2 \geq \Omega(\beta^2 \delta^2)$. 
\end{mainproposition}

\begin{proof}
    W.l.o.g., we assume $n = |\cH| + |\cB|$ is even. We consider the following two sets of gradients, both with $|\cH|$ honest clients, $|\cB|$ Byzantine clients, zero inner variation, and outer variation bounded by $\delta$
    
    Gradient set 1:
    \begin{align*}
        \vg_i = \begin{cases}
            + \frac{|\cH|}{n} \delta, & i = 1, \cdots, \frac{n}{2} \\
            - \frac{|\cH|}{n} \delta, & i = \frac{n}{2} + 1, \cdots, n
        \end{cases}, \quad \cH = {1, \cdots, |\cH|}, \quad \cB = {|\cH| + 1, \cdots, n}
    \end{align*}

    Gradient set 2:
    \begin{align*}
        \vg_i = \begin{cases}
            + \frac{|\cH|}{n} \delta, & i = 1, \cdots, \frac{n}{2} \\
            - \frac{|\cH|}{n} \delta, & i = \frac{n}{2} + 1, \cdots, n
        \end{cases}, \quad \cH = {|\cB| + 1, \cdots, n}, \quad \cB = {1, \cdots, |\cB|}
    \end{align*}

    For the gradient set 1, 
    \begin{align*}
        \bbE \vmu^{(1)} = \frac{1}{|\cH|} \left( \frac{n}{2} \left( \frac{|\cH|}{n} \delta \right) + \left( |\cH| - \frac{n}{2}\right) \left(- \frac{|\cH|}{n} \delta \right)  \right) = \frac{|\cB|}{n} \delta
    \end{align*}
    while for the gradient set 2, $\bbE \vmu^{(2)} = - \frac{|\cB|}{n} \delta$. 

    Notice that the two gradient sets have the \textit{same gradient values}; the only difference is the identity of Byzantine clients. Since the input is identical, any AGR will give identical aggregation result for both gradient sets. Thus, 
    \begin{align*}
        \max\{\| \hat\vmu - \bbE \vmu^{(1)} \|_2, \| \hat\vmu - \bbE \vmu^{(2)} \|_2\} 
        &\geq \frac{1}{2} \left( \| \hat\vmu - \bbE \vmu^{(1)} \|_2, \| \hat\vmu - \bbE \vmu^{(2)} \|_2 \right) \\
        &\geq \frac{1}{2} \| \bbE \vmu^{(1)} - \bbE \vmu^{(2)} \|_2 \\
        &= \frac{|\cB|}{n} \delta \\
        &= \beta \delta
    \end{align*}
    equivalently, 
    \begin{align*}
        \max\{\| \hat\vmu - \bbE \vmu^{(1)} \|_2^2, \| \hat\vmu - \bbE \vmu^{(2)} \|_2^2\} \geq \beta^2 \delta^2
    \end{align*}
    which means that there exists one gradient set among set 1 and 2, such that the gradient estimation error is at least $\beta^2\delta^2$. 
\end{proof}

\begin{remark}
    Notice that this result is \textit{not} in contradiction with Theorem III in \cite{bucket}. Theorem III in \cite{bucket} gives an lower bound of $\Omega(\delta \zeta^2)$, where $\delta$ represents the fraction of Byzantines (equivalent to our $\beta$) and $\zeta$ represents the expected norm of outer variation (similar to our $\delta$). The discrepancy arises from a slight difference in the definition of outer variation. We define $\delta$ as the maximum norm of outer variation, while \cite{bucket} define $\zeta$ as the expected norm of outer variation. 
\end{remark}


\subsubsection{Lower Bounds of Gradient Estimation Error for Krum, CooMed and GeoMed}
\label{appendix:proof:lower_bound_existing_agr}

In this subsubsection, we prove that the gradient estimation error for Krum \citep{krum}, CooMed \citep{trmean} and GeoMed \citep{geomed} cannot be better than $\cO(\epsilon^2 + \delta^2)$. To prove this, we construct example where the gradient estimation error for the above three AGRs are all $\Omega(\epsilon^2 + \delta^2)$, even when $\beta = 0$ (no attacks). 

We construct a simple 3-client setting: 
\begin{align*}
    \vg_1 &= \frac{\delta}{2} \cdot \frac{1}{\sqrt{2}}\left[\begin{matrix} -1 \\ 1 \end{matrix}\right] + \epsilon (2 Z_1  - 1) \cdot \frac{1}{\sqrt{2}}\left[\begin{matrix} 1 \\ 1 \end{matrix}\right], \quad \text{where $Z_1 \sim$ Bernoulli}(0.5) \\
    \vg_2 &= \frac{\delta}{2} \cdot \frac{1}{\sqrt{2}}\left[\begin{matrix} -1 \\ 1 \end{matrix}\right] + \epsilon \cdot \frac{1}{\sqrt{2}}(2 Z_2  - 1) \left[\begin{matrix} 1 \\ 1 \end{matrix}\right], \quad \text{where $Z_2 \sim$ Bernoulli}(0.5) \\
    \vg_3 &= -\delta \cdot \frac{1}{\sqrt{2}}\left[\begin{matrix} -1 \\ 1 \end{matrix}\right]
\end{align*}
Their expectations: 
\begin{align*}
    \bbE \vg_1 = \bbE \vg_2 = \frac{\delta}{2} \cdot \frac{1}{\sqrt{2}}\left[\begin{matrix} -1 \\ 1 \end{matrix}\right], \quad 
    \bbE \vg_3 = -\delta \cdot \frac{1}{\sqrt{2}}\left[\begin{matrix} -1 \\ 1 \end{matrix}\right], \quad 
    \bbE\vmu = \frac{\bbE \vg_1 + \bbE \vg_2 + \bbE \vg_3}{3} = \left[\begin{matrix} 0 \\ 0 \end{matrix}\right]
\end{align*}
Moreover, we consider $\delta > 2 \epsilon$. We can easily verify the bounded inner/outer variations. 

\paragraph{Krum}
No matter how $Z_1, Z_2$ are chosen, we always have $\| \vg_1 - \vg_2 \|_2 < \| \vg_1 - \vg_3 \|_2$ and $\| \vg_1 - \vg_2 \|_2 < \| \vg_2 - \vg_3 \|_2$. Therefore, Krum will always choose from $\vg_1$ and $\vg_2$. In both case, $\| \hat\vmu - \bbE \vmu \|_2^2 = \epsilon^2 + \frac{\delta^2}{4}$. 

\paragraph{CooMed}
\begin{align*}
    \hat\vmu = \begin{cases}
        \left[\begin{matrix} - \frac{\delta - 2\epsilon}{2\sqrt{2}}, \frac{\delta - 2\epsilon}{2\sqrt{2}}\end{matrix}\right]^\top, & (Z_1, Z_2) \in \{(1, 0), (0, 1)\} \\
        \left[\begin{matrix} - \frac{\delta - 2\epsilon}{2\sqrt{2}}, \frac{\delta + 2\epsilon}{2\sqrt{2}}\end{matrix}\right]^\top, & (Z_1, Z_2) = (1, 1) \\
        \left[\begin{matrix} - \frac{\delta + 2\epsilon}{2\sqrt{2}}, \frac{\delta - 2\epsilon}{2\sqrt{2}}\end{matrix}\right]^\top, & (Z_1, Z_2) = (0, 0) \\
    \end{cases}
\end{align*}
Therefore, 
\begin{align*}
    \bbE \| \hat\vmu - \bbE \vmu \|_2^2 &= \frac{1}{2} \cdot 2 \left( \frac{\delta - 2\epsilon}{2\sqrt{2}} \right)^2 + \frac{1}{2} \cdot \left[ \left( \frac{\delta + 2\epsilon}{2\sqrt{2}} \right)^2 + \left( \frac{\delta - 2\epsilon}{2\sqrt{2}} \right)^2 \right] \\
    &= \frac{(\delta - 2\epsilon)^2}{8} + \frac{\delta^2}{8} + \frac{\epsilon^2}{2} \\
    &> \frac{\delta^2}{8} + \frac{\epsilon^2}{2}
\end{align*}

\paragraph{GeoMed}
\begin{align*}
    \hat\vmu = \begin{cases}
        - (\frac{\delta}{2} - \frac{\epsilon}{\sqrt{3}}) \cdot \frac{1}{\sqrt{2}} [-1, 1]^\top, & (Z_1, Z_2) \in \{(1, 0), (0, 1)\} \\
        \left[\begin{matrix} - \frac{\delta - 2\epsilon}{2\sqrt{2}}, \frac{\delta + 2\epsilon}{2\sqrt{2}}\end{matrix}\right]^\top, & (Z_1, Z_2) = (1, 1) \\
        \left[\begin{matrix} - \frac{\delta + 2\epsilon}{2\sqrt{2}}, \frac{\delta - 2\epsilon}{2\sqrt{2}}\end{matrix}\right]^\top, & (Z_1, Z_2) = (0, 0) \\
    \end{cases}
\end{align*}
Therefore, 
\begin{align*}
    \bbE \| \hat\vmu - \bbE \vmu \|_2^2 &= \frac{1}{2} \cdot \left( \frac{\delta}{2} - \frac{\epsilon}{\sqrt{3}} \right)^2 + \frac{1}{2} \cdot \left[ \left( \frac{\delta + 2\epsilon}{2\sqrt{2}} \right)^2 + \left( \frac{\delta - 2\epsilon}{2\sqrt{2}} \right)^2 \right] \\
    &= \frac{1}{2} \cdot \left( \frac{\delta}{2} - \frac{\epsilon}{\sqrt{3}} \right)^2 + \frac{\delta^2}{8} + \frac{\epsilon^2}{2} \\
    &> \frac{\delta^2}{8} + \frac{\epsilon^2}{2}
\end{align*}






\newpage
\subsubsection{Impossible Unbiasedness and Robustness without Server Data}

As mentioned in Subsection 3.2, AGR in label skewness faces two challenges: selection bias and increased vulnerability. In other words, we aim for AGR to possess unbiasedness and robustness, which are rigorously defined in Definition \ref{def:unbiased} and \ref{def:robust}, 

\begin{definition}[Unbiasedness] \label{def:unbiased}
    An AGR is unbiased if for any $\vw_G$, $\| \hat\vmu - \bbE \vmu \|_2^2 \to 0$ when $\epsilon \to 0$ and $\beta = 0$ (i.e., no attacks). 
\end{definition}

\begin{definition}[Robustness]  \label{def:robust}
    An AGR is robust if there exist $\Delta > 0$ such that for any $\vw_G$, $\| \hat\vmu - \bbE \vmu \|_2^2 \leq \Delta^2$. 
\end{definition}

\begin{proposition}[Trade-Off Between Unbiasedness and Robustness]
    For any AGR, if it can only utilize $n$ gradients without relying on any other information, then it is impossible for it to be both unbiased and robust. 
\end{proposition}

\begin{proof}
    We consider the following machine learning task: 
    \begin{align*}
        \min_{b \in \bbR} \cL = \bbE_{(x, y)} \ell(x, y | w), \quad \text{where } \ell(x, y | w) = [y - (x - w)]^2
    \end{align*}
    whose gradient w.r.t. $w$ is $\frac{\partial \ell}{\partial w} = 2(w - (x - y))$. We let $x, w \in \bbR$ and $y \in \{-1, +1\}$. We start with $w_G^{(0)} = 0$. 

    We consider the following two sets of gradients. 
    
    Gradient set 1: 
    \begin{itemize}
        \item Client 1 and 2 are honest, with data $(x, y) = (0.5, 1)$
        \item Client 3 and 4 are honest, with data $(x, y) = (-0.5, -1)$
        \item Client 5 is Byzantine
    \end{itemize}
    This results in the following gradients: 
    \begin{align*}
        g_1 = g_2 = + 1, \quad g_3 = g_4 = - 1, \quad g_5 = k, \quad \cH = \{1, 2, 3, 4\}, \quad \cB = \{5\}
    \end{align*}

    Gradient set 2: 
    \begin{itemize}
        \item Client 5 is honest with $(x, y) = (1 - \frac{k}{2}, 1)$
        \item Client 3 and 4 are honest, with data $(x, y) = (-0.5, -1)$
        \item Client 1 and 2 are honest, with $\frac{1}{k+1}$ of their data as $(x, y) = (1 - \frac{k}{2}, 1)$ amd $\frac{k}{k+1}$ of their data as $(x, y) = (-0.5, -1)$
    \end{itemize}
    This results in the following gradients: 
    \begin{align*}
        g_1 = g_2 = + 1, \quad g_3 = g_4 = - 1, \quad g_5 = k, \quad \cH = \{1, 2, 3, 4, 5\}, \quad \cB = \emptyset
    \end{align*}

    For gradient set 1, we have inner variation upper bound $\epsilon = 0$, outer variation bound $\delta = 1$; for gradient set 2, we have inner variation upper bound $\epsilon = 0$, outer variation bound $\delta = \frac{4}{5}k$. 

    Notice that the two gradient sets have the \textit{same gradient values}; the only difference is the identity of Byzantine clients. Therefore, any AGR only utilizing $n$ gradients will give identical aggregation results for two gradient sets. To achieve unbiasedness in gradient set 2, for all $k > 1$, the aggregation result must be $\hat\mu^{(1)} = \frac{1}{5}k$. Its aggregation result on gradient set 1 will also be $\hat\mu^{(2)} = \frac{1}{5}k$. Let $k \to \infty$, then the gradient estimation error on gradient set 1 will be unbounded, which violates robustness. 
\end{proof}

\begin{remark}
    This unbiasedness-robustness trade-off can be circumvented by using additional server data. Consider that the two server gradients are $\gamma_1 = +1.5, \gamma_2 = -1.5$, then for any $k \gg 1.5$, the AGR can guarantee that it must be a Byzantine gradient, i.e., gradient set 2 is not valid. 
\end{remark}

\newpage

\subsection{Computation Complexity of {\method}}
\label{appendix:proof:complexity}

\setcounter{algorithm}{0}

\begin{algorithm}[h]
    \caption{{\method} Framework \label{alg:BOBA_appendix}} 
    \small
    \begin{algorithmic}[1]
        \REQUIRE $\vG = [\vg_1, \cdots, \vg_n]$, $\vGamma = [\vgamma_1, \cdots, \vgamma_c]$, $n, f, c, p_{\min}$
        \ENSURE Aggregation result $\hat\vmu$
        \STATE Initialize subspace $\hat\cP$: $\vm, \vU, \vSigma, \vV = \text{TrSVD}_{c-1}(\vGamma)$
        \WHILE{not converge}
            \STATE Update $\vr$: $\vG_{[n - f]} = \{ n - f$ gradients in $\vG$ with smallest $\| \vg_i - \Pi_{\hat\cP}(\vg_i) \|_2\}$ where $\Pi_{\hat\cP}(\vg_i) = \vU \vU^\top (\vg_i - \vm) + \vm$
            \STATE Update $\hat\cP$: $\vm, \vU, \vSigma, \vV = \text{TrSVD}_{c-1}(\vG_{[n-f]})$
        \ENDWHILE
        \STATE Encode: $\tilde{\vg_i} = \vU^\top (\vg_i - \vm), \forall i$; $\tilde{\vGamma} = \vU^\top (\vGamma - \vm \boldsymbol{1}^\top)$
        \STATE Estimate: $\hat \vp_i = \left[\begin{matrix}\tilde{\vGamma} \\ \boldsymbol{1}^\top \end{matrix} \right]^{-1} \left[ \begin{matrix} \tilde{\vg}_i \\ 1 \end{matrix} \right], \forall i$ 
        \STATE Filter: $\va = \cA(\{\hat\vp_i\}_{i=1}^n)$
        \STATE Aggregate: $\tilde \vmu = \sum_{i=1}^n a_i \tilde \vg_i / \sum_{i=1}^n a_i$
        \STATE Decode: $\hat\vmu = \vU \tilde\vg_G + \vm$
    \end{algorithmic}
\end{algorithm}

In this subsection, we provide a detail analysis of the complexity of {\method} (Algorithm \ref{alg:BOBA}). We use the results that the complexity of TrSVD is $\cO(cnd)$ \citep{random_svd}. 

\begin{itemize}
    \item Line 1: The complexity is $\cO(cnd)$. 
    \item Line 3: The complexity is $\cO(cnd + n \log n)$, where $\cO(cnd)$ comes from computing $\| \vg_i - \Pi_{\hat\cP}(\vg_i)\|_2$ for $n$ gradients $\vg_1, \cdots, \vg_n$, and $\cO(n \log n)$ comes from sorting all $n$ distances and select the smallest $n - f$. 
    \item Line 4: The complexity is $\cO(cnd)$. 
    \item Line 5: The complexity is $\cO(cnd)$. 
    \item Line 6: The complexity is $\cO(c^3 + c^2 n)$, where $\cO(c^3)$ comes from computing the inverse matrix $\left[\begin{matrix}\tilde{\vGamma} \\ \boldsymbol{1}^\top \end{matrix} \right]^{-1}$ and $\cO(c^2 n)$ arises from computing $\hat \vp_i$ for $i = 1, \cdots, n$. 
    \item Line 7: The complexity is $\cO(cn + n \log n)$, where $\cO(cn)$ comes from computing $\min_z p_{iz}$ for each client $i$, and $\cO(n \log n)$ arises from computing the quantile. 
    \item Line 8: The complexity is $\cO(cn)$. 
    \item Line 9: The complexity is $\cO(cd)$. 
\end{itemize}

Assuming that $c < n < d$, the overall complexity is dominated by Line 3 and 4, which are conducted by $k$ times. Therefore, the total complexity is $\cO(kcnd)$.

\newpage
\section{ADDITIONAL EXPERIMENTS}

\subsection{Experimental Setup}
\label{appendix:exp:setup}

In this part, we provide detailed experimental setup. 

\begin{table}[h]
  \centering
  \vspace{-1ex}
  \caption{Experimental settings summary  \label{tab:setup}}
  \vspace{1ex}
  \small
  \begin{tabular}{lcccc}
    \toprule
    \ & MNIST & CIFAR-10 & AG-News & AG-News (Ablation Study) \\
    \midrule
    \# Training Samples & 60,000 & 50,000 & 120,000 & 120,000 \\
    \# Testing Samples & 10,000 & 10,000 & 7,600 & 7,600 \\
    \# Classes $c$ & 10 & 10 & 4 & 4 \\
    \# Rounds & 200 & 2,000 & 200 & 200 \\
    Initial LR $\eta_0$ & 0.1         & 0.2         & 1.0   & 1.0  \\
    LR Decay $(T_s; T_i; \alpha)$ & (100; 10; 0.95) & (1,000; 100; 0.8) & (100; 10; 0.95) & (100; 10; 0.95) \\
    \# Honest Clients $|\cH|$ & 100 & 100 & 160 & 16\\
    Real \# Byzantine Clients $|\cB|$ & 0 or 15 & 0 or 15 & 0 or 54 & 0 or 2\\
    Declared \# Byzantine Clients $f$ & 16 & 16 & 60 & 2 \\
    \# Server Samples Per Class & 20 & 20 & 30 & 30 \\
    \bottomrule
  \end{tabular}
\end{table}

\paragraph{Training setup}
The setup for FL training is summarized in Table \ref{tab:setup}. Specifically, 
\begin{itemize}
    \item \textit{Data partition}. We use the pathological data partitioning proposed by \cite{FedAvg}. We first sort data samples based on labels and evenly divided the training set into $n_s \cdot |\cH|$ shards. As a result, each shard only contains one class of data \footnote{For MNIST, since the dataset is not strictly balanced, the size of each shard is slightly different to ensure that each shard only contains one class of data. }. We then assign $n_s$ shards to each honest client, so that most clients have only $n_s$ classes of samples. We let $n_s = 2$ for MNIST, CIFAR-10 and AG-News. 
    \item \textit{Models}. For MNIST \citep{mnist}, we train a 3-layer MLP with hidden layers of width 200. This network is the same as ``2NN'' in \citep{FedAvg}. For CIFAR-10 \citep{cifar}, we train a 5-layer CNN model as it in the TensorFlow tutorial \footnote{https://www.tensorflow.org/tutorials/images/cnn}. For AG News \citep{ag_news}, we train a RNN model containing a uni-directional GRU layer with 32 hidden units followed by a global pooling layer and a linear layer. 
    \item \textit{Learning rate strategy}. We use a constant learning rate at the early stage, and exponential learning rate decay at the end to stabilize the training. In detail, we start with an initial learning rate $\eta = \eta_0$ until the $T_s$ round, and then exponentially decrease it with $\eta \leftarrow \alpha \eta$ every $T_i$ rounds. 
\end{itemize}

\paragraph{Attacks}
We consider six types of attacks. 
\begin{itemize}
    \item \textit{Gauss} \citep{krum}, a non-colluding attack uploading large-scale vectors from Gaussian distribution $\cN(0, 200\vI)$. 
    \item \textit{IPM} \citep{inner_prod}, a colluding attack uploading $\vg_k = - \gamma \cdot \frac{1}{|\cH|}\sum_{i \in \cH}\vg_i$. While \cite{inner_prod} test $\gamma \in \{-10, 0, 0.1, 10\}$, we choose the strongest $\gamma = 10$, making Average to perform gradient ascent. 
    \item \textit{LIE} \citep{little}, a colluding attack uploading vectors within the scope of honest ones to bias the model while avoiding being detected. We set the hyper-parameter according to the original paper, i.e., $z = \phi^{-1}\left((n - \lfloor n/2 + 1 \rfloor) / (n - |\cB|)\right)$. 
    \item \textit{Mimic} \citep{bucket}, a colluding attack inserting consistent bias by always copying the gradient from a particular client with biased label distribution. This attack is on the 
    \item \textit{MinMax} and \textit{MinSum} \citep{dnc}, a colluding attack that maximize the effect of attack while not being detected. We use coordinate-wise standard deviation as the perturbation vector $\nabla^p$, and optimize the magnitude according to Algorithm 1 with $\gamma_{\text{init}} = 10$ and $\tau = 10^{-5}$. 
\end{itemize}

\paragraph{Baseline AGRs}
We consider 15 baseline AGRs
\begin{itemize}
    \item \textit{Average} \citep{FedAvg} simply averages all gradients. It is unbiased but vulnerable to attacks. 
    \item \textit{Server} only uses server data to fit a model. We use it to verify that one cannot train a good model with server data only. 
    \item \textit{CooMed} and \textit{TrMean} \citep{trmean} use coordinate-wise median or trimmed mean as the aggregation. For TrMean we trimmed the largest and smallest $f$ entries. 
    \item \textit{Krum} and \textit{Multi-Krum} \citep{krum} find the one or $m$ gradients that is closest to its $k$ nearest neighbors. We use $k = n - f - 2$ and $m = n - f$, according to the original paper. 
    \item \textit{GeoMed} \citep{geomed, rfa} computes the geometric median as the aggregation. We use the implementation in \textit{hdmedians} Python package.  
    \item \textit{SelfRej} and \textit{AvgRej} \citep{rej} evaluate client gradients with their loss on server data. SelfRej selects $n-f$ clients whose local models $\vw_i = \vw_G - \eta \vg_i$ have smallest loss, while AvgRej selects $n-f$ clients whose gradients can lower the loss of averaged model the most. 
    \item \textit{Zeno} \citep{zeno} considers both loss and gradient scales, select $n-f$ gradients with small loss and small gradient. We optimize $\rho$ on MNIST, and finally use $\rho = 5 \times 10^{-4}$ for all experiments. 
    \item \textit{FLTrust} \citep{fltrust} uses server data to estimate one server gradient, and use each gradient' clipped cosine similarity as the weight to re-weight each gradient and aggregate. 
    \item \textit{ByGARS} \citep{bygars} optimizes the aggregation weights of client gradients with server data as training set. Different from the original implementation, we optimize the aggregation weight $\vq$ for each communication round independently. We optimize hyperparameters on MNIST and finally use $k = 3$ and $\alpha = 0.05$. 
    \item \textit{Bucketing} \citep{bucket}. We consider bucketing ($s=2$) with Krum (B-Krum) / MKrum (B-MKrum). 
    \item \textit{RAGE} \citep{rage}. Considering that $C$ is usually unknown to the server, we run the while loop for fixed $f$ iterations, to make sure it successfully mitigate the Gauss attack. 
\end{itemize}

\paragraph{Image corruptions}
We simulate feature skewness by applying different image corruptions to each client. For each client, we randomly choose one kind of corruption (severity $= 3$) for its local training dataset. We do not add corruptions to the server data and the testing data, in order to make comparison with the corruption-free setting. 

\paragraph{Computation}
We did our experiments with single NVIDIA Tesla V100 GPU. 

\paragraph{AGR running time (RQ2)}
In the main text, we record the running time for all AGRs. For a fair comparison, we run all AGRs with an Intel Core i9-11900 Processor. Specifically, 
\begin{itemize}
    \item For {\method} and FLTrust, we do not include the time taken to compute the server gradient because this computation can be finished simultaneously with the client-side gradient computations. 
    \item However, for SelfRej, AvgRej, Zeno, and ByGARS, we include the time to perform inference or compute gradients using server data, because these computations must occur after the server receives the gradient from each client.
\end{itemize}

\newpage
\subsection{Majority-based AGRs with Additional Server Data (RQ1)}
\label{appendix:exp:boost_agr}

Reference-based AGRs, including {\method}, use additional server data for aggregation. However, majority-based AGRs do not require server data. To make a fair comparison, we study whether server data can further enhance baseline majority-based AGRs, especially the strongest ones. Specifically, we enhance the baselines majority-based AGRs with 
\begin{align*}
    \hat\vmu = (1 - \lambda) \agg(\{\vg_i\}_{i=1}^n) + \lambda \left(\frac{1}{c} \sum_{z=1}^c \vgamma_z\right)
\end{align*}
where $\agg(\{\vg_i\}_{i=1}^n)$ is the aggregation given by baseline majority-based AGRs, $\frac{1}{c} \sum_{z=1}^c \vgamma_z$ is the averaged server gradient, and $\lambda \in [0, 1]$ is the hyperparameter for convex combination. The underlying intuition here is that the AGR's output has a smaller variance (due to computing gradients using more data from clients), while the server gradient has a smaller bias (as it is not affected by selection bias). By combining these two outputs, a potentially better bias-variance tradeoff can be achieved, leading to improved aggregation results.

We test $\lambda \in \{0, 0.25, 0.5, 0.75, 1.0\}$. Notice that $\lambda = 0$ refers to vanilla baseline AGRs without server data, and $\lambda = 1$ refers to training with server data only. We test these enhanced AGRs with MNIST dataset. 

\begin{figure*}[h!]
    \centering
    \includegraphics[width=1.0\linewidth]{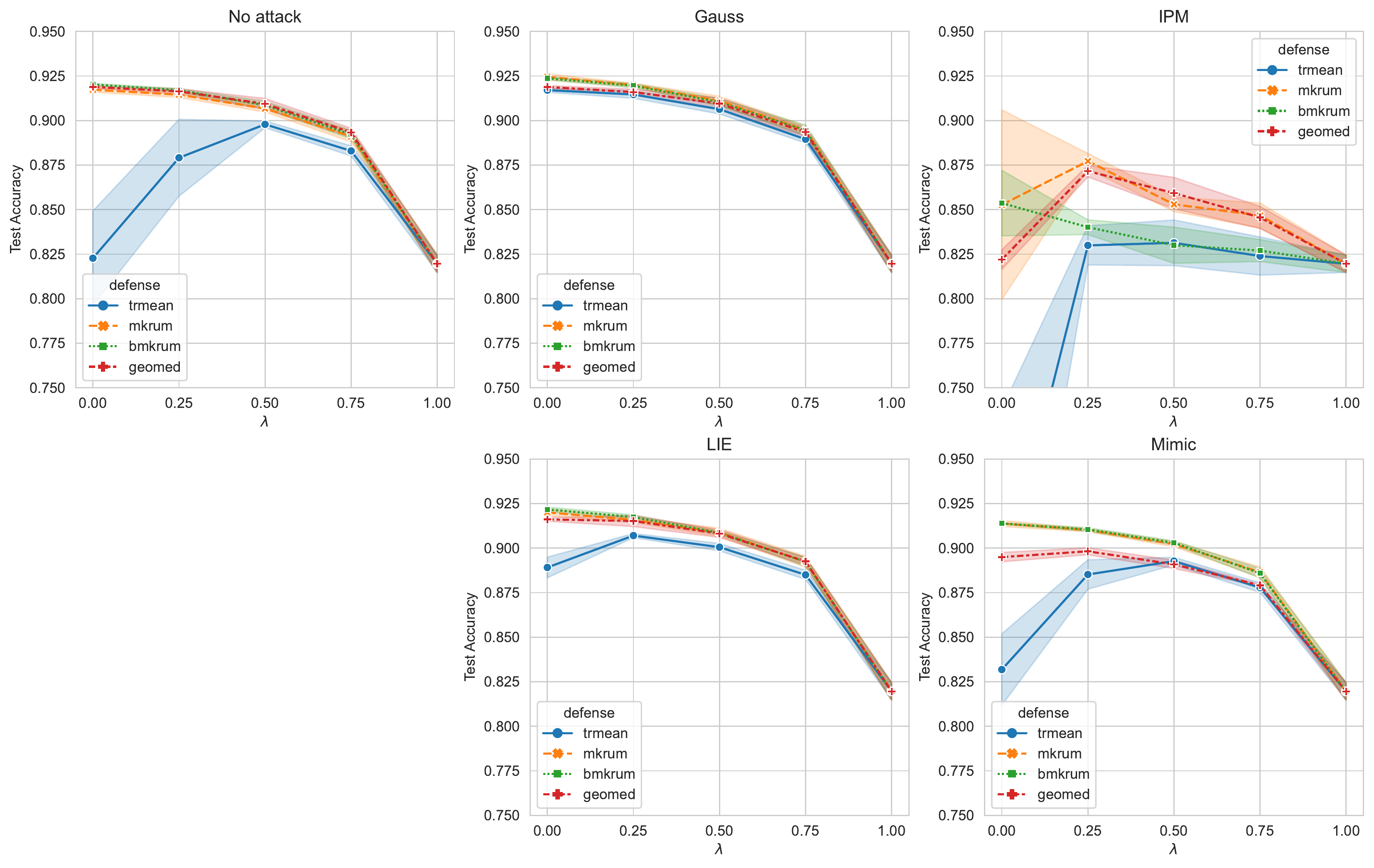}
    \caption{Performance of baseline majority-based AGRs when they use additional server data. }
    \label{fig:hyper_iid}
\end{figure*}

Figure \ref{fig:hyper_iid} shows that TrMean has better performance when it is combined with server data ($\lambda=0.25, 0.5$). However, the most competitive baseline majority-based AGRs (MKrum, BMKrum, GeoMed) get no performance improvement in most settings. We also notice that using additional server data can improve the worst-case test accuracy for most robust AGRs (usually under IPM attack). However, they are still significantly worse than {\method}, whose worst-case test accuracy is 91.6\%. 

\newpage
\subsection{Effect of Server Data (RQ3)}
\label{appendix:exp:server_data}

{\method} relies on server data for aggregation. In this subsection, we investigate how the quality and quantity of server data impact {\method}. Specifically, regarding data quality, we examine whether {\method}'s performance is affected when server data contains noise (skewed feature distribution) or skewed label distribution. Concerning data quantity, we explore how much server data is sufficient for BOBA to perform robust aggregation.

\subsubsection{Server Data with Feature Skewness}

\begin{figure}[h]
    \centering
    \includegraphics[width=0.45\linewidth]{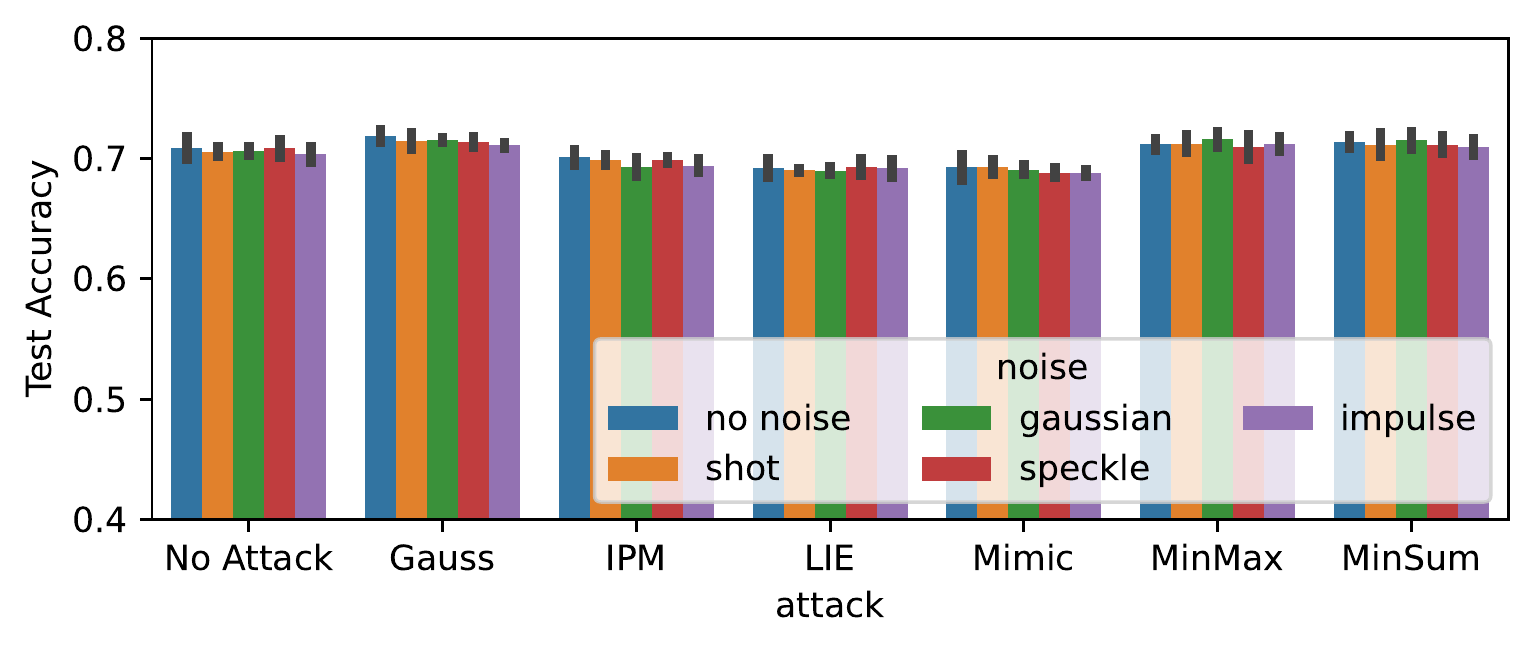}
    \caption{{\method} is robust to corrupted server data \label{fig:noise2}}
\end{figure}

We first investigate whether the performance of {\method} is robust to feature skewness of server data. To simulate low-quality data, we introduce four types of random noises to the server data, following the approach proposed by \citep{corruption}. As illustrated in Figure \ref{fig:noise2}, {\method} exhibits remarkable consistency across various noise types, highlighting its robustness to variations in server data quality. 

\newpage
\subsubsection{Server Data with Label Skewness}

In our experiments in the main text, the server dataset and testing dataset share the same label distribution. This assumption may be violated in real-world FL systems. In this subsubsection, we investigate the impact on reference-based AGR methods, including BOBA, when the server data also exhibits label skewness. 

\begin{figure*}[h!]
    \centering
    \includegraphics[width=1.0\linewidth]{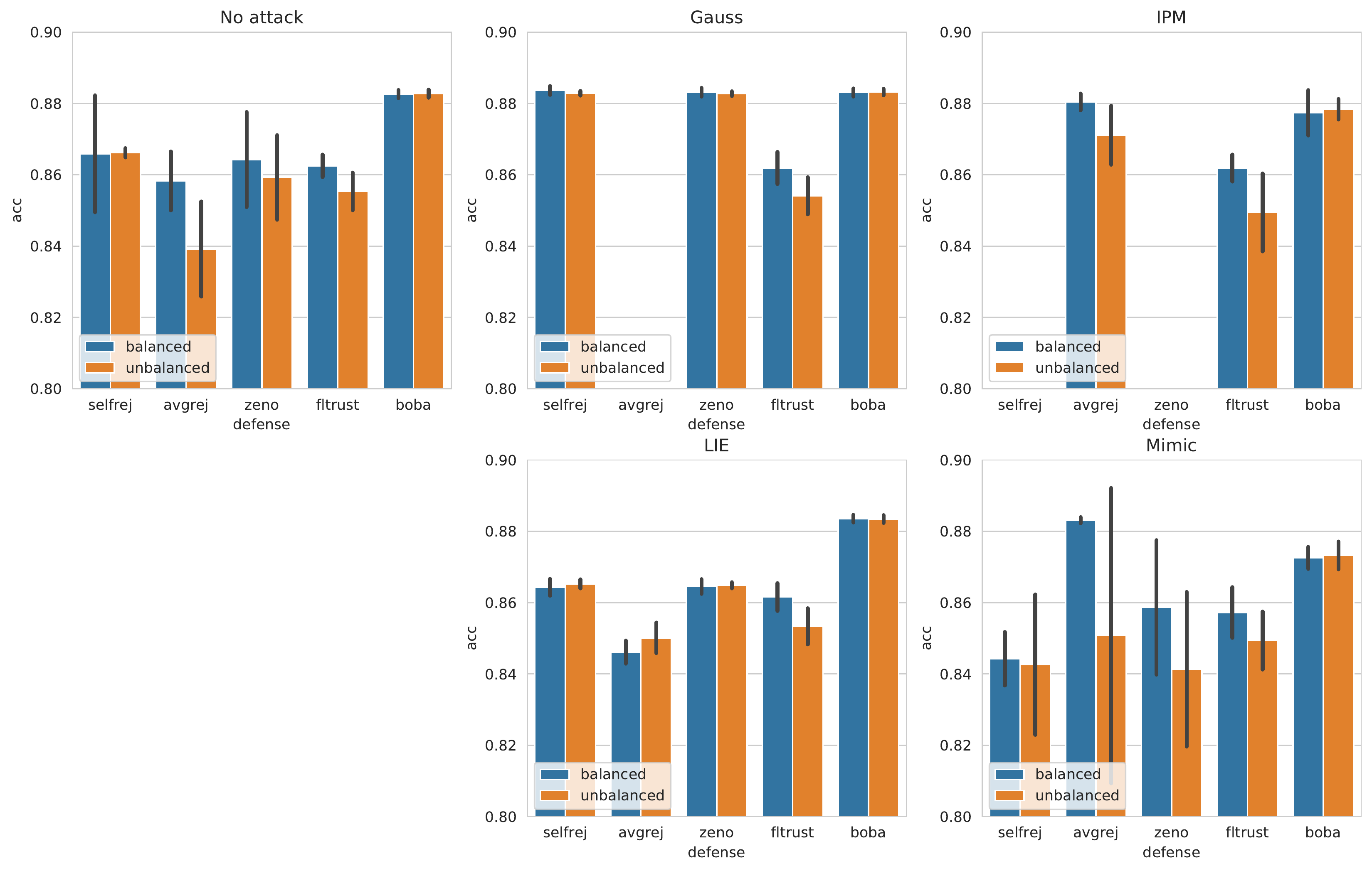}
    \caption{Comparison between aggregators using server data when server data is biased. }
    \label{fig:hyper_serverbias}
\end{figure*}

We conducted experiments on the AG-News dataset, and the results are presented in Figure \ref{fig:hyper_serverbias}. The ``balanced'' setting corresponds to the one in the main text, where the server has 30 samples for each class. In the ``unbalanced'' setting, the server has 40 samples for classes 0 and 1, and 20 samples for classes 2 and 3, thus introducing label skewness between the server data and test data, while the total amount of server data remains the same. 

As shown in Figure \ref{fig:hyper_serverbias}, the performance of baseline AGRs generally degrades when the server data becomes unbalanced. However, the performance of {\method} remains almost the same across non-attack settings and four attacks, showing that {\method} is also robust to the label skewness of server data. 

\newpage
\subsubsection{Quantity of Server Data}

Finally, we investigate the impact of the quantity of the server dataset on {\method}. We test on CIFAR-10 data set when the number of server data per class varies from 1 to 320, with $|\cB| = 15$ IPM attackers. 

\begin{figure*}[h!]
    \centering
    \includegraphics[width=0.45\linewidth]{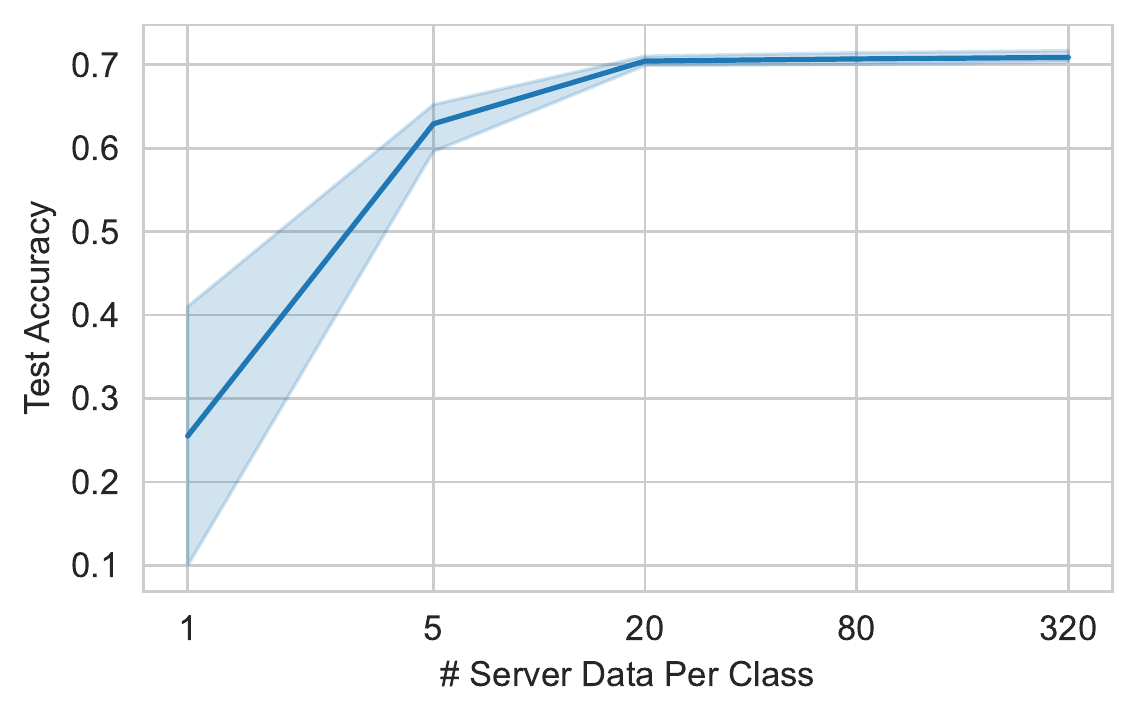}
    \caption{Effect of server data quantity. Error bars represent the s.d. of test accuracy over 5 random seeds.}
    \label{fig:hyper_num_data}
\end{figure*}

As depicted in Figure \ref{fig:hyper_num_data}, we observe that the test accuracy of our method increases as the number of server data per class rises from 1 to 20. It stabilizes once we have more than 20 samples per class. When the server data is limited to just one sample per class, our method's performance is suboptimal. However, with only 5 samples per class, our method already surpasses the highest-performing baseline AGR (MKrum with accuracy of 50.9\%). This demonstrates that our method demands only a small quantity of server data, which is readily achievable in real-world applications.

\newpage
\subsection{Effect of Hyperparameters (RQ3)}
\label{appendix:exp:hyperparameters}

\subsubsection{Effect of $f$ and $|\cB|$}

\begin{figure*}[h!]
    \centering
    \includegraphics[width=0.9\linewidth]{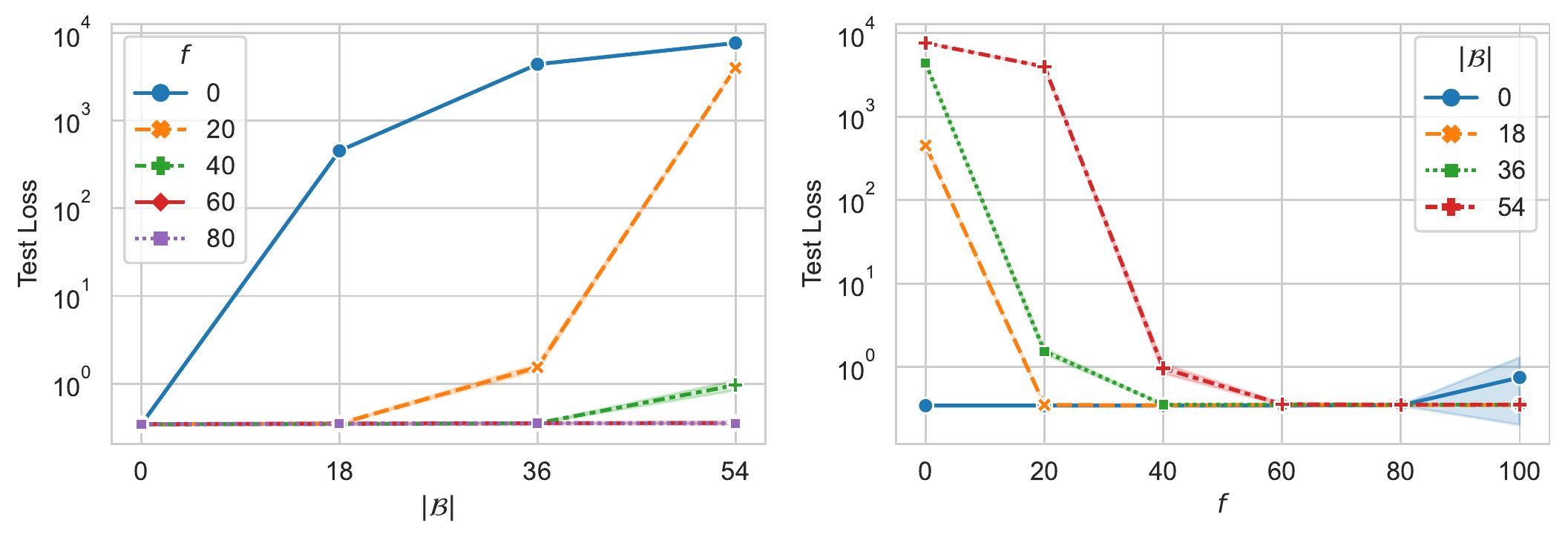}
    \caption{Effect of real number of Byzantines $|\cB|$ and Byzantine tolerance $f$. }
    \label{fig:hyper_Bf}
\end{figure*}

Similar to many robust AGRs (e.g., Krum \citep{krum} and TrMean \citep{trmean}), {\method} incorporates a hyperparameter denoted as $f$, which signifies Byzantine tolerance, i.e., the maximum number of attackers that the AGR is designed to withstand. Theoretically, an appropriate choice of $f$ should satisfy both $f \geq |\cB|$ and Assumption 5.3 simultaneously to achieve robustness. In this section, we empirically evaluate the performance of {\method} under various combinations of $f \in {0, 20, 40, 60, 80, 100}$ and $|\cB| \in {0, 18, 36, 54}$ using the AG-News dataset and the IPM attack. The results are depicted in Figure \ref{fig:hyper_Bf}. 

\paragraph{Effect of $|\cB|$ (given fixed $f$)}
In the main text, we investigate the influence of $|\cB|$ while keeping $f$ constant. Specifically, we examine two extreme scenarios: when $|\cB| = 0$ (where AGRs are most susceptible to selection bias) and when $|\cB| \approx f$ (where the AGRs are most susceptible to vulnerability). As depicted in Figure \ref{fig:hyper_Bf} (left), our findings reveal that, within a reasonable range of $f$ ($f \in [0, 80]$), the test loss remains minimal across all $|\cB| \in [0, f]$. This demonstrates that {\method} exhibits robustness to the actual number of Byzantines $|\cB|$ as long as $|\cB| \leq f$.

\paragraph{Effect of $f$}
In Figure \ref{fig:hyper_Bf} (right), our observations indicate the following:

\begin{itemize}
    \item For small values of $f$, {\method} exhibits robustness to only a limited number of Byzantine clients. For instance, when $f = 20$, {\method} demonstrates robustness when $|\cB| = 0$ and $|\cB| = 18$ but not when $|\cB| = 36$ or $|\cB| = 54$.
    \item As $f$ increases moderately, {\method} becomes more resilient to Byzantine clients while preserving its unbiasedness in scenarios with few or no Byzantines.
    \item However, when $f$ becomes excessively large (e.g., $f = 100$ in the context of $|\cH| = 160$), {\method} loses its unbiasedness in scenarios with small $|\cB|$. It's important to note that, when $|\cH| = 160$ and $|\cB| = 0$, we have $f = 100 > 80 = \frac{n}{2}$, implying an assumption that over half of the clients are Byzantine. Achieving optimal-order robustness under these conditions becomes impossible.
\end{itemize}

In summary, {\method} exhibits robustness across a broad range of $f$ values, and we recommend selecting a moderately larger $f$ to avoid underestimating $|\cB|$. 

\newpage
\subsubsection{Effect of $p_{\min}$}

In addition to the hyperparameter $f$, {\method} incorporates another parameter, $p_{\min}$, which is a slightly negative number intended to prevent excessive removal of honest clients during stage 2. In all experiments presented in the main text, spanning various datasets and models, we maintain a consistent value of $p_{\min} = -0.5$. In this section, we explore a range of $p_{\min}$ values to assess the sensitivity of {\method} to this hyperparameter. Specifically, we conduct experiments on the CIFAR-10 dataset with $p_{\min} \in \{-1.0, -0.7, -0.5, -0.2, -0.1, 0.0\}$ under two scenarios: no-attacks and $|\cB| = 15$ IPM attackers.

\begin{figure*}[h!]
    \centering
    \includegraphics[width=0.45\linewidth]{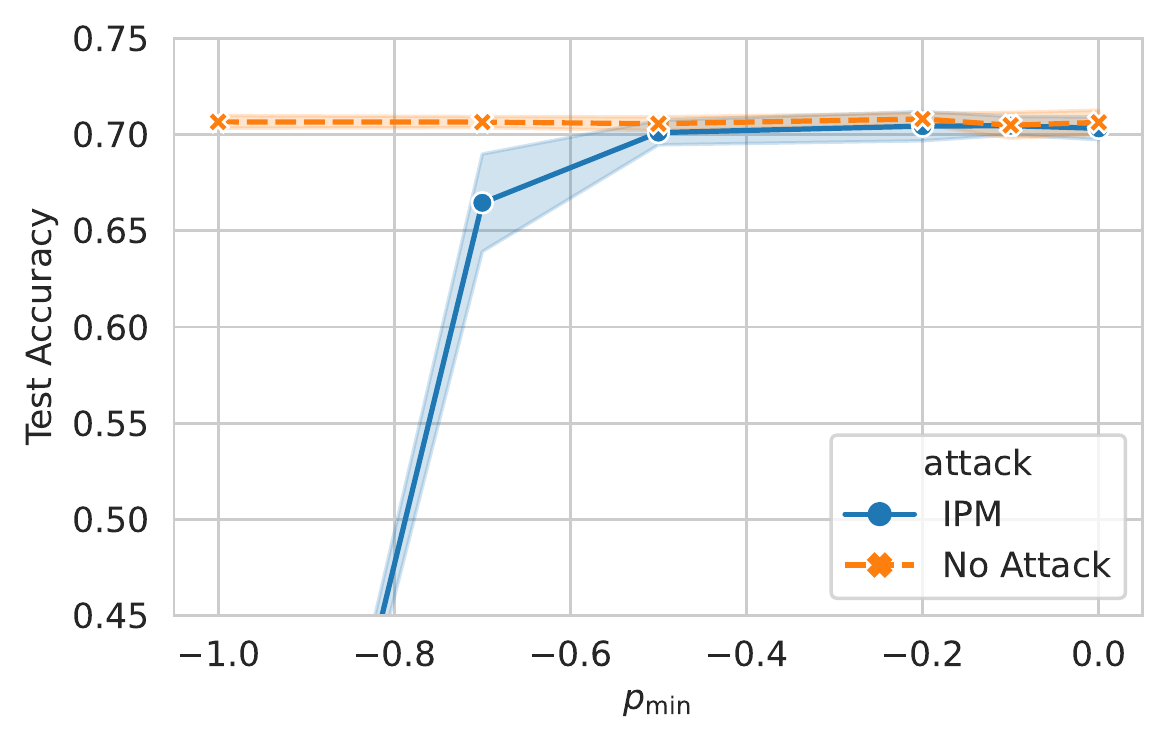}
    \caption{Effect of hyperparameter $p_{\min}$. }
    \label{fig:hyper_pmin}
\end{figure*}

The results displayed in Figure \ref{fig:hyper_pmin} illustrate that {\method} consistently delivers robust performance across a wide range of $p_{\min}$ values within the interval $[-0.5, 0.0]$. However, large absolute value for $p_{\min}$, such as $p_{\min} = -1.0$, fails to discard attackers and thus compromise the robustness of {\method}. Therefore, in practical applications, we recommend opting for a small absolute value for $p_{\min}$ to ensure robustness. 

\newpage
\subsection{More Label Skewness Settings (RQ3)}
\label{appendix:exp:noniid}

In the main test, we focus on pathological partition \citep{FedAvg}, a very challenging non-IID setting where each client only has two classes of data. In this setting, {\method} has state-of-the-art unbiasedness and robustness. In this part, we test {\method} with two more label skewness settings: step partition \citep{FedBE} and Dirichlet partition \citep{dirichlet}. We also test {\method} under partial participation. 

\subsubsection{Step Partition with Various Degrees of Non-IIDness}

In this part, we study how the performances of {\method} and other baseline AGRs change when the non-IID degree varies. We focus on the MNIST dataset with \textit{step partition} \citep{FedBE}: each client has 8 minor classes (with less data) and 2 major classes (with more data). We use a parameter $\alpha$ to control the ratio of major and minor class data size. Therefore, larger $\alpha$ indicates a larger non-IID degree. We test $\alpha \in \{1, 2, 4, 8, +\infty\}$. Notice that $\alpha = 1$ refers to the IID setting, and $\alpha = +\infty$ refers to pathological partition in the main text. 

\begin{figure*}[h!]
    \centering
    \includegraphics[width=1.0\linewidth]{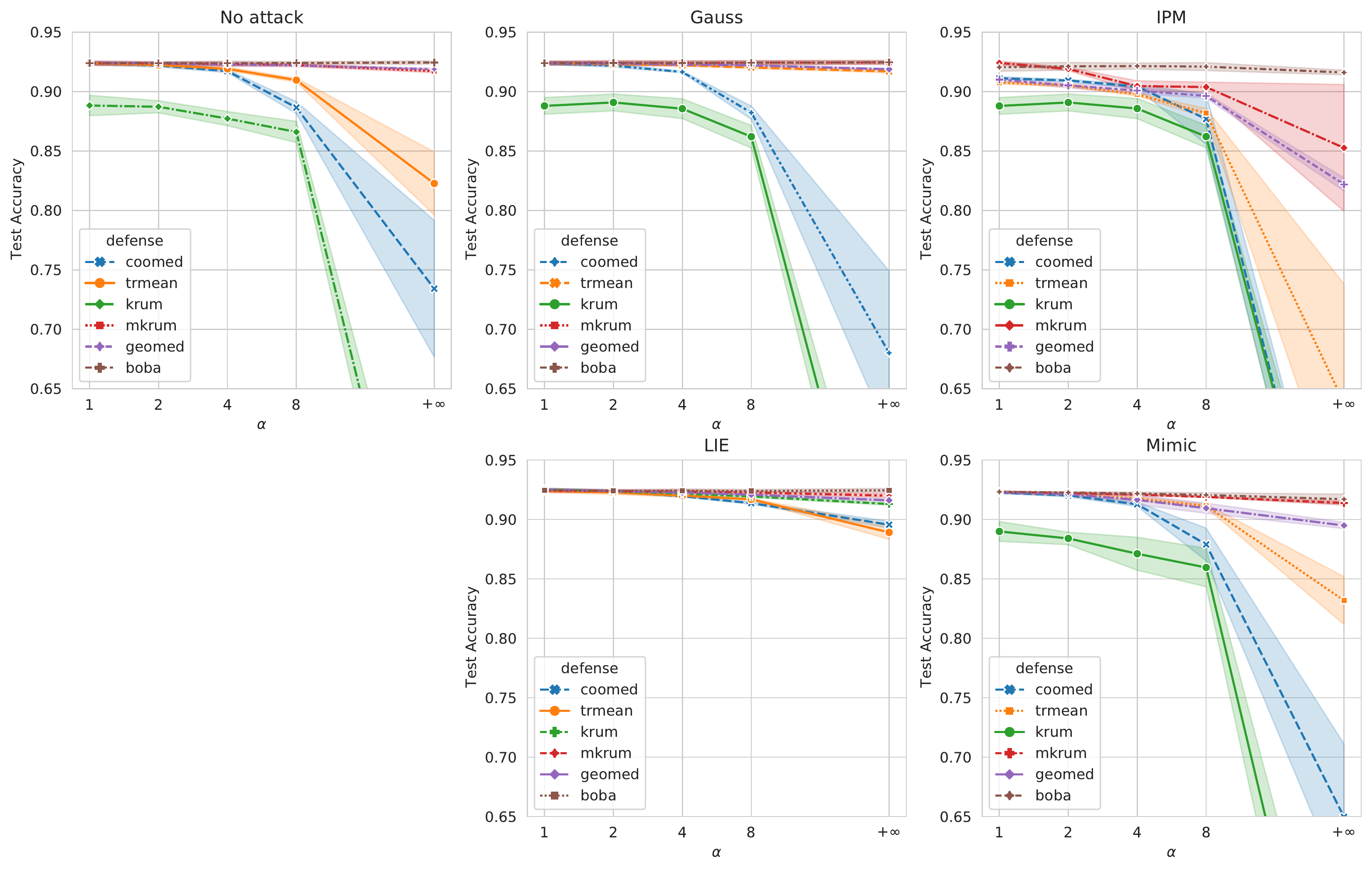}
    \caption{Effect of non-IIDness. Larger $\alpha$ indicates a larger non-IID degree. }
    \label{fig:hyper_noniid}
\end{figure*}

We show the test accuracy for both {\method} and selected baselines in Figure \ref{fig:hyper_noniid}. 
\begin{itemize}
    \item When the non-IID degree is small (e.g., $\alpha=1$), almost all the robust AGRs have satisfactory performance under all kinds of attacks. This observation matches the theoretical analysis of their Byzantine-robustness under the IID assumption. 
    \item However, when the non-IID degree ($\alpha$) increases, all the baseline AGRs degrade rapidly, especially under the IPM attack. This observation matches our claim that IID AGRs degrade under label skewness. 
    \item Finally, we notice that {\method} has almost constant performance under all attacks and non-IID degrees. This verifies our claim that {\method} has superior robustness and unbiasedness under label skewness. 
\end{itemize}

\subsubsection{Dirichlet Partition}

In this part, we use \textit{Dirichlet partition} ($\alpha = 0.01$) and compare {\method} to the strongest baselines. As shown in Table \ref{tab:dirichlet}, {\method} consistently outperforms baselines in both unbiasedness and robustness. 

\begin{table}[h]
\vspace{-1ex}
\caption{Performance (mean (s.d.) \%) under Dirichlet distribution ($\alpha = 0.01$)} \label{tab:dirichlet}
\vspace{1ex}
\centering
\resizebox{1.0\linewidth}{!}{
\begin{tabular}{cccccccccca}
\toprule
\multirow{2}*{Dataset} & \multirow{2}*{Method} & \multicolumn{2}{c}{$|\cB| = 0$} & \multicolumn{7}{c}{$|\cB| = 15$ (for MNIST and CIFAR-10) or $54$ (for AG-News) (Acc $\uparrow$)}\\
\cmidrule(lr){3-4} \cmidrule(lr){5-11}
& & Acc $\uparrow$ & MRD $\downarrow$ & Gauss & IPM & LIE & Mimic & MinMax & MinSum & Wst \\
\midrule
                & Average   & \meansd{\textbf{92.3}}{0.1} & - 
                            & \meansd{9.8}{0.0} & \meansd{9.8}{0.0} & \meansd{\textbf{92.3}}{0.1} 
                            & \meansd{\textbf{92.1}}{0.2} & \meansd{90.4}{0.1} & \meansd{90.5}{0.3} & 9.8 \\
MNIST           & MKrum     & \meansd{90.0}{1.1} & \meansd{24.0}{7.6}
                            & \meansd{\textbf{92.4}}{0.0} & \meansd{85.5}{3.5} & \meansd{91.0}{0.8} 
                            & \meansd{89.5}{1.4} & \meansd{83.8}{7.9} & \meansd{85.4}{4.1} & 83.8 \\
($|\cH|=100$)   & FLTrust   & \meansd{85.3}{1.0} & \meansd{20.1}{3.5}
                            & \meansd{85.3}{1.0} & \meansd{85.3}{1.0} & \meansd{87.9}{0.5} 
                            & \meansd{85.5}{0.6} & \meansd{85.2}{1.0} & \meansd{85.4}{0.6} & 85.2 \\
                & \method   & \meansd{\textbf{92.3}}{0.1} & \meansd{\textbf{1.9}}{1.9}
                            & \meansd{92.3}{0.1} & \meansd{\textbf{92.6}}{1.1} & \meansd{91.7}{0.4} 
                            & \meansd{\textbf{92.1}}{0.3} & \meansd{\textbf{91.8}}{0.4} & \meansd{\textbf{91.8}}{0.4} & \textbf{91.7} \\
                            
\midrule

                & Average   & \meansd{\textbf{71.6}}{0.4} & - 
                            & \meansd{10.0}{0.0} & \meansd{10.0}{0.0} & \meansd{35.6}{2.6} 
                            & \meansd{\textbf{70.3}}{0.7} & \meansd{35.3}{3.5} & \meansd{34.0}{2.4} & 34.0  \\
CIFAR-10        & MKrum     & \meansd{68.3}{0.4} & \meansd{27.3}{7.8}
                            & \meansd{\textbf{71.7}}{0.5} & \meansd{64.3}{2.7} & \meansd{43.6}{2.5}
                            & \meansd{67.2}{1.1} & \meansd{54.1}{6.9} & \meansd{41.1}{3.5} & 41.1 \\
($|\cH|=100$)   & FLTrust   & \meansd{49.1}{0.4} & \meansd{36.9}{6.0}
                            & \meansd{49.1}{0.5} & \meansd{47.9}{0.7} & \meansd{48.2}{0.8} 
                            & \meansd{49.5}{0.7} & \meansd{48.4}{0.8} & \meansd{48.3}{0.9} & 47.9 \\
                & \method   & \meansd{69.5}{1.2} & \meansd{\textbf{9.8}}{4.3}
                            & \meansd{\textbf{71.7}}{0.9} & \meansd{\textbf{70.9}}{0.5} & \meansd{\textbf{71.1}}{0.8} 
                            & \meansd{67.2}{2.0} & \meansd{\textbf{71.3}}{0.9} & \meansd{\textbf{71.2}}{0.9} & \textbf{67.2} \\

\midrule

                & Average   & \meansd{\textbf{88.3}}{0.1} & - 
                            & \meansd{24.3}{4.0} & \meansd{25.0}{0.0} & \meansd{87.0}{0.1} 
                            & \meansd{87.5}{0.5} & \meansd{36.0}{5.0} & \meansd{31.1}{3.9} & 24.3 \\
AG-News         & MKrum     & \meansd{80.8}{4.2} & \meansd{38.4}{19.8} 
                            & \meansd{\textbf{88.4}}{0.1} & \meansd{30.6}{9.2} & \meansd{83.2}{0.8} 
                            & \meansd{75.4}{7.3} & \meansd{85.7}{3.9} & \meansd{81.7}{1.7} & 30.6 \\
($|\cH|=160$)   & FLTrust   & \meansd{85.3}{0.6} & \meansd{7.8}{3.0}
                            & \meansd{85.5}{0.6} & \meansd{85.3}{0.6} & \meansd{85.3}{0.2} 
                            & \meansd{85.5}{0.5} & \meansd{85.3}{0.4} & \meansd{85.3}{0.4} & 85.3 \\
                & \method   & \meansd{88.2}{0.2} & \meansd{\textbf{1.3}}{2.4}
                            & \meansd{88.3}{0.1} & \meansd{\textbf{88.1}}{0.2} & \meansd{\textbf{88.3}}{0.1} 
                            & \meansd{\textbf{87.6}}{0.4} & \meansd{\textbf{88.1}}{0.2} & \meansd{\textbf{88.3}}{0.2} & \textbf{87.6} \\

\bottomrule
\end{tabular}
}
\end{table}

\subsubsection{Partial Participation}

{\method} also works under partial participation, i.e., only a subset of clients are selected for each round. We conduct experiments with AG-News dataset, under participation rate in $\{0.25, 0.50, 0.75, 1.00\}$. As shown in Table \ref{tab:partial}, {\method} consistently outperforms baselines across different participation rates. 

\begin{table}[h]
\vspace{-1ex}
\caption{Performance (mean (s.d.) \%) under partial participation on AG-News ($|\cH| = 160$)} \label{tab:partial}
\vspace{1ex}
\centering
\resizebox{1.0\linewidth}{!}{
\begin{tabular}{cccccccccca}
\toprule
\multirow{2}*{Participation Rate} & \multirow{2}*{Method} & \multicolumn{2}{c}{$|\cB| = 0$} & \multicolumn{7}{c}{$|\cB| = 54$ (Acc $\uparrow$)}\\
\cmidrule(lr){3-4} \cmidrule(lr){5-11}
& & Acc $\uparrow$ & MRD $\downarrow$ & Gauss & IPM & LIE & Mimic & MinMax & MinSum & Wst \\
\midrule
\multirow{4}*{0.25} 
& Average   & \meansd{\textbf{88.0}}{0.2} & - 
            & \meansd{25.7}{1.9} & \meansd{25.0}{0.0} & \meansd{\textbf{87.9}}{0.2} 
            & \meansd{\textbf{86.9}}{0.5} & \meansd{33.8}{5.0} & \meansd{81.3}{0.7} & 25.0 \\
& MKrum     & \meansd{87.3}{0.8} & \meansd{6.0}{2.9}
            & \meansd{\textbf{88.1}}{0.1} & \meansd{86.2}{0.6} & \meansd{87.8}{0.1} 
            & \meansd{82.6}{0.7} & \meansd{\textbf{87.9}}{0.2} & \meansd{86.1}{0.4} & 82.6 \\
& FLTrust   & \meansd{86.2}{0.4} & \meansd{7.6}{1.8}
            & \meansd{86.2}{0.6} & \meansd{86.2}{0.5} & \meansd{87.1}{0.4} 
            & \meansd{86.0}{0.4} & \meansd{85.8}{0.4} & \meansd{85.9}{0.5} & 85.8 \\
& \method   & \meansd{87.9}{0.2} & \meansd{\textbf{3.2}}{1.5}
            & \meansd{\textbf{88.1}}{0.2} & \meansd{\textbf{87.7}}{0.3} & \meansd{87.7}{0.3} 
            & \meansd{86.8}{0.5} & \meansd{87.7}{0.4} & \meansd{\textbf{87.8}}{0.3} & \textbf{86.8} \\
                            
\midrule

\multirow{4}*{0.50} 
& Average   & \meansd{\textbf{88.2}}{0.2} & -
            & \meansd{23.3}{1.9} & \meansd{25.0}{0.0} & \meansd{87.9}{0.3} 
            & \meansd{\textbf{87.2}}{0.4} & \meansd{36.4}{4.6} & \meansd{41.7}{21.1} & 23.3 \\
& MKrum     & \meansd{87.6}{0.5} & \meansd{5.1}{3.1}
            & \meansd{\textbf{88.2}}{0.2} & \meansd{84.4}{1.2} & \meansd{87.4}{0.5} 
            & \meansd{83.2}{1.9} & \meansd{\textbf{88.1}}{0.1} & \meansd{85.8}{0.3} & 83.2 \\
& FLTrust   & \meansd{86.3}{0.3} & \meansd{4.9}{1.5}
            & \meansd{86.3}{0.4} & \meansd{86.0}{1.0} & \meansd{86.7}{0.7} 
            & \meansd{86.0}{0.9} & \meansd{85.7}{0.9} & \meansd{85.6}{0.9} & 85.6 \\
& \method   & \meansd{\textbf{88.2}}{0.2} & \meansd{\textbf{2.9}}{2.5}
            & \meansd{88.1}{0.3} & \meansd{\textbf{87.8}}{0.2} & \meansd{\textbf{88.0}}{0.2} 
            & \meansd{87.1}{0.6} & \meansd{\textbf{88.1}}{0.2} & \meansd{\textbf{88.1}}{0.2} & \textbf{87.1} \\

\midrule

\multirow{4}*{0.75} 
& Average   & \meansd{\textbf{88.3}}{0.1} & - 
            & \meansd{25.2}{3.6} & \meansd{25.0}{0.0} & \meansd{87.9}{0.1} 
            & \meansd{\textbf{87.3}}{0.4} & \meansd{30.5}{3.3} & \meansd{29.8}{4.7} & 25.0 \\
& MKrum     & \meansd{87.8}{0.2} & \meansd{4.9}{1.6}
            & \meansd{88.3}{0.0} & \meansd{48.8}{32.6} & \meansd{87.1}{0.6} 
            & \meansd{83.2}{1.1} & \meansd{\textbf{88.2}}{0.1} & \meansd{86.2}{0.4} & 48.8 \\
& FLTrust   & \meansd{86.3}{0.4} & \meansd{5.4}{1.6}
            & \meansd{86.3}{0.5} & \meansd{86.2}{0.4} & \meansd{86.4}{0.3} 
            & \meansd{85.9}{0.7} & \meansd{86.0}{0.4} & \meansd{85.9}{0.5} & 85.9 \\
& \method   & \meansd{\textbf{88.3}}{0.1} & \meansd{\textbf{1.7}}{0.3}
            & \meansd{\textbf{88.4}}{0.1} & \meansd{\textbf{88.0}}{0.2} & \meansd{\textbf{88.3}}{0.1} 
            & \meansd{87.1}{0.5} & \meansd{88.0}{0.3} & \meansd{\textbf{88.1}}{0.3} & \textbf{87.1} \\

\midrule

\multirow{4}*{1.00} 
& Average   & \meansd{\textbf{88.3}}{0.1} & - 
            & \meansd{25.4}{2.6} & \meansd{25.0}{0.0} & \meansd{87.5}{0.2} 
            & \meansd{87.2}{0.3} & \meansd{35.9}{3.6} & \meansd{30.5}{3.0} & 25.0 \\
& MKrum     & \meansd{88.0}{0.1} & \meansd{4.6}{2.1} 
            & \meansd{\textbf{88.3}}{0.2} & \meansd{80.7}{6.0} & \meansd{86.6}{0.2} 
            & \meansd{83.4}{0.6} & \meansd{\textbf{88.3}}{0.1} & \meansd{85.9}{0.3} & 80.7 \\
& FLTrust   & \meansd{86.3}{0.4} & \meansd{5.8}{1.0} 
            & \meansd{86.2}{0.5} & \meansd{86.2}{0.4} & \meansd{86.2}{0.4} 
            & \meansd{85.7}{0.8} & \meansd{85.8}{0.9} & \meansd{85.8}{0.5} & 85.7 \\
& \method   & \meansd{\textbf{88.3}}{0.1} & \meansd{\textbf{0.2}}{0.1} 
            & \meansd{\textbf{88.3}}{0.1} & \meansd{\textbf{87.7}}{0.7} & \meansd{\textbf{88.4}}{0.1} 
            & \meansd{\textbf{87.3}}{0.3} & \meansd{88.1}{0.1} & \meansd{\textbf{88.3}}{0.2} & \textbf{87.3} \\

\bottomrule
\end{tabular}
}
\end{table}

\newpage
\subsection{Experiments with Both Label and Feature Skewness (RQ4)}
\label{appendix:exp:feat}

{\method} is motivated by label skewness, where each honest client possesses a different label distribution and the same label-conditioned data distribution. However, practical FL systems may have more complex non-IIDness, with both label and feature distribution potentially varying. For example, as mentioned in our introduction with the example of animal image classification, different users may not only capture different prevalent species in their region but also exhibit variations in image appearance due to different camera settings. To validate whether {\method} remains effective in such a more complex non-IID setting, alongside generating label skewness using pathological partition, we inject different image corruption to each client. 

Specifically, each client randomly selects one from the 15 common image corruptions \citep{corruption} and applies it to all of their training data. Consequently, even for images of the same class, there will be varying feature distributions across different clients. To facilitate a comparison with results obtained without image corruption, we do not add image corruptions to the testing data.

\begin{table}[h]
\vspace{-1ex}
\caption{Performance (mean (s.d.) \%) on CIFAR-10 with label skewness and image corruptions} \label{tab:corruption_full}
\vspace{1ex}
\centering
\begin{tabular}{ccccccccca}
\toprule
\multirow{2}*{Method} & \multicolumn{2}{c}{$|\cB| = 0$} & \multicolumn{7}{c}{$|\cB| = 15$ (Acc $\uparrow$)}\\
\cmidrule(lr){2-3} \cmidrule(lr){4-10}
& Acc $\uparrow$ & MRD $\downarrow$ & Gauss & IPM & LIE & Mimic & MinMax & MinSum & Wst \\
\midrule
Average & \meansd{68.7}{0.4} & - 
        & \meansd{10.0}{0.0} & \meansd{10.0}{0.0} & \meansd{64.6}{0.7} & \meansd{67.5}{0.5} & \meansd{27.9}{4.9} & \meansd{21.6}{7.5} & 10.0 \\
CooMed  & \meansd{19.1}{4.9} & \meansd{78.4}{2.0} 
        & \meansd{20.1}{1.5} & \meansd{9.4}{1.8} & \meansd{24.0}{2.1} & \meansd{17.8}{1.8} & \meansd{17.7}{1.2} & \meansd{17.7}{1.2} & 9.4 \\
TrMean  & \meansd{24.1}{2.9} & \meansd{78.8}{2.7}
        & \meansd{53.8}{1.8} & \meansd{14.2}{4.4} & \meansd{30.6}{1.1} & \meansd{20.1}{5.4} & \meansd{20.5}{0.6} & \meansd{22.2}{2.0} & 14.2\\
Krum    & \meansd{33.4}{2.9} & \meansd{79.3}{3.7}
        & \meansd{34.4}{3.0} & \meansd{33.1}{2.0} & \meansd{38.7}{2.3} & \meansd{31.0}{3.2} & \meansd{34.1}{1.8} & \meansd{33.2}{2.6} & 31.0 \\
MKrum   & \meansd{66.8}{1.1} & \meansd{16.7}{11.7} 
        & \meansd{68.2}{0.7} & \meansd{52.9}{10.2} & \meansd{63.1}{1.1} & \meansd{54.9}{25.1} & \meansd{67.2}{0.4} & \meansd{62.3}{2.1} & 52.9 \\
GeoMed  & \meansd{68.2}{0.6} & \meansd{5.3}{1.5} 
        & \meansd{67.9}{0.9} & \meansd{55.8}{4.6} & \meansd{42.5}{2.7} & \meansd{53.6}{5.0} & \meansd{42.7}{3.0} & \meansd{42.6}{2.9} & 42.5 \\
SelfRej & \meansd{66.3}{1.4} & \meansd{21.8}{13.4}
        & \meansd{67.8}{0.4} & \meansd{26.6}{6.5} & \meansd{63.1}{1.1} & \meansd{65.9}{0.8} & \meansd{25.0}{2.5} & \meansd{25.0}{2.8} & 25.0\\
AvgRej  & \meansd{67.1}{1.7} & \meansd{25.2}{20.3}
        & \meansd{10.0}{0.0} & \meansd{66.5}{0.8} & \meansd{63.6}{0.8} & \meansd{68.1}{0.6} & \meansd{54.5}{6.2} & \meansd{53.4}{4.7} & 10.0\\
Zeno    & \meansd{66.3}{1.6} & \meansd{23.8}{15.6} 
        & \meansd{67.8}{0.6} & \meansd{26.7}{6.5} & \meansd{63.2}{1.3} & \meansd{66.2}{1.4} & \meansd{23.8}{4.0} & \meansd{23.5}{2.3} & 23.5\\
FLTrust & \meansd{50.1}{0.9} & \meansd{29.1}{2.1} & \meansd{50.0}{1.1} & \meansd{47.8}{1.7} & \meansd{47.3}{2.5} & \meansd{49.8}{0.9} & \meansd{49.0}{1.8} & \meansd{49.1}{1.9} & 47.3 \\
ByGARS  & \meansd{29.2}{2.0} & \meansd{57.7}{4.2}
        & \meansd{29.1}{2.0} & \meansd{50.9}{0.8} & \meansd{27.3}{2.8} & \meansd{29.7}{1.7} & \meansd{24.4}{0.9} & \meansd{24.4}{0.7} & 24.4\\
B-Krum  & \meansd{52.4}{2.1} & \meansd{69.6}{10.0}
        & \meansd{58.1}{1.2} & \meansd{55.8}{1.2} & \meansd{41.1}{1.2} & \meansd{43.5}{2.3} & \meansd{57.7}{1.7} & \meansd{57.9}{1.1} & 41.1\\
B-MKrum & \meansd{68.1}{0.4} & \meansd{6.1}{3.0}
        & \meansd{68.2}{0.8} & \meansd{50.2}{6.6} & \meansd{63.1}{1.1} & \meansd{65.6}{2.2} & \meansd{49.7}{3.4} & \meansd{45.3}{6.7} & 45.3\\
RAGE    & \meansd{57.2}{4.4} & \meansd{45.6}{17.6} 
        & \meansd{65.7}{0.5} & \meansd{57.5}{2.0} & \meansd{45.3}{2.8} & \meansd{59.3}{2.1} & \meansd{58.5}{3.7} & \meansd{58.2}{4.9} & 45.3 \\
\method & \meansd{66.5}{1.0} & \meansd{6.7}{2.6} & \meansd{68.5}{0.3} & \meansd{66.0}{0.7} & \meansd{62.8}{1.6} & \meansd{66.2}{0.7} & \meansd{67.7}{0.5} & \meansd{67.5}{0.6} & 62.8 \\
\bottomrule
\end{tabular}
\end{table}

We have summarized the experimental results in Table \ref{tab:corruption_full}. Due to the introduction of perturbation in our training data, the performance of virtually all aggregators has deteriorated. It is worth noting that even in the absence of attacks, the accuracy of the average aggregator has decreased from 71.7 to 68.7. However, in this scenario, {\method} still exhibits better robustness than all the baseline methods, while also being more unbiased than the majority of baselines. This suggests that {\method} can generalize to more complex non-IID settings that exhibit both feature and label skewness.

\newpage
\subsection{Extension to More FL Frameworks (RQ4)}
\label{appendix:exp:more_fl}

In this part, we empirically show that our {\method} can be integrated with more FL algorithms with different local update function. Specifically, we consider FedAvg \citep{FedAvg} with $E = 5$ local epochs (instead of $E=1$ for FedSGD) and FedProx \citep{FedProx} with $E = 5$ and local regularization hyperparameter $\mu=0.01$. 

With multiple local gradient descent steps, slightly abusing notation, we define the pseudo-gradient as follows: 
\begin{align*}
    \vg_i = - (\vw_i - \vw_G)
\end{align*}
where $\vw_G$ is the global parameter send to client $i$ (before local update), and $\vg_i$ is the local parameter after local update. 

\begin{figure*}[h!]
    \centering
    \includegraphics[width=1.0\linewidth]{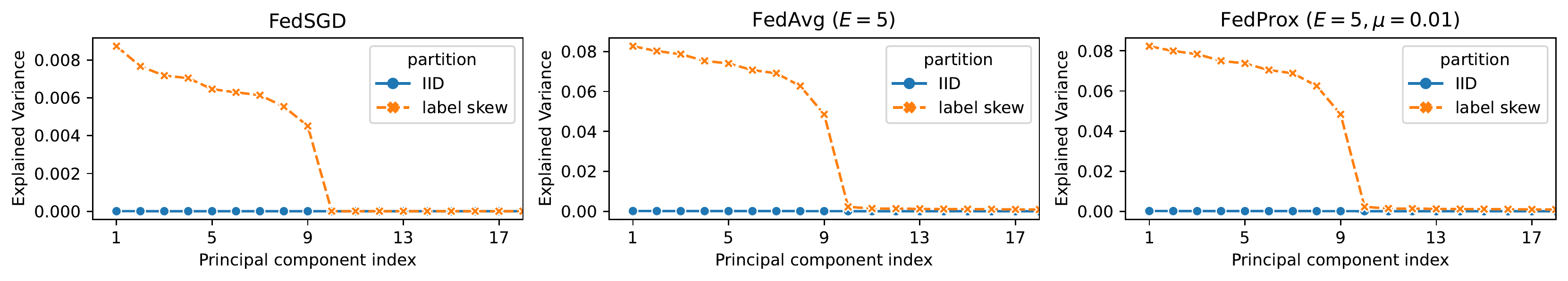}
    \caption{PCA of honest gradients on MNIST ($c = 10$)}
    \label{fig:pca2}
\end{figure*}

First of all, we empirically verify that Proposition 1 still approximately holds, even when using different local update functions. With random initialization, we save all 100 honest (pseudo-)gradients and conduct principal component analysis on them, under both IID and label skew settings. We sort all principal components with their explained variance (from large to small), and plot them in Figure \ref{fig:pca2}. 

It shows that (1) the total variance for label skewness setting is much larger than IID setting, and (2) most of the variances among honest gradients concentrate in the first $c - 1= 9$ principal components. It verifies that Proposition \ref{proposition:dist} still approximately holds for FedAvg and FedProx. 

Then, we evaluate whether {\method} can generalize these two FL frameworks. We run experiments with MNIST data set. 

\begin{figure*}[h!]
    \centering
    \includegraphics[width=0.45\linewidth]{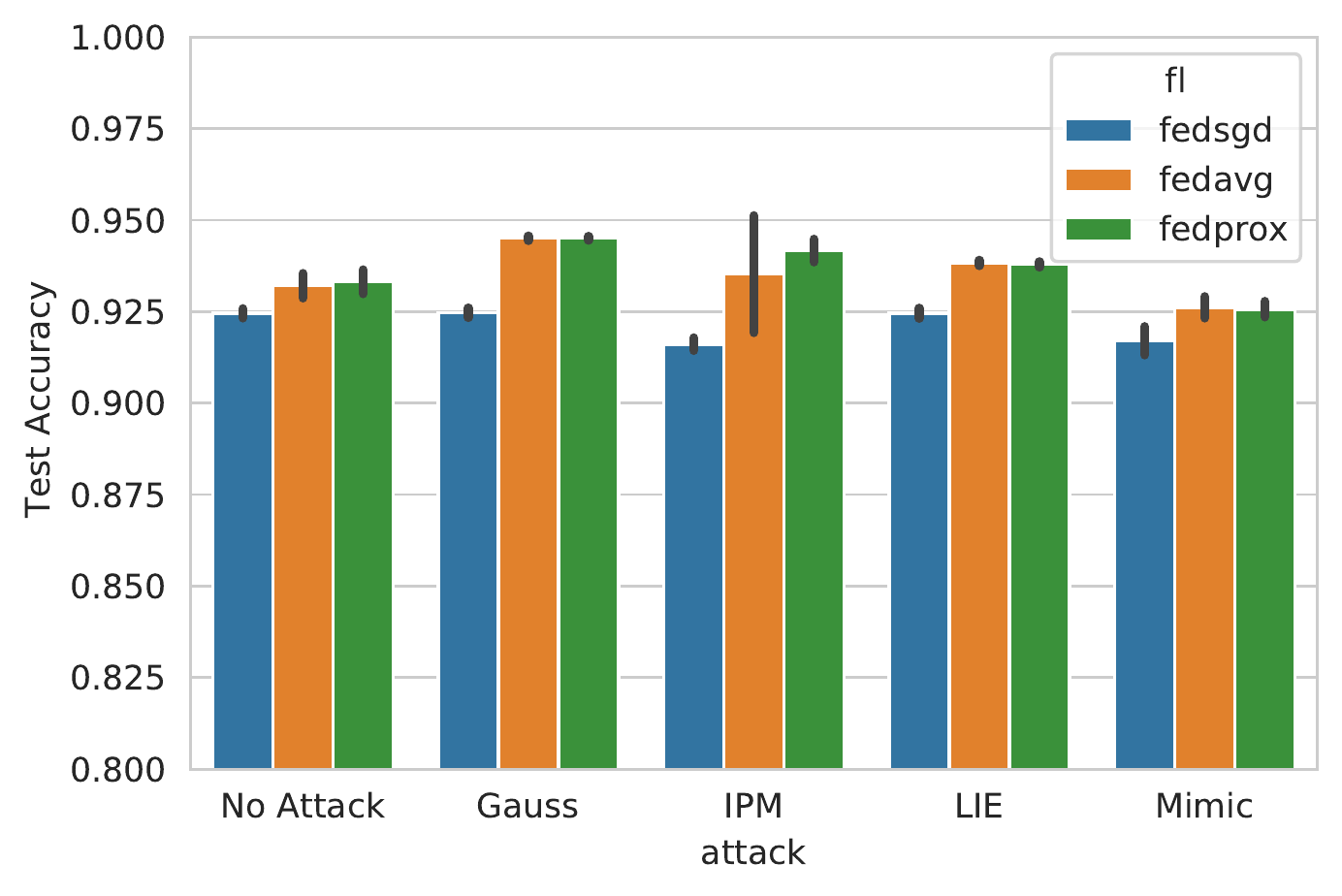}
    \caption{Applying {\method} on more FL frameworks}
    \label{fig:more_fl}
\end{figure*}

Figure \ref{fig:more_fl} shows that {\method} can generalize FedAvg and FedProx. With the same number of communication rounds, {\method} + FedAvg/FedProx achieve higher accuracy, indicating their faster convergence. Meanwhile, {\method} remains its unbiasedness and robustness across all attacks.

\end{document}